\documentclass[11pt]{article}%
\usepackage{lettrine}%
\usepackage{xcolor}%
\usepackage[breakable]{tcolorbox}%
\usepackage{parskip} 
\usepackage{iftex}%
\usepackage{makecell}
\usepackage{xcolor}
\usepackage{array}
\usepackage{tikz}
\usetikzlibrary{arrows.meta,positioning}

\ifPDFTeX%
    \usepackage[T1]{fontenc}%
    \usepackage{mathpazo}%
\else%
    \usepackage{fontspec}%
\fi%
\usepackage{fancyhdr}%
    \usepackage{graphicx}%
    \usepackage{caption}%

    \usepackage{float}%
    \usepackage{xcolor} 
    \usepackage{enumerate} 
    \usepackage{geometry} 
    \usepackage{amsmath} 
    \usepackage{amssymb} 
    \usepackage{textcomp} 
    \AtBeginDocument{%
    }%
    \usepackage{upquote} 
    \usepackage{eurosym} 
    \ifPDFTeX
    \usepackage[mathletters]{ucs} 
    \fi
    \usepackage{soul}%
    \usepackage{fancyvrb} 
    \usepackage{grffile} 
    \makeatletter 
    \@ifpackagelater{grffile}{2019/11/01}%
    {%
    }%
    {%
      \def\Gread@@xetex#1{%
        \IfFileExists{"\Gin@base".bb}%
        {\Gread@eps{\Gin@base.bb}}%
        {\Gread@@xetex@aux#1}%
      }%
    }%
    \makeatother%
    \usepackage[Export]{adjustbox} 
    \adjustboxset{max size={0.9\linewidth}{0.9\paperheight}}%

    \usepackage{hyperref}%
    \usepackage{titling}%
    \usepackage{longtable} 
    \usepackage{booktabs}  
    \usepackage[inline]{enumitem} 
    \usepackage[normalem]{ulem} 
    \usepackage{mathrsfs}%

    \definecolor{urlcolor}{rgb}{0,.145,.698}%
    \definecolor{linkcolor}{rgb}{.71,0.21,0.01}%
    \definecolor{citecolor}{rgb}{.12,.54,.11}%

    \definecolor{ansi-black}{HTML}{3E424D}%
    \definecolor{ansi-black-intense}{HTML}{282C36}%
    \definecolor{ansi-red}{HTML}{E75C58}%
    \definecolor{ansi-red-intense}{HTML}{B22B31}%
    \definecolor{ansi-green}{HTML}{00A250}%
    \definecolor{ansi-green-intense}{HTML}{007427}%
    \definecolor{ansi-yellow}{HTML}{DDB62B}%
    \definecolor{ansi-yellow-intense}{HTML}{B27D12}%
    \definecolor{ansi-blue}{HTML}{208FFB}%
    \definecolor{ansi-blue-intense}{HTML}{0065CA}%
    \definecolor{ansi-magenta}{HTML}{D160C4}%
    \definecolor{ansi-magenta-intense}{HTML}{A03196}%
    \definecolor{ansi-cyan}{HTML}{60C6C8}%
    \definecolor{ansi-cyan-intense}{HTML}{258F8F}%
    \definecolor{ansi-white}{HTML}{C5C1B4}%
    \definecolor{ansi-white-intense}{HTML}{A1A6B2}%
    \definecolor{ansi-default-inverse-fg}{HTML}{FFFFFF}%
    \definecolor{ansi-default-inverse-bg}{HTML}{000000}%

    \definecolor{outerrorbackground}{HTML}{FFDFDF}%

    \DefineVerbatimEnvironment{Highlighting}{Verbatim}{commandchars=\\\{\}}%
    \let\Oldtex\TeX%
    \let\Oldlatex\LaTeX%
    \renewcommand{\TeX}{\textrm{\Oldtex}}%
    \renewcommand{\LaTeX}{\textrm{\Oldlatex}}%
\makeatletter%
\def\PY@reset{\let\PY@it=\relax \let\PY@bf=\relax%
    \let\PY@ul=\relax \let\PY@tc=\relax%
    \let\PY@bc=\relax \let\PY@ff=\relax}%
\def\PY@tok#1{\csname PY@tok@#1\endcsname}%
\def\PY@toks#1+{\ifx\relax#1\empty\else%
    \PY@tok{#1}\expandafter\PY@toks\fi}%
\def\PY@do#1{\PY@bc{\PY@tc{\PY@ul{%
    \PY@it{\PY@bf{\PY@ff{#1}}}}}}}%
\def\PY#1#2{\PY@reset\PY@toks#1+\relax+\PY@do{#2}}%

\@namedef{PY@tok@w}{\def\PY@tc##1{\textcolor[rgb]{0.73,0.73,0.73}{##1}}}%
\@namedef{PY@tok@c}{\let\PY@it=\textit\def\PY@tc##1{\textcolor[rgb]{0.25,0.50,0.50}{##1}}}%
\@namedef{PY@tok@cp}{\def\PY@tc##1{\textcolor[rgb]{0.74,0.48,0.00}{##1}}}%
\@namedef{PY@tok@k}{\let\PY@bf=\textbf\def\PY@tc##1{\textcolor[rgb]{0.00,0.50,0.00}{##1}}}%
\@namedef{PY@tok@kp}{\def\PY@tc##1{\textcolor[rgb]{0.00,0.50,0.00}{##1}}}%
\@namedef{PY@tok@kt}{\def\PY@tc##1{\textcolor[rgb]{0.69,0.00,0.25}{##1}}}%
\@namedef{PY@tok@o}{\def\PY@tc##1{\textcolor[rgb]{0.40,0.40,0.40}{##1}}}%
\@namedef{PY@tok@ow}{\let\PY@bf=\textbf\def\PY@tc##1{\textcolor[rgb]{0.67,0.13,1.00}{##1}}}%
\@namedef{PY@tok@nb}{\def\PY@tc##1{\textcolor[rgb]{0.00,0.50,0.00}{##1}}}%
\@namedef{PY@tok@nf}{\def\PY@tc##1{\textcolor[rgb]{0.00,0.00,1.00}{##1}}}%
\@namedef{PY@tok@nc}{\let\PY@bf=\textbf\def\PY@tc##1{\textcolor[rgb]{0.00,0.00,1.00}{##1}}}%
\@namedef{PY@tok@nn}{\let\PY@bf=\textbf\def\PY@tc##1{\textcolor[rgb]{0.00,0.00,1.00}{##1}}}%
\@namedef{PY@tok@ne}{\let\PY@bf=\textbf\def\PY@tc##1{\textcolor[rgb]{0.82,0.25,0.23}{##1}}}%
\@namedef{PY@tok@nv}{\def\PY@tc##1{\textcolor[rgb]{0.10,0.09,0.49}{##1}}}%
\@namedef{PY@tok@no}{\def\PY@tc##1{\textcolor[rgb]{0.53,0.00,0.00}{##1}}}%
\@namedef{PY@tok@nl}{\def\PY@tc##1{\textcolor[rgb]{0.63,0.63,0.00}{##1}}}%
\@namedef{PY@tok@ni}{\let\PY@bf=\textbf\def\PY@tc##1{\textcolor[rgb]{0.60,0.60,0.60}{##1}}}%
\@namedef{PY@tok@na}{\def\PY@tc##1{\textcolor[rgb]{0.49,0.56,0.16}{##1}}}%
\@namedef{PY@tok@nt}{\let\PY@bf=\textbf\def\PY@tc##1{\textcolor[rgb]{0.00,0.50,0.00}{##1}}}%
\@namedef{PY@tok@nd}{\def\PY@tc##1{\textcolor[rgb]{0.67,0.13,1.00}{##1}}}%
\@namedef{PY@tok@s}{\def\PY@tc##1{\textcolor[rgb]{0.73,0.13,0.13}{##1}}}%
\@namedef{PY@tok@sd}{\let\PY@it=\textit\def\PY@tc##1{\textcolor[rgb]{0.73,0.13,0.13}{##1}}}%
\@namedef{PY@tok@si}{\let\PY@bf=\textbf\def\PY@tc##1{\textcolor[rgb]{0.73,0.40,0.53}{##1}}}%
\@namedef{PY@tok@se}{\let\PY@bf=\textbf\def\PY@tc##1{\textcolor[rgb]{0.73,0.40,0.13}{##1}}}%
\@namedef{PY@tok@sr}{\def\PY@tc##1{\textcolor[rgb]{0.73,0.40,0.53}{##1}}}%
\@namedef{PY@tok@ss}{\def\PY@tc##1{\textcolor[rgb]{0.10,0.09,0.49}{##1}}}%
\@namedef{PY@tok@sx}{\def\PY@tc##1{\textcolor[rgb]{0.00,0.50,0.00}{##1}}}%
\@namedef{PY@tok@m}{\def\PY@tc##1{\textcolor[rgb]{0.40,0.40,0.40}{##1}}}%
\@namedef{PY@tok@gh}{\let\PY@bf=\textbf\def\PY@tc##1{\textcolor[rgb]{0.00,0.00,0.50}{##1}}}%
\@namedef{PY@tok@gu}{\let\PY@bf=\textbf\def\PY@tc##1{\textcolor[rgb]{0.50,0.00,0.50}{##1}}}%
\@namedef{PY@tok@gd}{\def\PY@tc##1{\textcolor[rgb]{0.63,0.00,0.00}{##1}}}%
\@namedef{PY@tok@gi}{\def\PY@tc##1{\textcolor[rgb]{0.00,0.63,0.00}{##1}}}%
\@namedef{PY@tok@gr}{\def\PY@tc##1{\textcolor[rgb]{1.00,0.00,0.00}{##1}}}%
\@namedef{PY@tok@ge}{\let\PY@it=\textit}%
\@namedef{PY@tok@gs}{\let\PY@bf=\textbf}%
\@namedef{PY@tok@gp}{\let\PY@bf=\textbf\def\PY@tc##1{\textcolor[rgb]{0.00,0.00,0.50}{##1}}}%
\@namedef{PY@tok@go}{\def\PY@tc##1{\textcolor[rgb]{0.53,0.53,0.53}{##1}}}%
\@namedef{PY@tok@gt}{\def\PY@tc##1{\textcolor[rgb]{0.00,0.27,0.87}{##1}}}%
\@namedef{PY@tok@err}{\def\PY@bc##1{{\setlength{\fboxsep}{\string -\fboxrule}\fcolorbox[rgb]{1.00,0.00,0.00}{1,1,1}{\strut ##1}}}}%
\@namedef{PY@tok@kc}{\let\PY@bf=\textbf\def\PY@tc##1{\textcolor[rgb]{0.00,0.50,0.00}{##1}}}%
\@namedef{PY@tok@kd}{\let\PY@bf=\textbf\def\PY@tc##1{\textcolor[rgb]{0.00,0.50,0.00}{##1}}}%
\@namedef{PY@tok@kn}{\let\PY@bf=\textbf\def\PY@tc##1{\textcolor[rgb]{0.00,0.50,0.00}{##1}}}%
\@namedef{PY@tok@kr}{\let\PY@bf=\textbf\def\PY@tc##1{\textcolor[rgb]{0.00,0.50,0.00}{##1}}}%
\@namedef{PY@tok@bp}{\def\PY@tc##1{\textcolor[rgb]{0.00,0.50,0.00}{##1}}}%
\@namedef{PY@tok@fm}{\def\PY@tc##1{\textcolor[rgb]{0.00,0.00,1.00}{##1}}}%
\@namedef{PY@tok@vc}{\def\PY@tc##1{\textcolor[rgb]{0.10,0.09,0.49}{##1}}}%
\@namedef{PY@tok@vg}{\def\PY@tc##1{\textcolor[rgb]{0.10,0.09,0.49}{##1}}}%
\@namedef{PY@tok@vi}{\def\PY@tc##1{\textcolor[rgb]{0.10,0.09,0.49}{##1}}}%
\@namedef{PY@tok@vm}{\def\PY@tc##1{\textcolor[rgb]{0.10,0.09,0.49}{##1}}}%
\@namedef{PY@tok@sa}{\def\PY@tc##1{\textcolor[rgb]{0.73,0.13,0.13}{##1}}}%
\@namedef{PY@tok@sb}{\def\PY@tc##1{\textcolor[rgb]{0.73,0.13,0.13}{##1}}}%
\@namedef{PY@tok@sc}{\def\PY@tc##1{\textcolor[rgb]{0.73,0.13,0.13}{##1}}}%
\@namedef{PY@tok@dl}{\def\PY@tc##1{\textcolor[rgb]{0.73,0.13,0.13}{##1}}}%
\@namedef{PY@tok@s2}{\def\PY@tc##1{\textcolor[rgb]{0.73,0.13,0.13}{##1}}}%
\@namedef{PY@tok@sh}{\def\PY@tc##1{\textcolor[rgb]{0.73,0.13,0.13}{##1}}}%
\@namedef{PY@tok@s1}{\def\PY@tc##1{\textcolor[rgb]{0.73,0.13,0.13}{##1}}}%
\@namedef{PY@tok@mb}{\def\PY@tc##1{\textcolor[rgb]{0.40,0.40,0.40}{##1}}}%
\@namedef{PY@tok@mf}{\def\PY@tc##1{\textcolor[rgb]{0.40,0.40,0.40}{##1}}}%
\@namedef{PY@tok@mh}{\def\PY@tc##1{\textcolor[rgb]{0.40,0.40,0.40}{##1}}}%
\@namedef{PY@tok@mi}{\def\PY@tc##1{\textcolor[rgb]{0.40,0.40,0.40}{##1}}}%
\@namedef{PY@tok@il}{\def\PY@tc##1{\textcolor[rgb]{0.40,0.40,0.40}{##1}}}%
\@namedef{PY@tok@mo}{\def\PY@tc##1{\textcolor[rgb]{0.40,0.40,0.40}{##1}}}%
\@namedef{PY@tok@ch}{\let\PY@it=\textit\def\PY@tc##1{\textcolor[rgb]{0.25,0.50,0.50}{##1}}}%
\@namedef{PY@tok@cm}{\let\PY@it=\textit\def\PY@tc##1{\textcolor[rgb]{0.25,0.50,0.50}{##1}}}%
\@namedef{PY@tok@cpf}{\let\PY@it=\textit\def\PY@tc##1{\textcolor[rgb]{0.25,0.50,0.50}{##1}}}%
\@namedef{PY@tok@c1}{\let\PY@it=\textit\def\PY@tc##1{\textcolor[rgb]{0.25,0.50,0.50}{##1}}}%
\@namedef{PY@tok@cs}{\let\PY@it=\textit\def\PY@tc##1{\textcolor[rgb]{0.25,0.50,0.50}{##1}}}%

\makeatother%
    \makeatletter%
        \newbox\Wrappedcontinuationbox %
        \newbox\Wrappedvisiblespacebox %
        \newcommand*\Wrappedvisiblespace {\textcolor{red}{\textvisiblespace}} %
        \newcommand*\Wrappedcontinuationsymbol {\textcolor{red}{\llap{\tiny$\m@th\hookrightarrow$}}} %
        \newcommand*\Wrappedcontinuationindent {3ex } %
        \newcommand*\Wrappedafterbreak {\kern\Wrappedcontinuationindent\copy\Wrappedcontinuationbox} %
        \newcommand*\Wrappedbreaksatspecials {%
            \def\PYGZus{\discretionary{\char`\_}{\Wrappedafterbreak}{\char`\_}}%
            \def\PYGZob{\discretionary{}{\Wrappedafterbreak\char`\{}{\char`\{}}%
            \def\PYGZcb{\discretionary{\char`\}}{\Wrappedafterbreak}{\char`\}}}%
            \def\PYGZca{\discretionary{\char`\^}{\Wrappedafterbreak}{\char`\^}}%
            \def\PYGZam{\discretionary{\char`\&}{\Wrappedafterbreak}{\char`\&}}%
            \def\PYGZlt{\discretionary{}{\Wrappedafterbreak\char`\<}{\char`\<}}%
            \def\PYGZgt{\discretionary{\char`\>}{\Wrappedafterbreak}{\char`\>}}%
            \def\PYGZsh{\discretionary{}{\Wrappedafterbreak\char`\#}{\char`\#}}%
            \def\PYGZpc{\discretionary{}{\Wrappedafterbreak\char`\%}{\char`\%}}%
            \def\PYGZdl{\discretionary{}{\Wrappedafterbreak\char`\$}{\char`\$}}%
            \def\PYGZhy{\discretionary{\char`\-}{\Wrappedafterbreak}{\char`\-}}%
            \def\PYGZsq{\discretionary{}{\Wrappedafterbreak\textquotesingle}{\textquotesingle}}%
            \def\PYGZdq{\discretionary{}{\Wrappedafterbreak\char`\"}{\char`\"}}%
            \def\PYGZti{\discretionary{\char`\~}{\Wrappedafterbreak}{\char`\~}}%
        } %
        \newcommand*\Wrappedbreaksatpunct {%
            \lccode`\~`\.\lowercase{\def~}{\discretionary{\hbox{\char`\.}}{\Wrappedafterbreak}{\hbox{\char`\.}}}%
            \lccode`\~`\,\lowercase{\def~}{\discretionary{\hbox{\char`\,}}{\Wrappedafterbreak}{\hbox{\char`\,}}}%
            \lccode`\~`\;\lowercase{\def~}{\discretionary{\hbox{\char`\;}}{\Wrappedafterbreak}{\hbox{\char`\;}}}%
            \lccode`\~`\:\lowercase{\def~}{\discretionary{\hbox{\char`\:}}{\Wrappedafterbreak}{\hbox{\char`\:}}}%
            \lccode`\~`\?\lowercase{\def~}{\discretionary{\hbox{\char`\?}}{\Wrappedafterbreak}{\hbox{\char`\?}}}%
            \lccode`\~`\!\lowercase{\def~}{\discretionary{\hbox{\char`\!}}{\Wrappedafterbreak}{\hbox{\char`\!}}}%
            \lccode`\~`\/\lowercase{\def~}{\discretionary{\hbox{\char`\/}}{\Wrappedafterbreak}{\hbox{\char`\/}}}%
            \catcode`\.\active%
            \catcode`\,\active %
            \catcode`\;\active%
            \catcode`\:\active%
            \catcode`\?\active%
            \catcode`\!\active%
            \catcode`\/\active %
            \lccode`\~`\~ 	%
        }%
    \makeatother%

    \let\OriginalVerbatim=\Verbatim%
    \makeatletter%
    \renewcommand{\Verbatim}[1][1]{%
        \sbox\Wrappedcontinuationbox {\Wrappedcontinuationsymbol}%
        \sbox\Wrappedvisiblespacebox {\FV@SetupFont\Wrappedvisiblespace}%
        \def\FancyVerbFormatLine ##1{\hsize\linewidth%
            \vtop{\raggedright\hyphenpenalty\z@\exhyphenpenalty\z@%
                \doublehyphendemerits\z@\finalhyphendemerits\z@%
                \strut ##1\strut}%
        }%
        \def\FV@Space {%
            \nobreak\hskip\z@ plus\fontdimen3\font minus\fontdimen4\font%
            \discretionary{\copy\Wrappedvisiblespacebox}{\Wrappedafterbreak}%
            {\kern\fontdimen2\font}}%
        \Wrappedbreaksatspecials%
        \OriginalVerbatim[#1,codes*=\Wrappedbreaksatpunct]}%
    \makeatother%
    \definecolor{incolor}{HTML}{303F9F}%
    \definecolor{outcolor}{HTML}{D84315}%
    \definecolor{cellborder}{HTML}{CFCFCF}%
    \definecolor{cellbackground}{HTML}{F7F7F7}%
    \makeatletter%
    \newcommand{\boxspacing}{\kern\kvtcb@left@rule\kern\kvtcb@boxsep}%
    \makeatother%
    \hypersetup{%
      breaklinks=true,  
      colorlinks=true,%
      urlcolor=urlcolor,%
      linkcolor=linkcolor,%
      citecolor=citecolor,}%
    \title{Scalable Physics-Informed Neural Differential Equations and Data-Driven Algorithms for HVAC Systems}
    \vspace{15pt}%
    \author{\Large %
    {\sf Hanfeng Zhai$^{1}$}\thanks{Work done during internship at MERL; E-mail: \href{mailto:hzhai@stanford.edu}{\tt hzhai@stanford.edu}}, 
    {\sf Hongtao Qiao$^{2}$}\thanks{Corresponding author; E-mail: \href{mailto:qiao@merl.com}{\tt qiao@merl.com}}, 
    {\sf Hassan Mansour$^{2}$}, %
     {\sf Christopher Laughman$^{2}$}\vspace{10pt}%
\\
$^{1}$\emph{{Department of Mechanical Engineering}},\\
{\it Stanford University, Stanford, CA}%
\vspace{5pt}\\
$^{2}$\emph{{Mitsubishi Electric Research Laboratories}}, 
{\it Cambridge, MA}%
\\
\vspace{5pt}
}%

\usepackage{algorithm}%
\usepackage{algorithmicx}%
\usepackage{algpseudocode}%
\algrenewcommand\algorithmiccomment[1]{\hfill\textcolor{gray}{$\triangleright$~#1}}%
\usepackage{amsmath}%
\usepackage{amssymb}%
\usepackage{bm}%
\makeatletter
\@removefromreset{equation}{section}%
\@removefromreset{figure}{section}%
\@removefromreset{table}{section}%
\@removefromreset{algorithm}{section}%
\makeatother
\begin{document}%
\maketitle%
\begin{abstract}%
We present a scalable, data-driven simulation framework for large-scale heating, ventilation, and air conditioning (HVAC) systems that couples physics-informed neural ordinary differential equations (PINODEs) with differential-algebraic equation (DAE) solvers.
At the component level, we learn heat-exchanger dynamics using an implicit PINODE formulation that predicts conserved quantities (refrigerant mass $M_r$ and internal energy $E_\text{hx}$) as outputs, enabling physics-informed training via automatic differentiation of mass/energy balances.
Stable long-horizon prediction is achieved through gradient-stabilized latent evolution with gated architectures and layer normalization.
At the system level, we integrate learned components with DAE solvers (IDA and DASSL) that explicitly enforce junction constraints (pressure equilibrium and mass-flow consistency), and we use Bayesian optimization to tune solver parameters for accuracy--efficiency trade-offs.
To reduce residual system-level bias, we introduce a lightweight corrector network trained on short trajectory segments.
Across dual-compressor and scaled network studies, the proposed approach attains multi-fold speedups over high-fidelity simulation while keeping errors low (MAPE below a few percent) and scales to systems with up to 16 compressor--condenser pairs.

\vspace{25pt}

\end{abstract}%

\clearpage

\section*{Highlights}
\begin{itemize}
    \item Implicit physics-informed neural ODE formulation that treats conserved quantities as outputs, enabling physics-informed training via automatic differentiation.
    \item Gradient-stabilized latent evolution with gated architectures ensures stable training and long-horizon predictions for stiff thermo-fluid systems.
    \item Lightweight corrector network compensates for system-level errors, improving mass and energy prediction accuracy.
    \item DAE solvers (IDA and DASSL) with adaptive high-resolution stepping explicitly enforce algebraic constraints for pressure equilibrium and mass flow conservation.
    \item Scalable to systems with 16 compressor--condenser pairs, representing one of the largest data-driven HVAC surrogates reported.
    \item Achieves $4$--$9\times$ speedup with MAPE $< 2.5\%$ compared to high-fidelity simulators (best: 2.04\% MAPE, 58.91 s).
    \item Bayesian optimization automatically tunes solver parameters across multiple solver types for optimal accuracy--efficiency trade-offs.
\end{itemize}
\clearpage
\tableofcontents
\clearpage

\section*{Nomenclature}
\subsection*{State Variables}
\begin{tabular}{ll}
$M_r$ & Refrigerant mass within heat exchanger control volume (kg) \\
$E_{\text{hx}}$ & Internal energy within heat exchanger (J) \\
$p$ & Pressure (Pa) \\
$T$ & Temperature (K) \\
$h$ & Specific enthalpy (J/kg) \\
$\bm{y}$ & System state vector \\
$\bm{y}_d$ & Differential component of state vector \\
$\bm{y}_a$ & Algebraic component of state vector \\
$\dot{\bm{y}}$ & Time derivative of state vector \\
\end{tabular}
\subsection*{Rate Variables}
\begin{tabular}{ll}
$\dot{M}_r$ & Rate of change of refrigerant mass (kg/s) \\
$\dot{E}_{\text{hx}}$ & Rate of change of internal energy (J/s) \\
$\dot{m}$ & Mass flow rate (kg/s) \\
$\dot{m}_{r,\text{in}}$ & Inlet refrigerant mass flow rate (kg/s) \\
$\dot{m}_{r,\text{out}}$ & Outlet refrigerant mass flow rate (kg/s) \\
$\dot{Q}_a$ & Heat transfer rate to air stream (W) \\
\end{tabular}
\subsection*{DAE Solver Variables}
\begin{tabular}{ll}
$\mathbf{F}$ & DAE residual function \\
$\mathbf{G}$ & Nonlinear system function \\
$\mathbf{J}$ & Jacobian matrix \\
$h_n$ & Time step size at step $n$ \\
$\Delta t$ & Time step size \\
$h_\text{max}$ & Maximum time step size \\
$h_\text{min}$ & Minimum time step size \\
$\Delta t_\text{out}^\text{IDA}$ & IDA solver output interval \\
$\Delta t_\text{out,min}^\text{DASSL}$ & DASSL minimum output interval \\
$N_\text{max}^\text{DASSL}$ & DASSL maximum number of internal steps \\
$\Delta t_\text{target}$ & Target output interval for adaptive stepping \\
$t_n$ & Time at step $n$ \\
$k$ & BDF order \\
$\alpha_{i,k}$ & BDF coefficients (subscript $i$ for coefficient index, subscript $k$ for order) \\
$\text{ATOL}$ & Absolute tolerance \\
$\text{RTOL}$ & Relative tolerance \\
$\epsilon_{\Delta t}$ & Time step tolerance \\
$\epsilon_\text{soln}$ & Solution tolerance for root-finding \\
$\tau$ & Control jump detection threshold \\
$n_\text{pre}$ & Number of steps before control jump (high-res) \\
$n_\text{post}$ & Number of steps after control jump (high-res) \\
\end{tabular}
\subsection*{Neural Network Variables}
\begin{tabular}{ll}
$\mathbf{h}(t)$ & Latent state vector in neural ODE \\
$\mathbf{h}_0$ & Initial latent state \\
$\bm{X}_\text{enc}$ & Encoder input sequence \\
$\mathbf{S}_\text{enc}$ & Encoder state sequence \\
$\mathbf{S}_\text{dec}$ & Decoder state sequence \\
$\bm{x}(t)$ & Time-dependent feature vector \\
$\mathbf{s}(t)$ & Time-dependent state vector \\
$\hat{\bm{y}}$ & Predicted output vector \\
$\mathbf{W}$ & Neural network weight matrices \\
$\boldsymbol{\theta}$ & Neural network parameters \\
$d_\text{latent}$ & Latent dimension \\
$d_x$ & Feature dimension \\
$d_s$ & State dimension \\
$T_\text{enc}$ & Encoder sequence length \\
$T_\text{dec}$ & Decoder sequence length \\
$\Phi_\text{enc}$ & Encoder function (GRU) \\
$\psi$ & Decoder function (GRU) \\
$\zeta$ & Latent projection network \\
$f_\theta$ & Neural ODE dynamics function \\
$\phi_\text{corr}$ & Corrector network function \\
$\mathbf{z}_\text{in}$ & Corrector network input vector \\
$\boldsymbol{\phi}_\text{corr}$ & Corrector network output (correction term) \\
$\boldsymbol{\phi}_\text{corr}^\text{raw}$ & Raw correction from network \\
$\boldsymbol{\phi}_\text{corr}^\text{GP}$ & GP-smoothed correction \\
$\boldsymbol{\phi}_\text{corr}^\text{smooth}$ & Final smoothed correction \\
$\hat{\mathbf{m}}_\text{pred}$ & Normalized predicted mass/energy values \\
$\hat{\mathbf{m}}_\text{bench}$ & Normalized benchmark mass/energy values \\
$\hat{\mathbf{m}}_\text{corr}$ & Normalized corrected mass/energy values \\
$d_\text{in}$ & Corrector network input dimension \\
\end{tabular}
\subsection*{Loss Functions}
\begin{tabular}{ll}
$\mathcal{L}_\text{total}$ & Total training loss \\
$\mathcal{L}_\text{data}$ & Data fidelity loss \\
$\mathcal{L}_\text{phys}$ & Physics-informed loss \\
$\mathcal{L}_\text{cons}$ & Conservation loss \\
$\mathcal{L}_\text{corr}$ & Corrector network loss \\
$\lambda_\text{phys}$ & Physics loss weight \\
$\lambda_\text{cons}$ & Conservation loss weight \\
$N$ & Batch size \\
\end{tabular}
\subsection*{System Parameters}
\begin{tabular}{ll}
$n_c$ & Number of compressors (or compressor--condenser pairs) \\
$n_v$ & Number of valves (or valve--evaporator pairs) \\
$n_{\text{cond}}$ & Number of condensers \\
$n_{\text{evap}}$ & Number of evaporators \\
$n_p$ & Number of pressure variables (junction pressures) \\
$n$ & Total system dimension \\
$N_\text{steps}$ & Number of simulation time steps \\
$p_\text{liq}$ & Liquid manifold pressure node \\
$p_\text{suct}$ & Suction manifold pressure node \\
$\mathcal{G}(n_c, n_v)$ & System topology graph with $n_c$ compressors and $n_v$ evaporators \\
\end{tabular}%
\subsection*{Mathematical Symbols}
\begin{tabular}{ll}
$\sigma$ & Sigmoid activation function \\
$\odot$ & Element-wise multiplication \\
$\mathbb{R}$ & Set of real numbers \\
$\|\cdot\|$ & Euclidean norm \\
$\partial$ & Partial derivative \\
$\nabla$ & Gradient operator \\
$\text{MSE}$ & Mean squared error \\
$\text{MAPE}$ & Mean absolute percentage error \\
$\arg\min$ & Argument of the minimum \\
$\mathcal{D}$ & Dataset \\
$\mathcal{M}$ & Model \\
$\mathcal{X}$ & Parameter space \\
\end{tabular}
\subsection*{Abbreviations}
\begin{tabular}{ll}
PINODE & Physics-Informed Neural ODE \\
DAE & Differential-Algebraic Equation \\
ODE & Ordinary Differential Equation \\
DASSL & Differential-Algebraic System Solver \\
BDF & Backward Differentiation Formula \\
GRU & Gated Recurrent Unit \\
RK4 & Fourth-order Runge-Kutta method \\
HVAC & Heating, Ventilation, and Air Conditioning \\
GP & Gaussian Process \\
EI & Expected Improvement \\
\end{tabular}%
%


\clearpage

\section{Introduction\label{intro}}%

Heating, ventilation, and air conditioning (HVAC) systems account for a substantial portion of global energy consumption, with residential and commercial buildings responsible for approximately 40\% of total energy usage~\cite{eia2021buildings}.
As the demand for energy-efficient building operations grows, the need for accurate, real-time modeling and control of HVAC systems has become increasingly critical~\cite{Afram2014_hvac_review}.
Large-scale HVAC networks, comprising multiple compressors, condensers, evaporators, and valves, exhibit complex thermo-fluid dynamics governed by nonlinear differential-algebraic equations (DAEs) that couple mass, energy, and momentum conservation laws~\cite{Chakrabarty2021_HVAC_DAE_BO}.
Traditional approaches to HVAC modeling and simulation face a fundamental trade-off: high-fidelity physics-based models implemented in commercial software (e.g., Modelica~\cite{modelica2005specification}, EnergyPlus~\cite{crawley2008energyplus}) achieve high accuracy but require computationally expensive numerical solvers that preclude real-time applications, while simplified reduced-order models sacrifice physical consistency for computational efficiency.

Current physics-based simulators, despite their accuracy, suffer from significant computational bottlenecks that limit their practical deployment.
High-fidelity simulations of large-scale HVAC systems are computationally intensive~\cite{qiao2015transient}, make it challenging for real-time control, optimization, or co-simulation with building energy management systems~\cite{chen2018building}.
The computational burden stems from the need to solve stiff, nonlinear DAEs with adaptive time-stepping, handle complex thermodynamic property calculations, and resolve fast transients in multi-component networks \cite{Chakrabarty2021_HVAC_DAE_BO, Sahlin2004_IDA_DAE, Anantharaman2023PhD_MIT}.
This research gap between accuracy and computational efficiency has motivated the development of data-driven surrogates that can approximate the behavior of physics-based simulators at a fraction of the computational cost~\cite{Ma2024, Chakrabarty2021_hvac_bo, Zhao2023, Taboga2024, Chen2021_scale_efficiency}.

To bridge this gap, we propose a hybrid modeling framework that combines the expressiveness of neural networks with the physical consistency of conservation laws, enabling both accurate predictions and efficient computation.
Our approach integrates physics-informed neural ordinary differential equations (PINODEs)~\cite{Sholokhov2023} for component-level dynamics with specialized DAE solvers~\cite{IDA_SUNDIALS, petzold1982dassl} for system-level constraint satisfaction.
This dual-level strategy allows us to leverage the flexibility of machine learning for capturing complex, data-driven patterns while explicitly enforcing fundamental physical principles, such as mass and energy conservation, that are essential for thermodynamic validity~\cite{Fleming2013_energy_conserv_hvac, Mamadou2025}.

Existing machine learning surrogates for HVAC systems face several critical limitations that compromise their reliability and applicability.
Conventional neural network approaches, such as feedforward networks or recurrent architectures (LSTMs, GRUs), could produce predictions that violate conservation laws, leading to unphysical results that accumulate errors over long prediction horizons~\cite{raissi2019physics}.
Similarly, standard neural ODE formulations~\cite{chen2018neural} struggle with numerical stability during training and inference, particularly when dealing with stiff dynamics or long time horizons~\cite{de2019gru_ode, thummerer2023eigen_stable}.
On the system-level side, most existing solvers for HVAC networks either simplify the algebraic constraints to make the system easier to solve (losing accuracy) or employ generic ODE integrators that cannot properly handle the differential-algebraic structure, which may lead to constraint violations and pressure imbalances~\cite{Ma2024}.
Our framework addresses these limitations through gradient-stabilized latent evolution in the neural ODE component and explicit algebraic constraint enforcement in the DAE solver, ensuring both stability and physical consistency.
To solve these challenges, we draw upon recent advances in continuous-time neural models and differential-algebraic equation solvers.
Neural ODEs~\cite{chen2018neural} and physics-informed neural networks~\cite{raissi2019physics, meng2020ppinn, zhai2022pinn, lu2021deepxde} have demonstrated how to incorporate physical laws, though existing work often combines these ideas superficially without deeply embedding conservation laws into the latent dynamics or addressing stability issues in stiff systems.
For the system-level solver, DASSL (Differential-Algebraic System Solver)~\cite{petzold1982dassl} provides a robust foundation for handling index-1 DAEs, but its application to large-scale HVAC networks requires careful parameter tuning and adaptive time-stepping strategies.
To ensure fair and systematic comparison across different solver types (algebraic, DAE-IDA, and DAE-DASSL), we employ Bayesian optimization~\cite{frazier2018bayesian, zhai2022computational, Zhao2023, Chakrabarty2021_HVAC_DAE_BO} to systematically explore the parameter space and characterize each solver's performance, enabling objective evaluation of accuracy and computational efficiency.

The key novelty of our proposed methodology consists of four core innovations:
\begin{itemize}
    \item \textbf{Implicit PINODE formulation with conservation-aware states:} Conserved quantities (refrigerant mass $M_r$ and internal energy $E_\text{hx}$) are treated as model outputs rather than prescribed inputs, enabling direct use of conservation structure in learning and downstream system simulation.
    \item \textbf{Stable long-horizon latent dynamics:} A gradient-stabilized evolution scheme with gated architectures and layer normalization~\cite{ba2016layer} improves stability for stiff thermo-fluid dynamics and supports accurate long-horizon prediction.
    \item \textbf{DAE-based system integration with explicit constraint enforcement:} A specialized system-level DAE formulation explicitly enforces junction constraints (e.g., pressure equilibrium and mass flow consistency), improving thermodynamic consistency across coupled components.
    \item \textbf{Automated solver tuning and scalability assessment:} Bayesian optimization is used to tune solver parameters for fair accuracy--efficiency comparison, and the framework is demonstrated on large-scale HVAC configurations.
\end{itemize}
These innovations enable scalable and thermodynamically consistent modeling of large HVAC configurations, while achieving $4$--$9\times$ computational speedup with MAPE $< 2.5\%$ relative to high-fidelity simulators.

The remainder of this paper is organized as follows.
Section~\ref{sec:pinode} presents the physics-informed neural ODE framework for heat exchanger modeling, including the encoder-decoder architecture, latent dynamics formulation, and physics-informed training strategy.
Section~\ref{sec:system_solver} describes the DAE solver for large-scale HVAC systems, detailing the differential--algebraic formulation, DASSL integration, and Bayesian optimization for parameter tuning.
Section~\ref{sec:results} presents experimental results on real HVAC datasets, demonstrating accuracy, efficiency, and scalability.
Section~\ref{sec:scalability} discusses the scalability analysis and computational performance.
Finally, Section~\ref{sec:conclusions} summarizes the contributions and discusses future research directions.
\clearpage

\section{Methodologies\label{sec:method}}
\subsection{Physics-Informed Neural ODE\label{sec:pinode}}

We develop a physics-informed neural ordinary differential equation (PINODE) framework for modeling heat exchanger dynamics in HVAC systems.
The approach combines the expressiveness of neural networks with the physical consistency of conservation laws, enabling accurate long-horizon predictions while maintaining thermodynamic validity.

\subsubsection{Implicit Formulation}

A key design decision in our framework is the treatment of conserved quantities. Specifically, the refrigerant mass $M_r$ and internal energy $E_\text{hx}$ are treated as model outputs rather than known inputs.
This implicit formulation contrasts with the explicit approach in \cite{Ma2024}, where mass and energy are provided as inputs to the heat exchanger model, following a discrete-time dynamical system form with fixed time steps.

The heat exchanger model maps an 8-dimensional input vector $\bm{x}_\text{in} \in \mathbb{R}^8$ to a 9-dimensional output vector $\hat{\bm{y}} \in \mathbb{R}^9$:
\begin{equation}
\hat{\bm{y}} = \mathcal{M}_\text{PINODE}(\bm{x}_\text{in}, \mathbf{s}_\text{in}),
\end{equation}
where the input vector $\bm{x}_\text{in}$ contains the boundary conditions:
\begin{equation}
\bm{x}_\text{in} = \left[T_{a,\text{in}}, \phi_{a,\text{in}}, \dot{m}_a, P_\text{amb}, \dot{m}_{r,\text{in}}, h_{r,\text{in}}, h_{r,\text{out}}, P_{r,\text{out}}\right]^\top,
\end{equation}
comprising air-side properties (inlet temperature $T_{a,\text{in}}$, humidity ratio $\phi_{a,\text{in}}$, mass flow rate $\dot{m}_a$, ambient pressure $P_\text{amb}$) and refrigerant-side properties (inlet mass flow rate $\dot{m}_{r,\text{in}}$, inlet enthalpy $h_{r,\text{in}}$, outlet enthalpy $h_{r,\text{out}}$, outlet pressure $P_{r,\text{out}}$).
The output vector $\hat{\bm{y}}$ includes both observable quantities and the conserved state variables:
\begin{equation}
\hat{\bm{y}} = \left[p_1, p_N, h_1, h_N, T_{a,\text{out}}, \dot{Q}_a, M_r, E_\text{hx}, \dot{Q}_\text{lat}\right]^\top,
\end{equation}
where $p_1$ and $p_N$ are refrigerant pressures at the first and last nodes, $h_1$ and $h_N$ are corresponding enthalpies, $T_{a,\text{out}}$ is the air outlet temperature, $\dot{Q}_a$ is the heat transfer rate to air, $M_r$ and $E_\text{hx}$ are the conserved quantities (refrigerant mass and internal energy), and $\dot{Q}_\text{lat}$ is the latent heat transfer rate.

The key distinction from the explicit formulation in \cite{Ma2024} lies in the treatment of $M_r$ and $E_\text{hx}$.
In the explicit approach, these quantities are provided as inputs: $\bm{x}_\text{in}^\text{explicit} = [\bm{x}_\text{in}, M_r, E_\text{hx}]^\top \in \mathbb{R}^{10}$, and the model predicts only the observable outputs.
In contrast, our implicit formulation treats $M_r$ and $E_\text{hx}$ as outputs, enabling direct computation of their time derivatives through automatic differentiation:
\begin{equation}
\dot{M}_r^{\text{pred}} = \frac{d}{dt} \hat{M}_r, \quad \dot{E}_\text{hx}^{\text{pred}} = \frac{d}{dt} \hat{E}_\text{hx},
\end{equation}
where the derivatives are computed with respect to time through the neural network's computational graph.
This capability enables the incorporation of gradient information into physics-informed loss terms that enforce conservation laws during training, as the model must predict rates of change that are consistent with mass and energy balance:
\begin{equation}
\dot{M}_r^{\text{true}} = \dot{m}_{r,\text{in}} - \dot{m}_{r,\text{out}}, \quad \dot{E}_\text{hx}^{\text{true}} = \dot{m}_{r,\text{in}} h_{r,\text{in}} - \dot{m}_{r,\text{out}} h_{r,\text{out}} - \dot{Q}_a.
\end{equation}
This physics-informed training approach is essential for learning dynamics that respect thermodynamic constraints, ensuring that the predicted rates $\dot{M}_r^{\text{pred}}$ and $\dot{E}_\text{hx}^{\text{pred}}$ align with the true conservation-based rates $\dot{M}_r^{\text{true}}$ and $\dot{E}_\text{hx}^{\text{true}}$.

\begin{figure}[htbp]
    \centering
    \includegraphics[width=\linewidth]{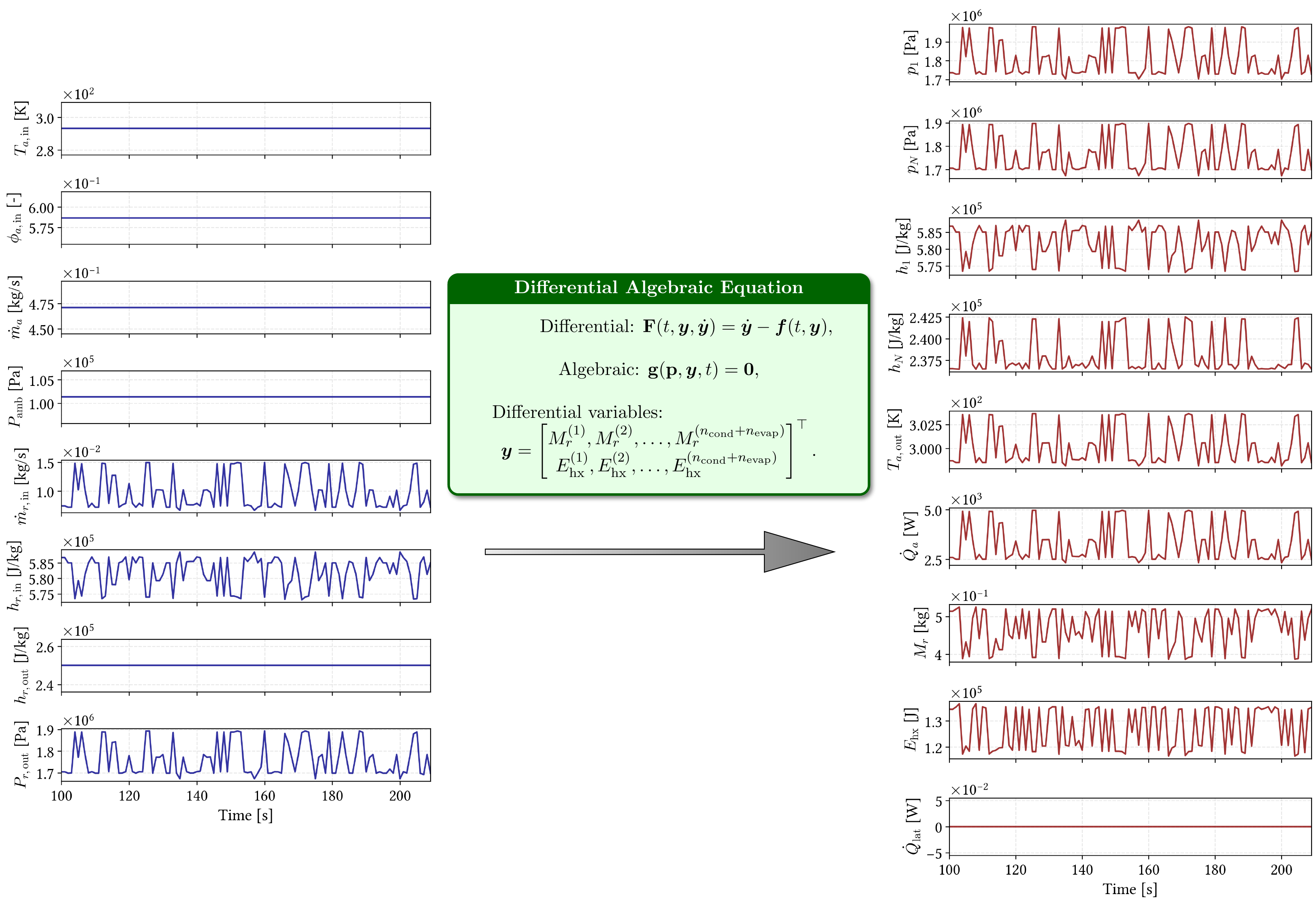}
    \caption{Heat exchanger model inputs and outputs for the implicit PINODE formulation. The top panel shows the 8 input variables (air and refrigerant boundary conditions), while the bottom panel shows the 9 output variables, including the conserved quantities $M_r$ and $E_\text{hx}$. The figure displays a representative 100-step segment (steps 100--210) of a simulation; in practice, the model can simulate thousands of time steps.}
    \label{fig:hex_schematic}
\end{figure}

\begin{figure}[htbp]
    \centering
    \begin{tikzpicture}[x=1cm,y=1cm, font=\small]
        \draw[rounded corners=2pt, thick, fill=gray!10] (0,0) rectangle (6,2);
        \node at (3,1) {Heat exchanger (control volume)};

        \draw[-{Stealth[length=2.2mm]}, thick, blue!70!black] (-1,0.55) -- (0,0.55);
        \draw[-{Stealth[length=2.2mm]}, thick, blue!70!black] (6,0.55) -- (7,0.55);
        \node[blue!70!black, anchor=east] at (-1,0.55) {Refrigerant in};
        \node[blue!70!black, anchor=west] at (7,0.55) {Refrigerant out};

        \draw[-{Stealth[length=2.2mm]}, thick, red!70!black] (7,1.45) -- (6,1.45);
        \draw[-{Stealth[length=2.2mm]}, thick, red!70!black] (0,1.45) -- (-1,1.45);
        \node[red!70!black, anchor=west] at (7,1.45) {Air in};
        \node[red!70!black, anchor=east] at (-1,1.45) {Air out};

        \node[anchor=south] at (3,2.05) {States advanced by solver: $M_r$, $E_{\text{hx}}$};
    \end{tikzpicture}
    \caption{Simplified heat exchanger schematic (control volume view) illustrating the coupling between air-side and refrigerant-side flows. The conserved quantities $M_r$ and $E_{\text{hx}}$ are the differential states advanced by the system-level solver.}
    \label{fig:hex_physical_schematic}
\end{figure}
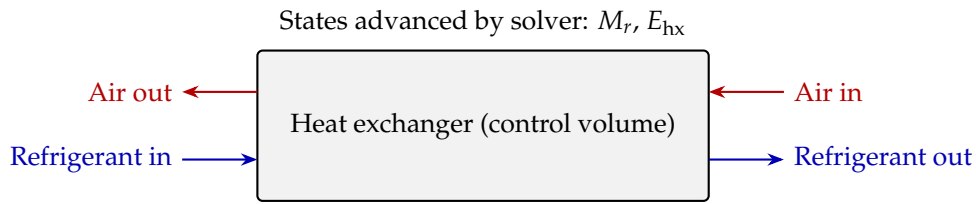

\begin{sloppypar}
Figure~\ref{fig:hex_schematic} illustrates the input--output structure of the implicit PINODE formulation for heat exchanger modeling.
The model receives 8 input variables that characterize the boundary conditions: air-side properties (inlet temperature, humidity ratio, mass flow rate, ambient pressure) and refrigerant-side properties (inlet and outlet mass flow rates, enthalpies, and outlet pressure).
The PINODE model processes these inputs through its encoder--neural ODE--decoder architecture to predict 9 output variables, including the conserved quantities $M_r$ and $E_\text{hx}$ that are treated as differential states in the system-level DAE formulation.
The figure shows a representative 100-step segment (steps 100--210) of a simulation trajectory; in practice, the model can simulate thousands of time steps to capture long-horizon system dynamics.
\end{sloppypar}

The implicit formulation is fundamentally more physically meaningful: in real-world HVAC simulations, the mass and energy within each heat exchanger are unknown state variables that emerge from solving the system dynamics, not known inputs that can be directly measured or specified.
Although this introduces training challenges due to potential error accumulation, we address these through careful architecture design, including gradient-stabilized latent evolution and regularization strategies detailed in the following sections.

\subsubsection{Model Framework}

\begin{figure}[htbp]
\centering%
\includegraphics[width=0.95\linewidth]{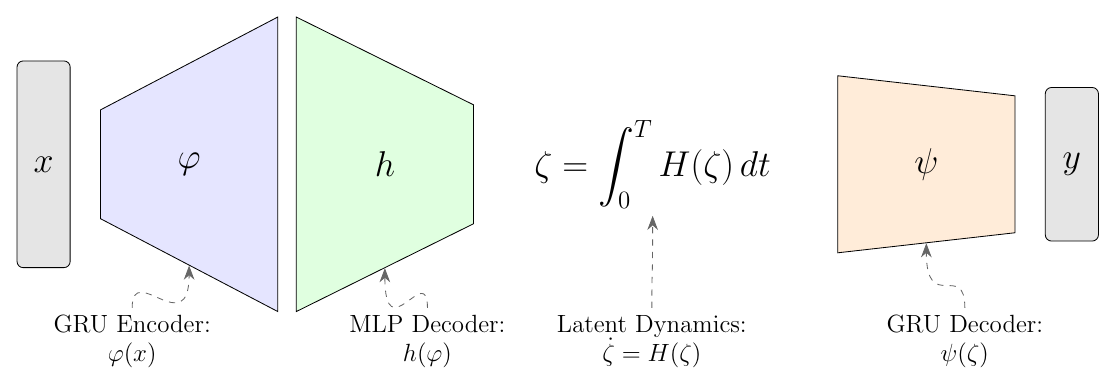}
\caption{Schematic of the physics-informed neural ODE (PINODE) architecture for heat exchanger modeling.
The model encodes input sequences into a latent space, evolves the latent dynamics using a neural ODE, and decodes back to observable states.
Physics constraints are enforced through a physics-informed loss term that penalizes violations of mass and energy conservation.}
\label{fig:pinode_schematic}
\end{figure}

The PINODE architecture consists of three main components: an encoder, a neural ODE, and a decoder, as illustrated in Figure~\ref{fig:pinode_schematic}.
The encoder processes historical input sequences to extract a compact latent representation, which is then evolved forward in time through a continuous-time neural ODE that captures the underlying dynamics.
The decoder maps the evolved latent trajectory back to the observable output space, enabling predictions of heat exchanger states and thermodynamic properties.
Given encoder sequences $\bm{X}_\text{enc} \in \mathbb{R}^{T_\text{enc} \times d_x}$ and decoder sequences $\mathbf{S}_\text{dec} \in \mathbb{R}^{T_\text{dec} \times d_s}$, where $d_x$ and $d_s$ denote feature and state dimensions respectively, the model operates as follows.

\begin{algorithm}[htbp]
\caption{PINODE Forward Pass for Heat Exchanger Prediction}
\label{alg:pinode_forward}
\begin{algorithmic}[1]
\Require Encoder sequences $\bm{X}_\text{enc}$, $\mathbf{S}_\text{enc}$; decoder sequences $\bm{X}_\text{dec}$, $\mathbf{S}_\text{dec}$; time step $\Delta t$
\Ensure Predicted outputs $\hat{\bm{y}}$ for each decoder time step
\State \textbf{Step 1: Encoding} - Extract latent representation from input history
\State $\mathbf{z}_T \gets \Phi_\text{enc}(\bm{X}_\text{enc}, \mathbf{S}_\text{enc})$ \Comment{GRU encoder processes input sequence}
\State $\bm{\zeta}_0 \gets \zeta(\mathbf{z}_T)$ \Comment{Project to latent space $\mathbb{R}^{d_\text{latent}}$}
\State \textbf{Step 2: Latent Evolution} - Integrate neural ODE over decoder horizon
\For{each decoder time step $t = 1, 2, \ldots, T_\text{dec}$}
    \State Extract feature and state vectors: $\bm{x}_t \gets \bm{X}_\text{dec}[t]$, $\mathbf{s}_t \gets \mathbf{S}_\text{dec}[t]$
    \State Compute latent derivative: $\frac{d\bm{\zeta}}{dt} \gets f_\theta(\bm{\zeta}_{t-1}, \bm{x}_t, \mathbf{s}_t)$ \Comment{Neural ODE with gated architecture}
    \State Integrate latent trajectory: $\bm{\zeta}_t \gets \text{RK4}(f_\theta, \bm{\zeta}_{t-1}, \bm{x}_t, \mathbf{s}_t, \Delta t)$ \Comment{4th-order Runge-Kutta}
\EndFor
\State \textbf{Step 3: Decoding} - Map evolved latent states to predictions
\For{each decoder time step $t = 1, 2, \ldots, T_\text{dec}$}
    \State $\hat{\bm{y}}_t \gets \psi(\bm{\zeta}_t)$ \Comment{GRU decoder maps latent to observable space}
\EndFor
\State \textbf{Step 4: Physics Constraints} - Compute derivatives for physics loss
\State $\dot{M}_r^{\text{pred}}, \dot{E}_\text{hx}^{\text{pred}} \gets \frac{d}{dt}[\hat{M}_r, \hat{E}_\text{hx}]$ \Comment{Automatic differentiation}
\State \Return Predicted outputs $\hat{\bm{y}} = [\hat{\bm{y}}_1, \ldots, \hat{\bm{y}}_{T_\text{dec}}]$ and derivatives
\end{algorithmic}
\end{algorithm}

\textbf{Encoder:} A GRU encoder $\Phi_\text{enc}$ processes the input sequence to extract a latent representation that captures the temporal dependencies in the input history.
The encoder takes as input the feature sequence $\bm{X}_\text{enc}$ and state sequence $\mathbf{S}_\text{enc}$, processing them through recurrent layers to produce a summary representation:
\begin{equation}
    \mathbf{z}_T = \Phi_\text{enc}(\bm{X}_\text{enc}, \mathbf{S}_\text{enc}),
\end{equation}
where $\mathbf{S}_\text{enc}$ represents the encoder state sequence and $\mathbf{z}_T$ is the final hidden state after processing all $T_\text{enc}$ time steps.
The encoder output is then projected into a lower-dimensional latent space via a feedforward network $\zeta$ to initialize the latent trajectory:
\begin{equation}
    \bm{\zeta}_0 = \zeta(\mathbf{z}_T) \in \mathbb{R}^{d_\text{latent}},
\end{equation}
where $d_\text{latent}$ is the latent dimension (typically $d_\text{latent} = 8$), and $\bm{\zeta}_0$ serves as the initial condition for the neural ODE integration.

\textbf{Neural ODE:} The latent trajectory $\bm{\zeta}(t)$ evolves continuously in time according to a neural ODE, which learns the underlying dynamics of the heat exchanger system in a low-dimensional latent space.
The evolution is governed by:
\begin{equation}
    \frac{d\bm{\zeta}}{dt} = f_\theta(\bm{\zeta}(t), \bm{x}(t), \mathbf{s}(t)),
\end{equation}
where $f_\theta$ is a neural network parameterized by $\theta$ that defines the vector field in the latent space, and $\bm{x}(t)$ and $\mathbf{s}(t)$ are the time-dependent feature and state vectors at the decoder input that provide external forcing and boundary conditions.
The neural ODE uses a gated architecture inspired by GRU cells to ensure stable gradient flow and prevent numerical instabilities during long-horizon integration:
\begin{equation}
    \begin{aligned}
        \mathbf{r}_t &= \sigma(\mathbf{W}_r [\bm{x}_t, \mathbf{s}_t] + \mathbf{U}_r \bm{\zeta}_t),\\
        \mathbf{z}_t &= \sigma(\mathbf{W}_z [\bm{x}_t, \mathbf{s}_t] + \mathbf{U}_z \bm{\zeta}_t),\\
        \tilde{\bm{\zeta}}_t &= \tanh(\mathbf{W}_h [\bm{x}_t, \mathbf{s}_t] + \mathbf{U}_h (\mathbf{r}_t \odot \bm{\zeta}_t)),\\
        \frac{d\bm{\zeta}}{dt} &= (1 - \mathbf{z}_t) \odot (\tilde{\bm{\zeta}}_t - \bm{\zeta}_t),
    \end{aligned}
\end{equation}
where $\sigma$ denotes the sigmoid function, $\odot$ is element-wise multiplication, and $\mathbf{W}_r, \mathbf{W}_z, \mathbf{W}_h, \mathbf{U}_r, \mathbf{U}_z, \mathbf{U}_h$ are learnable weight matrices.
The reset gate $\mathbf{r}_t$ controls how much of the previous latent state is retained, while the update gate $\mathbf{z}_t$ balances between the previous state and the candidate update $\tilde{\bm{\zeta}}_t$.
This gated formulation mitigates vanishing and exploding gradients by constraining the rate of change in the latent space, enabling stable training and accurate long-horizon predictions.

The latent trajectory is integrated forward in time using a fourth-order Runge-Kutta (RK4) scheme, which provides high accuracy while maintaining computational efficiency:
\begin{equation}
    \bm{\zeta}_{t+\Delta t} = \text{RK4}(f_\theta, \bm{\zeta}_t, \bm{x}_t, \mathbf{s}_t, \Delta t),
\end{equation}
where $\Delta t$ is the time step size.
The RK4 method evaluates the vector field $f_\theta$ at four intermediate points within each time step, combining these evaluations to produce a fourth-order accurate approximation of the latent state at the next time point.
Our work is inspired by \cite{Sholokhov2023}.

\textbf{Decoder:} A GRU decoder $\psi$ maps the evolved latent trajectory $\bm{\zeta}(t)$ back to the observable output space, transforming the low-dimensional latent representation into predictions of heat exchanger states and thermodynamic properties:
\begin{equation}
    \hat{\bm{y}}= \psi(\bm{\zeta}(t)),
\end{equation}
where $\hat{\bm{y}}$ contains the predicted outputs including pressures, enthalpies, temperatures, heat transfer rates, and the conserved quantities (mass and energy).

To improve training stability, we employ layer normalization~\cite{ba2016layer} at multiple stages: after the encoder, in the latent space, and before the final output layer.
Dropout regularization is applied to prevent overfitting during training.

\subsubsection{Physics-Informed Loss}

The training objective combines data fidelity with physics constraints to ensure the learned dynamics respect thermodynamic conservation laws.
For a heat exchanger, the key conserved quantities are the refrigerant mass $M_r$ and internal energy $E_\text{hx}$ within the control volume.

The total loss function is defined as:
\begin{equation}
    \begin{aligned}
        \mathcal{L}_\text{total} &= \mathcal{L}_\text{data} + \lambda_\text{phys} \mathcal{L}_\text{phys} + \lambda_\text{cons} \mathcal{L}_\text{cons},\\
        \mathcal{L}_\text{data} &= \frac{1}{N} \sum_{i=1}^{N} \|\hat{\bm{y}}_i - \bm{y}_i\|^2,\\
        \mathcal{L}_\text{phys} &= \frac{1}{N} \sum_{i=1}^{N} \left[ \|\dot{M}_r^{\text{pred}} - \dot{M}_r^{\text{true}}\|^2 + \|\dot{E}_\text{hx}^{\text{pred}} - \dot{E}_\text{hx}^{\text{true}}\|^2 \right],\\
        \mathcal{L}_\text{cons} &= \frac{1}{N} \sum_{i=1}^{N} \left[ \|\hat{M}_r - M_r^{\text{true}}\|^2 + \|\hat{E}_\text{hx} - E_\text{hx}^{\text{true}}\|^2 \right],
    \end{aligned}
\end{equation}
where $\lambda_\text{phys}$ and $\lambda_\text{cons}$ are weighting hyperparameters (typically set to 0.5), $N$ is the batch size, and $\hat{\bm{y}}_i$ and $\bm{y}_i$ are the predicted and ground-truth state vectors, respectively.

The physics loss $\mathcal{L}_\text{phys}$ enforces consistency between the model's predicted rates of change and the true rates computed from conservation laws.
For a heat exchanger control volume, the mass and energy rates are:
\begin{equation}
    \begin{aligned}
        \dot{M}_r^{\text{true}} &= \dot{m}_{r,\text{in}} - \dot{m}_{r,\text{out}},\\
        \dot{E}_\text{hx}^{\text{true}} &= \dot{m}_{r,\text{in}} h_{r,\text{in}} - \dot{m}_{r,\text{out}} h_{r,\text{out}} + \dot{Q}_a,
    \end{aligned}
\end{equation}
where $\dot{m}_{r,\text{in}}$ and $\dot{m}_{r,\text{out}}$ are the inlet and outlet mass flow rates, $h_{r,\text{in}}$ and $h_{r,\text{out}}$ are the corresponding specific enthalpies, and $\dot{Q}_a$ is the heat transfer rate to the air stream.

The predicted rates $\dot{M}_r^{\text{pred}}$ and $\dot{E}_\text{hx}^{\text{pred}}$ are computed via automatic differentiation of the model output with respect to time, leveraging the continuous-time formulation of the neural ODE:
\begin{equation}
    \frac{d\hat{\bm{y}}}{dt} = \frac{\partial \psi}{\partial \bm{\zeta}} \frac{d\bm{\zeta}}{dt},
\end{equation}
where the derivatives are evaluated using the chain rule through the decoder and neural ODE.
This automatic differentiation capability is a key advantage of the continuous-time formulation, as it enables direct computation of time derivatives without finite difference approximations, ensuring smooth and accurate gradient information for the physics-informed loss terms.

Since $M_r$ and $E_\text{hx}$ are treated as algebraic variables in the implicit formulation, the conservation loss $\mathcal{L}_\text{cons}$ provides additional supervision to ensure these quantities match the ground truth.
\subsubsection{Training Algorithms}

The training procedure alternates between validation and training phases within each epoch, as detailed in Algorithm~\ref{alg:training}.
During the validation phase, the model is evaluated on held-out data to monitor generalization performance and adjust the learning rate.
The training phase performs gradient-based optimization with physics-informed loss terms.

Key implementation details:
\begin{itemize}
    \item \textbf{Optimizer:} Adam optimizer with an initial learning rate of $10^{-3}$, reduced by a factor of 0.5 when validation loss plateaus (patience of 25 epochs).
    \item \textbf{Gradient clipping:} Gradients are clipped to a maximum norm of 1.0 to prevent instability during training.
    \item \textbf{Physics weight scheduling:} The physics loss weight $\lambda_\text{phys}$ is set to 0.5 throughout training, balancing data fidelity with physical consistency.
    \item \textbf{Batch processing:} Sequences are processed in batches, with encoder sequences of length $T_\text{enc}$ and decoder sequences of length $T_\text{dec}$.
\end{itemize}

The algorithm uses an implicit problem formulation, where algebraic constraints are enforced through the conservation loss term, which penalizes deviations in the conserved quantities $M_r$ and $E_\text{hx}$.

\begin{algorithm}[htbp]%
\caption{Physics-Informed Neural ODE Training for Heat Exchanger}
\label{alg:training}
\begin{algorithmic}[1]%
\Require Training data $\mathcal{D}_{\text{train}}$, validation data $\mathcal{D}_{\text{val}}$, model $\mathcal{M}$ (PINODE)%
\Ensure Trained model $\mathcal{M}^*$ with optimized parameters%
\State Initialize $\mathcal{M}$ with random weights $\mathbf{W}$, optimizer (Adam), learning rate scheduler%
\State Set $\lambda_{\text{phys}} = 0.5$, $\lambda_{\text{cons}} = 0.5$%

\For{epoch $e = 1, 2, \ldots, E$}
    \State \textbf{Validation:} Set model to evaluation mode%
    \For{each batch $(X_{\text{enc}}, S_{\text{enc}}, X_{\text{dec}}, S_{\text{dec}}, \Delta t, Y) \in \mathcal{D}_{\text{val}}$}
        \State $\hat{S} \gets \mathcal{M}(X_{\text{enc}}, S_{\text{enc}}, X_{\text{dec}}, S_{\text{dec}}, \Delta t)$%
        \State $\mathcal{L}_{\text{data}} \gets \text{MSE}(\hat{S}, Y)$%
        \State Compute $\dot{M}_r^{\text{true}}, \dot{E}_{\text{hx}}^{\text{true}}$ from physics; extract $\dot{M}_r^{\text{pred}}, \dot{E}_{\text{hx}}^{\text{pred}}$ from model%
        \State $\mathcal{L}_{\text{phys}} \gets \text{MSE}(\dot{M}_r^{\text{pred}}, \dot{M}_r^{\text{true}}) + \text{MSE}(\dot{E}_{\text{hx}}^{\text{pred}}, \dot{E}_{\text{hx}}^{\text{true}})$%
        \State $\mathcal{L}_{\text{cons}} \gets \text{MSE}(\hat{S}[M_r], Y[M_r]) + \text{MSE}(\hat{S}[E_{\text{hx}}], Y[E_{\text{hx}}])$%
        \State $\mathcal{L}_{\text{val}} \gets \mathcal{L}_{\text{data}} + \lambda_{\text{phys}} \cdot \mathcal{L}_{\text{phys}} + \lambda_{\text{cons}} \cdot \mathcal{L}_{\text{cons}}$%
    \EndFor%
    \State Update learning rate; if $\mathcal{L}_{\text{val}} < \mathcal{L}_{\text{val}}^{*}$, save $\mathcal{M}^* \gets \mathcal{M}$%
    \State \textbf{Training:} Set model to training mode%
    \For{each batch $(X_{\text{enc}}, S_{\text{enc}}, X_{\text{dec}}, S_{\text{dec}}, \Delta t, Y) \in \mathcal{D}_{\text{train}}$}
        \State $\hat{S} \gets \mathcal{M}(X_{\text{enc}}, S_{\text{enc}}, X_{\text{dec}}, S_{\text{dec}}, \Delta t)$%
        \State Compute $\mathcal{L}_{\text{data}}$, $\mathcal{L}_{\text{phys}}$, $\mathcal{L}_{\text{cons}}$ as in validation%
        \State $\mathcal{L}_{\text{total}} \gets \mathcal{L}_{\text{data}} + \lambda_{\text{phys}} \cdot \mathcal{L}_{\text{phys}} + \lambda_{\text{cons}} \cdot \mathcal{L}_{\text{cons}}$%
        \State $\nabla \mathbf{W} \gets \frac{\partial \mathcal{L}_{\text{total}}}{\partial \mathbf{W}}$; clip if needed; update $\mathbf{W} \gets \mathbf{W} - \eta \cdot \nabla \mathbf{W}$%
    \EndFor%
\EndFor
\State\Return Best model $\mathcal{M}^*$ with lowest validation loss%
\end{algorithmic}
\end{algorithm}

The PINODE framework provides a principled approach to learning HVAC dynamics that respects physical constraints while maintaining the flexibility of neural network models.
By embedding conservation laws directly into the training objective, the model learns representations that generalize better to unseen operating conditions and maintain thermodynamic consistency over long prediction horizons.
The stabilized latent dynamics, achieved through gated architectures and layer normalization, enable stable gradient flow during backpropagation through the neural ODE, addressing a key challenge in training continuous-time neural models.
In the following sections, we demonstrate how this framework integrates with DAE solvers to handle algebraic constraints in large-scale HVAC systems.

\paragraph{Compressor and expansion valve}

In addition to the PINODE heat exchanger models, the compressor and expansion valve are modeled as static (memoryless) mappings implemented with lightweight multilayer perceptrons (MLPs).
These models take local inlet/outlet thermodynamic conditions and actuation signals (e.g., compressor speed or valve opening) as inputs and output the corresponding mass flow rates and outlet enthalpies needed by the system solver.
We do not elaborate these static models in detail here; the training setup and feature definitions follow~\cite{Ma2024}.

\subsection{Corrector Network for Cyclic Systems\label{sec:corr_nn}}

While physics-informed neural ODEs (PINODEs) provide accurate predictions for individual heat exchangers, system-level simulations involving multiple interconnected components can accumulate errors due to algebraic coupling constraints and numerical integration inaccuracies.
To address this, we introduce a corrector neural network that learns to compensate for systematic biases in mass and energy predictions at the system level.
The corrector network is trained on a short initial segment of the simulation trajectory and then applied throughout the entire simulation to improve prediction accuracy.

\subsubsection{Network Architecture and Formulation}

The corrector network $\phi_\text{corr}: \mathbb{R}^{d_\text{in}} \rightarrow \mathbb{R}^4$ is a fully connected neural network that takes as input the concatenated normalized state and output vectors from both indoor heat exchangers (for dual-compressor simulation):
\begin{equation}
\bm{z}_\text{in} = \begin{bmatrix} \hat{\bm{x}}_{\text{HEX},1} \\ \hat{\bm{y}}_{\text{HEX},1} \\ \hat{\bm{x}}_{\text{HEX},2} \\ \hat{\bm{y}}_{\text{HEX},2} \end{bmatrix} \in \mathbb{R}^{d_\text{in}},
\end{equation}
where $\hat{\bm{x}}_{\text{HEX},i}$ and $\hat{\bm{y}}_{\text{HEX},i}$ are the normalized state and output vectors for the $i$-th heat exchanger, respectively.
The network outputs a correction term $\boldsymbol{\phi}_\text{corr} = \left[\phi^{(1)}_E, \phi^{(2)}_E, \phi^{(1)}_M, \phi^{(2)}_M\right]^\top$ for the four target variables: internal energy $E$ and refrigerant mass $M$ for both compressors.

The network architecture consists of three fully connected layers with sigmoid activations:
\begin{equation}
\boldsymbol{\phi}_\text{corr}(\bm{z}_\text{in}) = \mathtt{ConstrainTanh}\left( \mathbf{W}_3 \sigma(\mathbf{W}_2 \sigma(\mathbf{W}_1 \bm{z}_\text{in} + \mathbf{b}_1) + \mathbf{b}_2) + \mathbf{b}_3 \right),
\end{equation}
where $\sigma$ denotes the sigmoid activation function, $\mathbf{W}_i$ and $\mathbf{b}_i$ are the weight matrices and bias vectors, and $\mathtt{ConstrainTanh}$ applies a scaled hyperbolic tangent to constrain the output within a specified range.
The hidden layer dimension matches the outdoor heat exchanger model's hidden size to maintain consistency with the overall architecture.

\subsubsection{Training Regime Selection}

The corrector network is trained on a short initial segment of the simulation trajectory, typically spanning 850 time steps starting from step 950.
This training regime is selected to capture the system's transient behavior and initial coupling dynamics while keeping the computational cost minimal.
During training, the dual-compressor system simulation is run for this initial segment, collecting pairs of predicted and benchmark (Dymola) values for the mass and energy variables.

The training objective minimizes the mean squared error between the corrected predictions and the benchmark values:
\begin{equation}
    \mathcal{L}_\text{corr} = \frac{1}{T} \sum_{t=1}^{T} \left\| \left\{\hat{\mathbf{m}}_\text{pred}^{(t)} + \boldsymbol{\phi}_\text{corr}(\bm{z}_\text{in}^{(t)})\right\} - \hat{\mathbf{m}}_\text{bench}^{(t)} \right\|^2,
\end{equation}
where $\hat{\mathbf{m}}_\text{pred}^{(t)} = \left[\hat{E}_1^{(t)}, \hat{E}_2^{(t)}, \hat{M}_1^{(t)}, \hat{M}_2^{(t)}\right]^\top$ are the normalized predicted mass and energy values at time step $t$, $\hat{\mathbf{m}}_\text{bench}^{(t)}$ are the corresponding normalized benchmark values from Dymola, and $T$ is the number of training time steps.
The loss function penalizes deviations between the corrected predictions and the reference benchmark, encouraging the network to learn systematic correction patterns that compensate for accumulated errors.

The training procedure, detailed in Algorithm~\ref{alg:corrnn_training}, begins by running the dual-compressor system simulation for the selected training segment (typically 850 steps starting from step 950) using the physics-informed algebraic solver.
This generates a sequence of system states, from which we extract the normalized state and output vectors from both condensers to form the input vectors $\bm{z}_\text{in}^{(t)}$.
Simultaneously, we collect the corresponding benchmark mass and energy values from the high-fidelity Dymola simulation, which serve as the ground truth targets.

The network is trained using the Adam optimizer with a learning rate of $10^{-3}$ for 500,000 epochs, ensuring convergence to a stable correction mapping.
The training process iteratively adjusts the network weights to minimize the correction loss, learning to predict corrections that, when added to the raw predictions, bring them closer to the benchmark values.
This supervised learning approach allows the corrector network to capture systematic biases that arise from component coupling and numerical integration errors, which are difficult to eliminate through solver parameter tuning alone.

\begin{algorithm}[htbp]%
\caption{Corrector Network Training}%
\label{alg:corrnn_training}%
\begin{algorithmic}[1]%
\Require Dual-compressor system with PINODE models, benchmark Dymola data, training segment $[t_\text{start}, t_\text{end}]$%
\Ensure Trained corrector network $\phi_\text{corr}^*$%
\State Initialize corrector network $\phi_\text{corr}$ with random weights%
\State Initialize Adam optimizer with learning rate $\eta = 10^{-3}$%
\State \textbf{Data Collection Phase}%
\State Run system simulation for segment $[t_\text{start}, t_\text{end}]$ using algebraic solver%
\For{each time step $t \in [t_\text{start}, t_\text{end}]$}
    \State Extract normalized state/output vectors: $\hat{\bm{x}}_{\text{HEX},1}^{(t)}, \hat{\bm{y}}_{\text{HEX},1}^{(t)}, \hat{\bm{x}}_{\text{HEX},2}^{(t)}, \hat{\bm{y}}_{\text{HEX},2}^{(t)}$%
    \State Form input: $\bm{z}_\text{in}^{(t)} \gets [\hat{\bm{x}}_{\text{HEX},1}^{(t)}; \hat{\bm{y}}_{\text{HEX},1}^{(t)}; \hat{\bm{x}}_{\text{HEX},2}^{(t)}; \hat{\bm{y}}_{\text{HEX},2}^{(t)}]$%
    \State Extract predicted values: $\hat{\mathbf{m}}_\text{pred}^{(t)} \gets [\hat{E}_1^{(t)}, \hat{E}_2^{(t)}, \hat{M}_1^{(t)}, \hat{M}_2^{(t)}]$%
    \State Extract benchmark values: $\hat{\mathbf{m}}_\text{bench}^{(t)} \gets$ from Dymola simulation%
\EndFor
\State \textbf{Training Phase}%
\For{epoch $e = 1, 2, \ldots, E_\text{max}$ where $E_\text{max} = 500,000$}
    \For{each time step $t \in [t_\text{start}, t_\text{end}]$}
        \State Compute correction: $\boldsymbol{\phi}_\text{corr}^{(t)} \gets \phi_\text{corr}(\bm{z}_\text{in}^{(t)})$%
        \State Compute corrected prediction: $\hat{\mathbf{m}}_\text{corr}^{(t)} \gets \hat{\mathbf{m}}_\text{pred}^{(t)} + \boldsymbol{\phi}_\text{corr}^{(t)}$%
        \State Compute loss: $\mathcal{L}_\text{corr}^{(t)} \gets \|\hat{\mathbf{m}}_\text{corr}^{(t)} - \hat{\mathbf{m}}_\text{bench}^{(t)}\|^2$%
    \EndFor
    \State Compute average loss: $\mathcal{L}_\text{corr} \gets \frac{1}{T} \sum_{t} \mathcal{L}_\text{corr}^{(t)}$%
    \State Compute gradients: $\nabla \boldsymbol{\theta} \gets \frac{\partial \mathcal{L}_\text{corr}}{\partial \boldsymbol{\theta}}$%
    \State Update weights: $\boldsymbol{\theta} \gets \boldsymbol{\theta} - \eta \cdot \nabla \boldsymbol{\theta}$ using Adam optimizer%
\EndFor
\State\Return Trained network $\phi_\text{corr}^*$ with optimized parameters $\boldsymbol{\theta}^*$%
\end{algorithmic}
\end{algorithm}

\subsubsection{Deployment to HVAC Systems}

\begin{figure}[htbp]
    \centering
    \includegraphics[width=\linewidth]{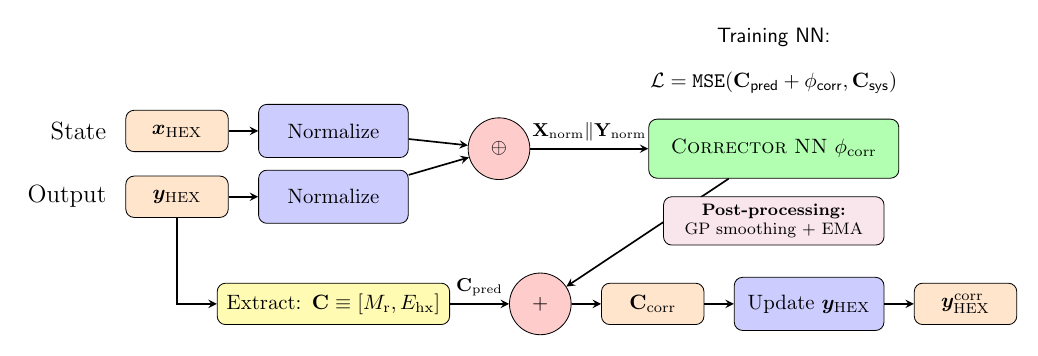}
    \caption{Schematic of the corrector network deployment in the dual-compressor HVAC system simulation. The network takes normalized state and output vectors from both condensers as input and outputs corrections for the mass and energy predictions, which are then smoothed and applied to improve system-level accuracy.}
    \label{fig:schamtic_corrnn}
\end{figure}

After training, the corrector network is deployed throughout the entire simulation to improve prediction accuracy.
Figure~\ref{fig:schamtic_corrnn} shows the schematic of applying the corrector neural network to the dual-compressor system simulation.

At each simulation time step, the corrector network generates a raw correction term $\boldsymbol{\phi}_\text{corr}^\text{raw}(t)$.
To ensure smooth temporal evolution and reduce high-frequency noise, the corrections are post-processed using a two-stage smoothing procedure:
\begin{enumerate}
    \item \textbf{Gaussian Process smoothing}: A Gaussian process regressor with a radial basis function (RBF) kernel and white noise kernel is fitted to the raw correction time series for each output dimension:
    \begin{equation}
    k(t, t') = C \exp\left(-\frac{(t-t')^2}{2\ell^2}\right) + \sigma_n^2 \delta(t-t'),
    \end{equation}
    where $C=1.0$, $\ell=2000$ s, and $\sigma_n^2 = 0.3$ are the kernel hyperparameters.
    The GP provides a smoothed correction $\boldsymbol{\phi}_\text{corr}^\text{GP}(t)$ with uncertainty quantification.
    
    \item \textbf{Exponential moving average}: The GP-smoothed corrections are further smoothed using an exponential moving average (EMA) with $\alpha = 0.95$:
    \begin{equation}
    \boldsymbol{\phi}_\text{corr}^\text{smooth}(t) = \alpha \boldsymbol{\phi}_\text{corr}^\text{smooth}(t-1) + (1-\alpha) \boldsymbol{\phi}_\text{corr}^\text{GP}(t).
    \end{equation}
\end{enumerate}

The final corrected predictions are computed as:
\begin{equation}
\hat{\mathbf{m}}_\text{corr}^{(t)} = \hat{\mathbf{m}}_\text{pred}^{(t)} + \boldsymbol{\phi}_\text{corr}^\text{smooth}(t),
\end{equation}
which are then denormalized and integrated back into the system state.
The correction is only applied if the corrected values remain within the valid normalized range $[-1, 1]$; otherwise, the uncorrected predictions are used to maintain numerical stability.

This approach enables the corrector network to learn and compensate for systematic biases that arise from the coupling between components, numerical integration errors, and model approximations, significantly improving the overall system-level prediction accuracy while maintaining computational efficiency.

\subsection{System Solver\label{sec:system_solver}}

While the PINODE framework captures the dynamics of individual heat exchangers, large-scale HVAC systems with multiple interconnected components introduce algebraic constraints that must be satisfied simultaneously.
These constraints arise from pressure equilibrium conditions at component junctions, mass and energy conservation across the entire system, and the coupling between differential states (mass charge $M_r$ and internal energy $E_{\text{hx}}$) and algebraic variables (pressures, temperatures, mass flow rates).
Standard ODE solvers cannot directly handle these algebraic constraints, necessitating a differential-algebraic equation (DAE) formulation.

\begin{figure}[htbp]
    \centering
    \includegraphics[width=0.35\linewidth]{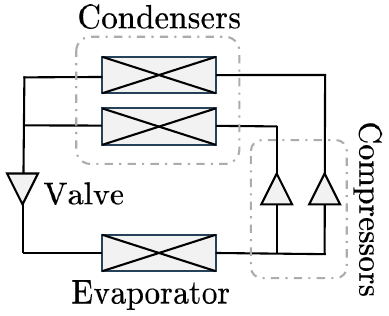}
    \caption{Schematic of the dual-compressor HVAC system topology. The system consists of two compressors, two condensers (indoor heat exchangers), one evaporator (outdoor heat exchanger), and one expansion valve. The components are interconnected through junctions where pressure equilibrium must be maintained, creating algebraic constraints that require a DAE formulation rather than a standard ODE solver.}
    \label{fig:dual_comp_schematic}
\end{figure}

Figure~\ref{fig:dual_comp_schematic} summarizes the dual-compressor cycle used in our system-level studies.
Refrigerant flows through two parallel compressor--condenser branches (indoor heat exchangers), merges at a high-pressure liquid region, expands through a single valve, passes through the outdoor evaporator, and returns through a low-pressure suction path that feeds both compressors.
Each compressor sets mass flow and work input into its condenser; the shared liquid and suction regions couple the branches so that junction pressures and split mass flows must be solved together with the differential states $M_r$ and $E_\text{hx}$ in each heat exchanger.
Those shared manifolds and tees are where mass-flow balance and pressure equilibrium appear as algebraic relations, while $M_r$ and $E_\text{hx}$ evolve according to the PINODE and conservation laws described above---precisely the mixed differential--algebraic structure a DAE solver is meant to treat.

\begin{figure}[htbp]
    \centering
    \includegraphics[width=0.7\linewidth]{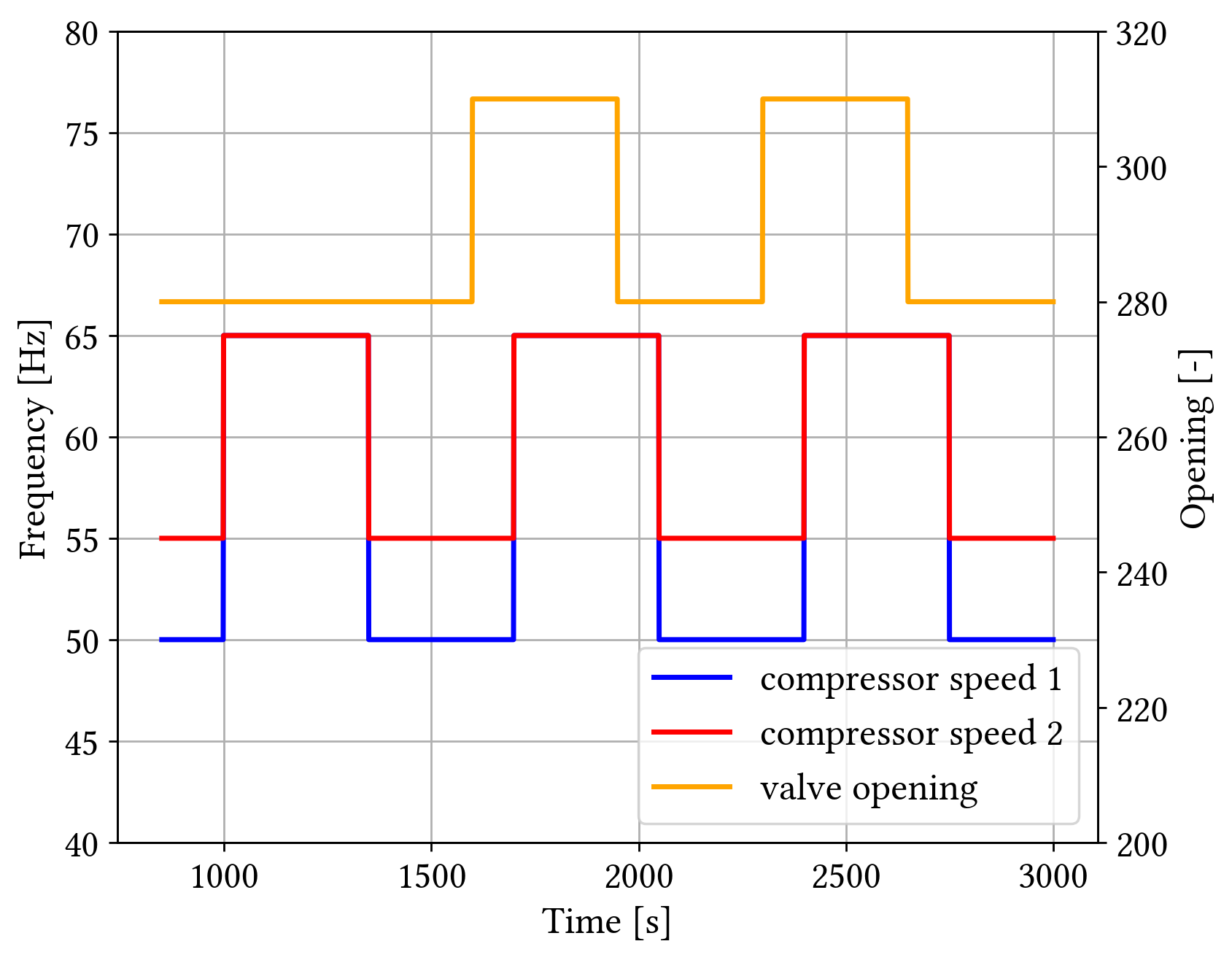}
    \caption{Time-varying actuation signals for the dual-compressor HVAC system of Figure~\ref{fig:dual_comp_schematic}. Compressor speeds and valve opening are shown versus time. Rapid changes in these inputs (jumps with $|\Delta u| > 5.0$) trigger adaptive high-resolution time stepping in the DAE integration, as detailed in Section~\ref{sec:system_solver}.}
    \label{fig:dual_comp_actuation}
\end{figure}

The same topology is driven in simulation by time-varying actuator commands: compressor speed (or capacity) for each compressor and opening of the expansion valve.
Figure~\ref{fig:dual_comp_actuation} plots these actuation signals for the system in Figure~\ref{fig:dual_comp_schematic} over the simulation horizon, so the curves should be read as the boundary inputs applied to the components labeled in the schematic (two compressors, one valve, with dynamics coupled through the interconnecting lines and junctions).

Together, Figures~\ref{fig:dual_comp_schematic} and~\ref{fig:dual_comp_actuation} specify \emph{which} components and couplings the solver resolves and \emph{which} excitations drive the transient: the schematic fixes the network structure and algebraic unknowns, while the actuation traces determine how aggressively that structure is perturbed over time.

\subsubsection{General System Solver Framework}

The system solver framework provides a unified approach for advancing the HVAC system state in time, regardless of whether an algebraic solver or a DAE solver is employed.
The key insight is that both approaches follow the same fundamental sequence: solve algebraic constraints (pressures), evaluate component models (PINODEs), compute state derivatives, and advance the solution.
The primary difference lies in how the time advancement is performed: algebraic solvers use explicit time stepping with fixed or adaptive intervals, while DAE solvers use implicit integration with adaptive error control.

The general framework operates as follows.
At each time step, the system state $\bm{y}$ (containing mass and energy for each heat exchanger) is known, and the solver must determine the state at the next time point.
The algebraic pressure system $\mathbf{g}(\mathbf{p}, \bm{y}, t) = \mathbf{0}$ must be solved first, as the component models (compressors, valves, heat exchangers) require pressure information to compute mass flow rates and other thermodynamic properties.
Once pressures are determined, each component model can be evaluated: compressors and valves compute mass flow rates based on pressure differences, while heat exchangers use PINODE models to predict outlet conditions and heat transfer rates.
These predictions are then used to compute the time derivatives of mass and energy for each heat exchanger, which are integrated to advance the system state.

\begin{algorithm}[htbp]
\caption{General System Solver Framework for HVAC Systems}
\label{alg:system_solver}
\begin{algorithmic}[1]
    \Require Initial state $\bm{y}_0$, time horizon $[t_0, t_{\text{end}}]$, actuation signals $\mathbf{u}(t)$, solver type (Algebraic: Powell hybrid, DAE-IDA, or DAE-DASSL)
    \Ensure Solution trajectory $\{\bm{y}_n, t_n\}_{n=0}^{N}$
    \State Initialize: $\bm{y} \gets \bm{y}_0$, $t \gets t_0$, $n \gets 0$: 
    \State Determine time step $\Delta t$ or $h_n$ based on solver type and adaptive control
    \While{$t < t_{\text{end}}$}
        \State \textbf{Step 1: Solve Algebraic Pressure System}
        \State Solve $\mathbf{g}(\mathbf{p}, \bm{y}, t) = \mathbf{0}$ for junction pressures $\mathbf{p}$
        \State Use Powell hybrid method (algebraic solver) or Levenberg--Marquardt (DAE solvers) with tolerance $\epsilon_\text{soln}$
        \State \textbf{Step 2: Evaluate Component Models}
        \For{each component $i$ in system}
            \State Extract inlet conditions from upstream components and pressures $\mathbf{p}$
            \If{component $i$ is heat exchanger}
                \State $\hat{\bm{y}}_i, \dot{Q}_{a,i} \gets \text{PINODE}_i(\bm{x}_{\text{in},i}, \mathbf{s}_{\text{in},i})$ \Comment{Predict outputs and heat transfer}
            \ElsIf{component $i$ is compressor or valve}
                \State $\dot{m}_i, h_{\text{out},i} \gets \text{ComponentModel}_i(p_{\text{in}}, p_{\text{out}}, h_{\text{in}})$ \Comment{Compute mass flow and enthalpy}
            \EndIf
        \EndFor
        \State \textbf{Step 3: Compute State Derivatives}
        \For{each heat exchanger $i$}
            \State $\dot{M}_{r,i} \gets \dot{m}_{\text{in},i} - \dot{m}_{\text{out},i}$ \Comment{Mass conservation}
            \State $\dot{E}_{\text{hx},i} \gets \dot{m}_{\text{in},i} h_{\text{in},i} - \dot{m}_{\text{out},i} h_{\text{out},i} - \dot{Q}_{a,i}$ \Comment{Energy conservation}
        \EndFor
        \State Assemble derivative vector: $\dot{\bm{y}} \gets [\dot{M}_{r,1}, \dot{E}_{\text{hx},1}, \ldots, \dot{M}_{r,N}, \dot{E}_{\text{hx},N}]^\top$
        \State \textbf{Step 4: Advance Solution}
        \If{solver type is algebraic (Powell hybrid)}
            \State $\bm{y}_{n+1} \gets \bm{y}_n + \dot{\bm{y}} \cdot \Delta t$ \Comment{Explicit Euler step}
            \State $t_{n+1} \gets t_n + \Delta t$
        \ElsIf{solver type is IDA or DASSL (DAE)}
            \State Solve implicit system: $\mathbf{F}(t_{n+1}, \bm{y}_{n+1}, \dot{\bm{y}}_{n+1}) = \mathbf{0}$ \Comment{DAE residual}
            \State Use BDF method with adaptive step size $h_n$ and order $k$
            \State $t_{n+1} \gets t_n + h_n$ (adaptively determined)
        \EndIf
        \State $n \gets n + 1$, $\bm{y} \gets \bm{y}_{n+1}$, $t \gets t_{n+1}$
    \EndWhile
    \State \Return Solution trajectory $\{(\bm{y}_n, t_n)\}_{n=0}^{N}$
\end{algorithmic}
\end{algorithm}

The algorithm highlights the common structure shared by all solver types: the algebraic pressure system must be solved before component evaluation, component models (especially PINODEs) provide the necessary predictions, and state derivatives are computed from conservation laws.
The key distinction is in Step 4: algebraic solvers use explicit time stepping with a fixed or adaptively chosen time step $\Delta t$, while DAE solvers solve an implicit system that enforces the DAE residual $\mathbf{F}(t, \bm{y}, \dot{\bm{y}}) = \mathbf{0}$ using adaptive BDF methods.
This unified framework ensures that regardless of the solver choice, the system maintains thermodynamic consistency through proper handling of algebraic constraints and conservation laws.

\subsubsection{Algebraic Solver}

The default algebraic solver uses the Powell hybrid method (via \texttt{scipy.optimize.root} with \texttt{method='hybr'}) to solve for junction pressures that satisfy mass flow conservation throughout the system.
At each time step, the solver finds the junction pressures $\mathbf{p} = [p_1, p_2, \ldots, p_{n_p}]^\top$ by solving the nonlinear system of residual equations:
\begin{equation}
\mathbf{r}(\mathbf{p}) = \mathbf{0},
\end{equation}
where each residual $r_i$ enforces mass flow balance at junction $i$:
\begin{equation}
r_i = \frac{\dot{m}_{\text{in},i} - \dot{m}_{\text{out},i}}{m_{\text{scale}}},
\end{equation}
with $m_{\text{scale}} = 0.01$ kg/s used for numerical scaling.
The mass flow rates $\dot{m}_{\text{in},i}$ and $\dot{m}_{\text{out},i}$ are computed by evaluating each component's mass flow model (compressor, valve, or heat exchanger) using the current junction pressures and upstream conditions.
The Powell hybrid method combines Powell's method with a quasi-Newton approach, making it robust for systems with $n_p$ unknowns where $n_p$ is the number of junction pressures.
For large systems with more than 10 pressures, the implementation automatically switches to a bounded least-squares solver (\texttt{scipy.optimize.least\_squares} with \texttt{method='trf'}) with pressure bounds $[2 \times 10^5, 6 \times 10^6]$ Pa to ensure physical feasibility and improve convergence.

The algebraic solver employs adaptive time stepping based on a Runge--Kutta--Fehlberg (RK45) method to determine the optimal time step size $\Delta t$ at each iteration.
The adaptive stepping mechanism leverages the PINODE model's latent dynamics to predict the time step that maintains numerical accuracy while maximizing computational efficiency.
Specifically, the RK45 method computes both 4th-order ($z_4$) and 5th-order ($z_5$) solutions to the latent ODE:
\begin{equation}
\frac{d\mathbf{h}}{dt} = f_\theta(\mathbf{h}(t), \bm{x}(t), \mathbf{s}(t)),
\end{equation}
where $\mathbf{h}(t)$ is the latent state, $\bm{x}(t)$ and $\mathbf{s}(t)$ are the feature and state vectors, and $f_\theta$ is the neural ODE function.
The error estimate $\epsilon = \|z_4 - z_5\|$ is used to adaptively adjust the time step:
\begin{equation}
\Delta t_{\text{new}} = 2.5 \left(\frac{0.6 \cdot \epsilon_{\Delta t} \cdot \Delta t_0}{\epsilon}\right)^{1/4},
\end{equation}
where $\epsilon_{\Delta t}$ is the time step tolerance (typically $1.1 \times 10^{-3}$ or $0.3$ for stability) and $\Delta t_0$ is the current time step.
This adaptive mechanism ensures that the time step is automatically reduced when the system dynamics are rapidly changing (e.g., during control input discontinuities) and increased when the dynamics are smooth, balancing accuracy and computational efficiency throughout the simulation.

\subsubsection{DAE Solver and Formulation}

For a large-scale HVAC system with $n_c$ compressors, $n_v$ valves, $n_{\text{cond}}$ condensers, and $n_{\text{evap}}$ evaporators, we formulate the system as an index-1 DAE.
The differential variables are the mass charge $M_r$ and internal energy $E_{\text{hx}}$ for each heat exchanger, while the algebraic variables are the junction pressures $p_j$, which are solved separately using a root-finding algorithm.

The state vector $\bm{y} \in \mathbb{R}^{2(n_{\text{cond}}+n_{\text{evap}})}$ contains only the differential variables:
\begin{equation}
\bm{y} = \begin{bmatrix}
M_{r}^{(1)}, M_{r}^{(2)}, \ldots, M_{r}^{(n_{\text{cond}}+n_{\text{evap}})}\\
E_{\text{hx}}^{(1)}, E_{\text{hx}}^{(2)}, \ldots, E_{\text{hx}}^{(n_{\text{cond}}+n_{\text{evap}})}
\end{bmatrix}^\top,
\end{equation}%
where the first $n_{\text{cond}}+n_{\text{evap}}$ elements correspond to refrigerant mass for each heat exchanger, and the remaining elements correspond to internal energy.
The DAE residual function $\mathbf{F}(t, \bm{y}, \dot{\bm{y}}) = \mathbf{0}$ is formulated as:
\begin{equation}
\mathbf{F}(t, \bm{y}, \dot{\bm{y}}) = \dot{\bm{y}} - \mathbf{f}(t, \bm{y}),
\end{equation}
where $\mathbf{f}(t, \bm{y})$ is the right-hand side function that computes the time derivatives of mass and energy for each heat exchanger:
\begin{align}
\dot{M}_{r,i} &= \dot{m}_{\text{in},i} - \dot{m}_{\text{out},i}, \\
\dot{E}_{\text{hx},i} &= \dot{m}_{\text{in},i} h_{\text{in},i} - \dot{m}_{\text{out},i} h_{\text{out},i} - \dot{Q}_{a,i},
\end{align}
where $\dot{m}_{\text{in},i}$ and $\dot{m}_{\text{out},i}$ are the inlet and outlet mass flow rates for heat exchanger $i$, $h$ denotes specific enthalpy, and $\dot{Q}_{a,i}$ is the heat transfer rate predicted by the PINODE model for that component.

The mass flow rates and enthalpies depend on the current system state and the junction pressures, which are determined by solving the algebraic constraint system:
\begin{equation}
\mathbf{g}(\mathbf{p}, \bm{y}, t) = \mathbf{0},
\end{equation}
where $\mathbf{p} = [p_1, p_2, \ldots, p_{n_p}]^\top$ are the junction pressures.
These constraints encode pressure equilibrium at component junctions, mass flow conservation at splitters and mergers, and component-specific relationships (e.g., compressor and valve characteristics).
The pressure system is solved using a Levenberg--Marquardt algorithm (via \texttt{scipy.optimize.least\_squares}) at each time step to ensure fast convergence.

We implement two DAE solvers to solve the implicit system $\mathbf{F}(t, \bm{y}, \dot{\bm{y}}) = \mathbf{0}$: the IDA solver and the DASSL solver.
Both solvers use variable-order, variable-step backward differentiation formula (BDF) methods, but differ in their implementation details and numerical characteristics.

At each time step $t_n$, both solvers approximate the derivative using a $k$-th order BDF formula:
\begin{equation}
\dot{\bm{y}}_n \approx \frac{1}{h_n} \sum_{i=0}^{k} \alpha_{i,k} \bm{y}_{n-i},
\end{equation}
where $h_n = t_n - t_{n-1}$ is the step size, and $\alpha_{i,k}$ are BDF coefficients for orders $k \in \{1,2,3,4,5\}$.
Substituting into the DAE residual yields a nonlinear system:
\begin{equation}
\mathbf{G}(\bm{y}_n) = \mathbf{F}\left(t_n, \bm{y}_n, \frac{1}{h_n}\sum_{i=0}^{k} \alpha_{i,k} \bm{y}_{n-i}\right) = \mathbf{0},
\end{equation}
which both solvers solve via modified Newton iteration.
Both solvers adaptively select the step size $h_n$ and BDF order $k$ based on local truncation error estimates, and use weighted root-mean-square error norms with absolute and relative tolerances (both set to $\text{ATOL} = \text{RTOL} = \epsilon_\text{soln}$, typically $10^{-6}$) to control accuracy.

\paragraph{IDA Solver}
We implement the IDA (Implicit Differential-Algebraic) solver using the SUNDAE library (specifically \texttt{sksundae}), which provides a Python interface to the SUNDIALS IDA solver.
The key distinguishing feature of IDA is its use of standard modified Newton iteration without explicitly forming the step-dependent Jacobian matrix.
Additional parameters include maximum and minimum step sizes ($h_\text{max}$, $h_\text{min}$) and the output interval ($\Delta t_\text{out}^\text{IDA}$), which controls the temporal resolution of saved solution points.
If the SUNDAE IDA solver is unavailable, the implementation falls back to CVODE (also from SUNDIALS) using the Adams--Moulton method, though this is less suitable for DAE systems.

\paragraph{DASSL Solver}
We employ the DASSL (Differential-Algebraic System Solver) algorithm~\cite{petzold1982dassl} via the PyDAS Python wrapper, which provides access to the original DASSL Fortran implementation.
The main difference from IDA is that DASSL explicitly forms and uses a step-dependent Jacobian matrix in its Newton iteration:
\begin{equation}
\mathbf{J}_n = \frac{\partial \mathbf{F}}{\partial \bm{y}} + \frac{\alpha_{0,k}}{h_n} \frac{\partial \mathbf{F}}{\partial \dot{\bm{y}}}.
\end{equation}
This explicit Jacobian formulation can provide better convergence properties for certain stiff systems.
The implementation uses adaptive steps with intermediate target times, saving solution points only when a minimum output interval ($\Delta t_\text{out,min}^\text{DASSL}$) has elapsed, allowing DASSL to choose its own internal time steps while maintaining control over output density.

\subsubsection{Integration with PINODE Framework}

The DAE solver integrates seamlessly with the PINODE models for each heat exchanger component through a tightly coupled iterative procedure.
At each time step, the following sequence is executed:

\begin{enumerate}
\item \textbf{Initialize state}: The DAE solver starts with the current system state $\bm{y}_n$ (mass and energy for each heat exchanger) and determines the time step $h_n$ based on adaptive error control.

\item \textbf{Solve pressure system}: The junction pressures $\mathbf{p}$ are solved using a Levenberg--Marquardt algorithm (via \texttt{scipy.optimize.least\_squares}) to satisfy the algebraic constraints:
   \begin{equation}
   \mathbf{g}(\mathbf{p}, \bm{y}_n, t_n) = \mathbf{0}.
   \end{equation}
   This ensures pressure equilibrium at junctions and mass flow conservation throughout the network.

\item \textbf{Predict heat exchanger outputs}: For each heat exchanger, the PINODE model receives the current inputs (inlet conditions from upstream components, boundary pressures from the solved pressure system) and predicts the outlet states (pressures, enthalpies, temperatures) and heat transfer rate $\dot{Q}_a$.

\item \textbf{Compute mass and energy derivatives}: The predicted $\dot{Q}_a$ and flow properties (mass flow rates $\dot{m}_{\text{in}}$, $\dot{m}_{\text{out}}$, enthalpies $h_{\text{in}}$, $h_{\text{out}}$) are used to evaluate the right-hand side of the mass and energy conservation equations:
   \begin{align}
   \dot{M}_{r,i} &= \dot{m}_{\text{in},i} - \dot{m}_{\text{out},i}, \\
   \dot{E}_{\text{hx},i} &= \dot{m}_{\text{in},i} h_{\text{in},i} - \dot{m}_{\text{out},i} h_{\text{out},i} - \dot{Q}_{a,i}.
   \end{align}

\item \textbf{Advance DAE solution}: The DAE solver (IDA or DASSL) advances to the next time step using the computed residual function $\mathbf{F}(t, \bm{y}, \dot{\bm{y}}) = \dot{\bm{y}} - \mathbf{f}(t, \bm{y})$, where $\mathbf{f}(t, \bm{y})$ contains the mass and energy derivatives computed in the previous step.
\end{enumerate}

This tight coupling ensures that the neural network predictions respect the system-wide algebraic constraints, maintaining thermodynamic consistency across all components.
The iterative nature of this process (pressure solving $\rightarrow$ PINODE prediction $\rightarrow$ derivative evaluation $\rightarrow$ DAE advancement) ensures that the solution satisfies both the differential equations and the algebraic constraints simultaneously.

\subsubsection{Bayesian Optimization for Parameter Tuning}

For large-scale systems, the DAE solver performance depends critically on solver-specific parameters that control time stepping, tolerances, and output resolution.
We employ Bayesian optimization~\cite{mockus1975bayesian} to automatically tune these parameters for each solver type.

For the algebraic solver (Powell hybrid method), we optimize two parameters: $\boldsymbol{\theta}_\text{alg} = [\epsilon_{\Delta t}, \epsilon_\text{soln}]$, where $\epsilon_{\Delta t}$ is the time step tolerance and $\epsilon_\text{soln}$ is the solution tolerance for the root-finding algorithm.

For the DAE-IDA solver, we optimize five parameters: $\boldsymbol{\theta}_{\text{IDA}} = [\epsilon_{\Delta t}, \epsilon_\text{soln}, h_\text{max}, h_\text{min}, \Delta t_\text{out}^\text{IDA}]$, where $h_\text{max}$ and $h_\text{min}$ control the adaptive step size bounds, and $\Delta t_\text{out}^\text{IDA}$ controls the output interval.

For the DAE-DASSL solver, we optimize six parameters: $\boldsymbol{\theta}_{\text{DASSL}} = [\epsilon_{\Delta t}, \epsilon_\text{soln}, h_\text{max}, h_\text{min}, \Delta t_\text{out,min}^\text{DASSL}, N_\text{max}^\text{DASSL}]$, where $\Delta t_\text{out,min}^\text{DASSL}$ controls the minimum spacing between saved solution points and $N_\text{max}^\text{DASSL}$ limits the maximum number of internal DASSL steps.

The optimization objective minimizes a weighted combination of the mean absolute percentage error (MAPE) over all system outputs and the simulation time:
\begin{equation}
\boldsymbol{\theta}^* = \arg\min_{\boldsymbol{\theta} \in \mathcal{X}} \left[ w_{\text{MAPE}} \cdot \text{MAPE}_{\text{all}}(\boldsymbol{\theta}) + w_{\text{time}} \cdot t_{\text{simulation}}(\boldsymbol{\theta}) \right],
\end{equation}
where $w_{\text{MAPE}}$ and $w_{\text{time}}$ are weighting factors (in our implementation, $w_{\text{MAPE}} = w_{\text{time}} = 0.5$), and $\mathcal{X}$ is the parameter space defined by log-uniform distributions over specified ranges.

Using a Gaussian process surrogate model with expected improvement (EI) acquisition function, the algorithm efficiently explores the parameter space with typically 100 function evaluations, converging to near-optimal solver settings that balance accuracy and computational efficiency.

\begin{algorithm}[htbp]%
\begin{algorithmic}[1]
\caption{Bayesian optimization of parameter optimization}
\label{alg:bayesian_param}
\State \textbf{Initialize:} Sample $n_0$ points $\{\boldsymbol{\theta}_i\}_{i=1}^{n_0}$ uniformly, evaluate $y_i = f(\boldsymbol{\theta}_i)$
\For{$n = n_0+1$ to $N$}
    \State Fit GP on $\mathcal{D}_{n-1}$ to get $\mu_{n-1}, \sigma_{n-1}$
    \State $\boldsymbol{\theta}_n = \arg\max_{\boldsymbol{\theta}} \mathrm{EI}(\boldsymbol{\theta})$ \Comment{Find next point}
    \State Evaluate $y_n = f(\boldsymbol{\theta}_n)$ \Comment{Run DAE simulation}
    \State Update $\mathcal{D}_n = \mathcal{D}_{n-1} \cup \{(\boldsymbol{\theta}_n, y_n)\}$
\EndFor
\State \Return $\boldsymbol{\theta}^* = \arg\min_{(\boldsymbol{\theta}, y) \in \mathcal{D}_N} y$
\end{algorithmic}
\end{algorithm}

\subsubsection{Adaptive High-Resolution Stepping}

To handle rapid changes in control inputs (e.g., compressor speed or valve opening), we implement adaptive high-resolution time stepping around control discontinuities for both IDA and DASSL solvers.
The algorithm pre-computes a high-resolution mask by detecting control jumps where $|\Delta u| > \tau$ (typically $\tau = 5.0$ units) in the actuation signals.
For each detected jump, a high-resolution window is created spanning $n_\text{pre} = 5$ steps before and $n_\text{post} = 50$ steps after the jump.

The target output spacing is adaptively determined based on the current time index:
\begin{equation}
\Delta t_\text{target} = \begin{cases}
2.5 \text{ s} & \text{if in high-resolution window}, \\
7.5 \text{ s} & \text{otherwise}.
\end{cases}
\end{equation}

However, the implementation differs between IDA and DASSL solvers due to their different interfaces and adaptive stepping mechanisms.

\paragraph{IDA Solver}
For the IDA solver (implemented via the SUNDAE library, which provides a Python interface to SUNDIALS IDA), the target spacing $\Delta t_\text{target}$ directly controls the output interval $\Delta t_\text{out}^\text{IDA}$.
The solver pre-computes evaluation points $t_\text{eval} = \{t_0, t_0 + \Delta t_\text{target}, t_0 + 2\Delta t_\text{target}, \ldots, t_1\}$ and passes them to the IDA solver via \texttt{ida\_solver.solve($t_\text{eval}$, $y_0$, $y'_0$)}.
The IDA solver then performs adaptive BDF stepping internally between these output points, automatically adjusting its internal time step size $h_n$ based on local truncation error estimates while ensuring outputs are provided at the specified evaluation points.

\paragraph{DASSL Solver}
For the DASSL solver (implemented via the PyDAS Python wrapper for the DASSL Fortran library), the adaptive stepping mechanism is more incremental.
The solver uses $\Delta t_\text{target}$ to determine an intermediate step increment $\Delta t_\text{increment} = 0.5 \cdot \Delta t_\text{target}$ and a minimum output spacing $\Delta t_\text{out,min}^\text{DASSL}$ (typically $0.1 \cdot \Delta t_\text{target}$ or from the \texttt{pydas\_min\_output\_dt} parameter).
During simulation, the solver calls \texttt{dae\_solver.step($t_\text{next}$)} repeatedly with $t_\text{next} = \min(t_\text{current} + \Delta t_\text{increment}, t_1)$, allowing DASSL to adaptively choose its internal step size.
Outputs are saved only when the time since the last saved output exceeds $\Delta t_\text{out,min}^\text{DASSL}$, ensuring appropriate temporal resolution while allowing DASSL to take smaller internal steps when needed for stability.

Both approaches ensure accurate resolution of transient dynamics during control changes while maintaining computational efficiency during steady-state operation.
The implementation also includes output caching to avoid redundant computations when transitioning from high-resolution to low-resolution regions.

\section{Results \& Discussions\label{sec:results}}

\subsection{Training PINODE Models\label{sec:pinode_results}}

The physics-informed neural ODE (PINODE) models for individual heat exchangers form the foundation of our system-level simulation framework.
We train separate PINODE models for indoor (condenser) and outdoor (evaporator) heat exchangers using the training procedure described in Section~\ref{sec:pinode}.

\begin{figure}[htbp]
    \centering
    \includegraphics[width=0.85\linewidth]{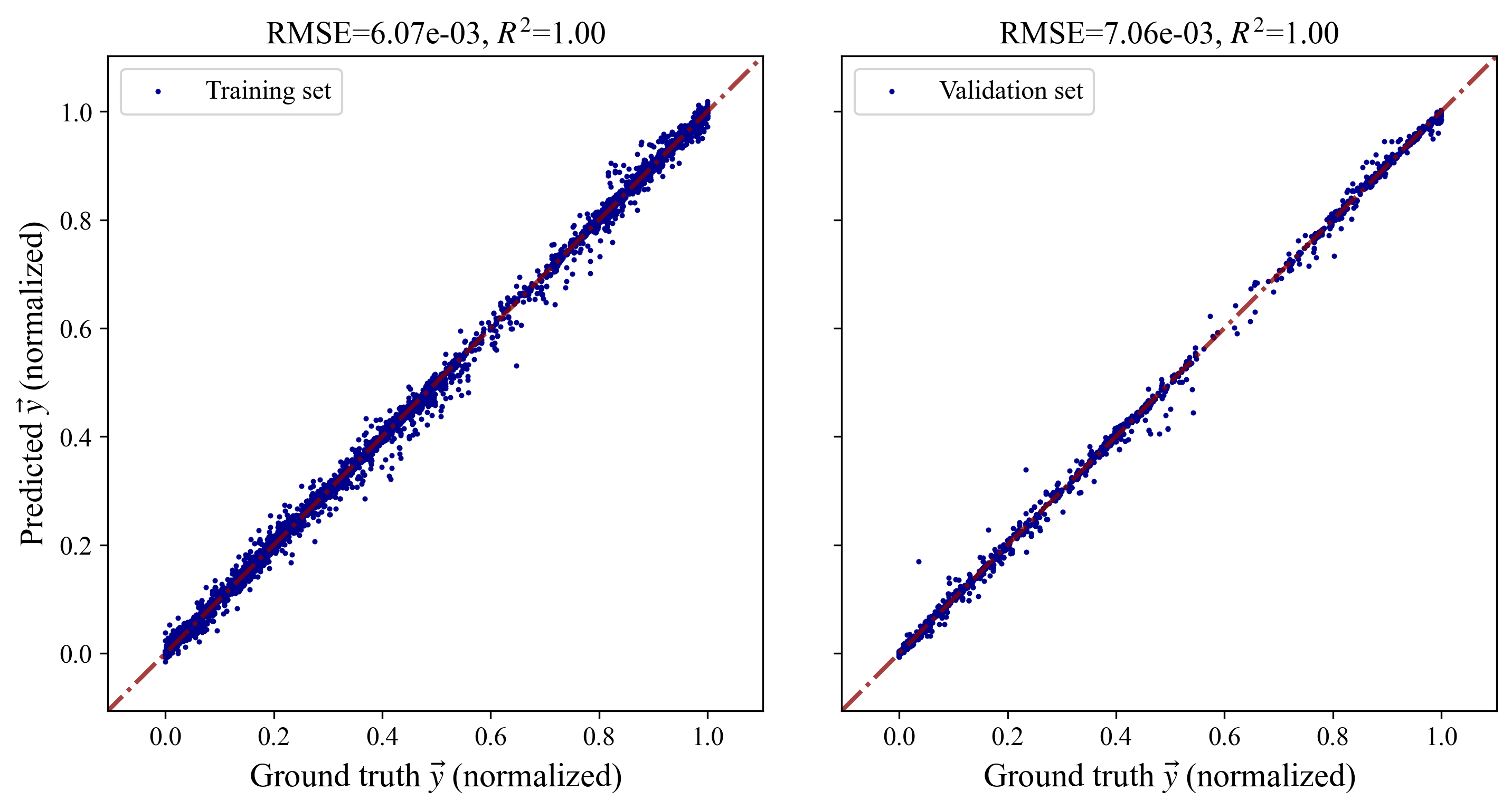}
    \caption{Parity plot comparing predicted versus true outputs for the training and testing sets of the condenser (indoor heat exchanger) PINODE model. The model demonstrates excellent agreement with the reference data across both datasets.}
    \label{fig:parity_indoor}
\end{figure}

\begin{figure}[htbp]
    \centering
    \includegraphics[width=0.85\linewidth]{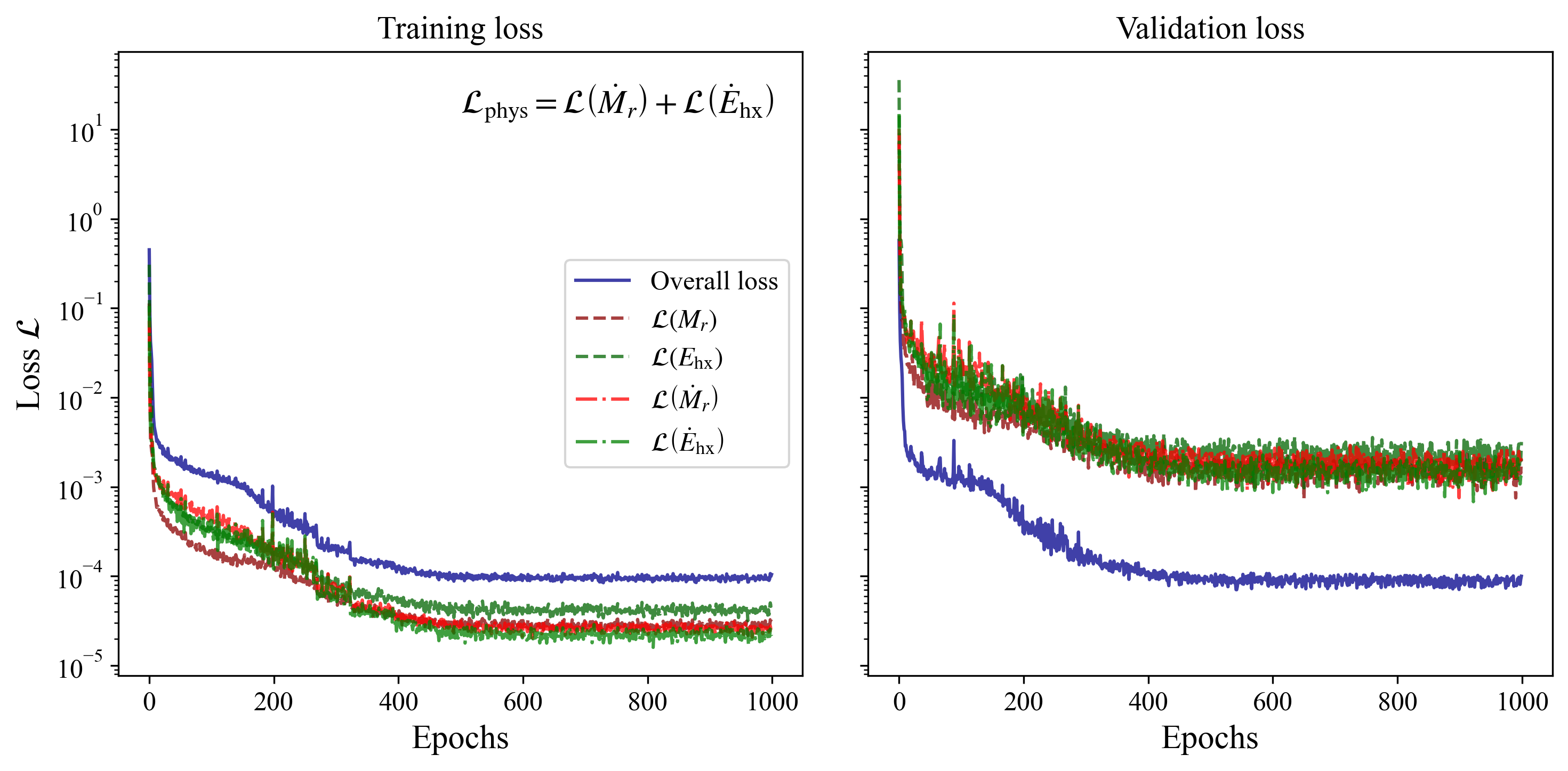}
    \caption{Loss history for the training and testing sets of the condenser (indoor heat exchanger) PINODE model. The stable convergence with minimal gap between training and testing losses indicates good generalization and minimal overfitting.}
    \label{fig:loss_indoor}
\end{figure}

The PINODE models demonstrate excellent performance on both training and testing datasets, as shown in Figure~\ref{fig:parity_indoor}.
The parity plots reveal strong linear correlation between predicted and true values, with data points closely aligned along the diagonal, indicating high prediction accuracy.
The loss history in Figure~\ref{fig:loss_indoor} indicates stable convergence with minimal overfitting, as evidenced by the parallel trajectories of training and testing losses.

The physics-informed loss component $\mathcal{L}_\text{phys}$, which penalizes violations of conservation laws through gradient information, and the conservation loss component $\mathcal{L}_\text{cons}$, which directly constrains mass and energy predictions, operate on similar data scales, enabling effective multi-objective optimization.
For the validation set, the overall loss is typically smaller because the mean squared error contributions from individual components (data, physics, and conservation terms) are balanced, demonstrating that the model successfully learns to satisfy both data fidelity and physical constraints simultaneously.
This validates the model architecture and training procedure, ensuring that the well-trained PINODE models serve as reliable building blocks for the system-level simulations discussed in the following sections.

\subsection{Corrector Network for System Simulations\label{sec:corrector_net_results}}

While individual PINODE models provide accurate predictions for isolated heat exchangers, system-level simulations involving multiple interconnected components can accumulate errors due to algebraic coupling constraints, numerical integration inaccuracies, and model approximations.
To address this, we employ a corrector network (described in Section~\ref{sec:corr_nn}) that learns to compensate for systematic discrepancies between predicted and true mass-energy (ME) terms at the system level.

We begin with manually inspected, physically justifiable parameters (e.g., $\epsilon_{\Delta t}=10^{-3}$) for the system solver.
The corrector network is trained on a short initial segment of the simulation trajectory (850 time steps) and then deployed throughout the entire simulation to improve prediction accuracy.
This lightweight neural network takes normalized state and output predictions from both condensers as input and outputs correction terms for the four ME components ($E_1$, $E_2$, $M_1$, $M_2$).

\begin{figure}[htbp]
    \centering
    \includegraphics[width=0.6\linewidth]{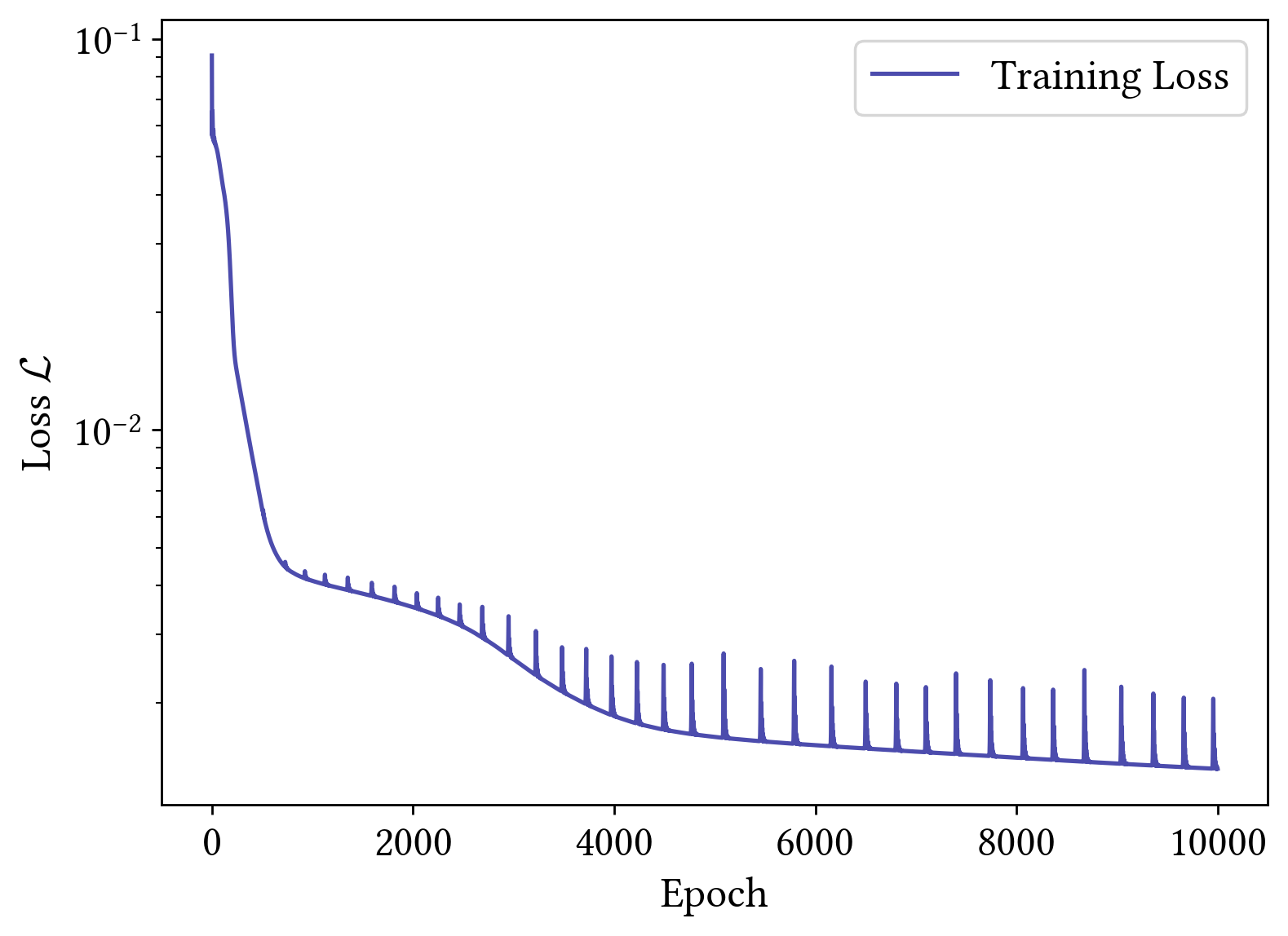}
    \caption{Training loss history of the corrector neural network. The network converges rapidly, demonstrating efficient learning of systematic biases in mass-energy predictions.}
    \label{fig:loss_corrnn}
\end{figure}

The training loss of the corrector network is shown in Figure~\ref{fig:loss_corrnn}.
The network converges rapidly, with training completing in approximately 30 seconds, demonstrating the computational efficiency of this approach.
The monotonic decrease in loss indicates stable learning, and the final low loss value suggests that the network successfully captures the systematic biases present in the system-level predictions.

\begin{figure}[htbp]
    \centering
    {\small\hspace{35pt}Before applying corrector NN\hspace{80pt}After applying corrector NN}\\
    (a)\includegraphics[width=0.39\linewidth]{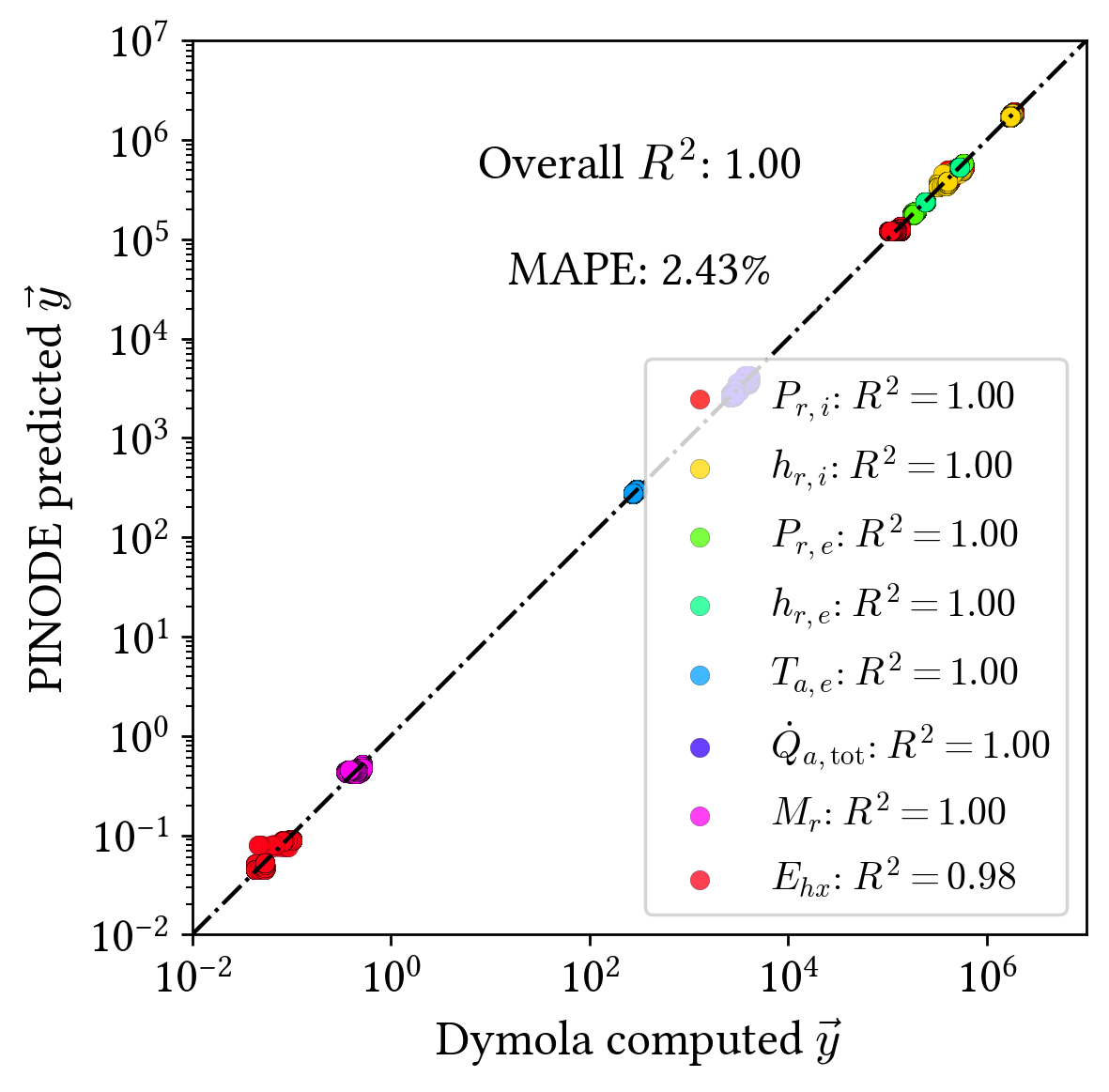}
    \hspace{10pt}
    (b)\includegraphics[width=0.39\linewidth]{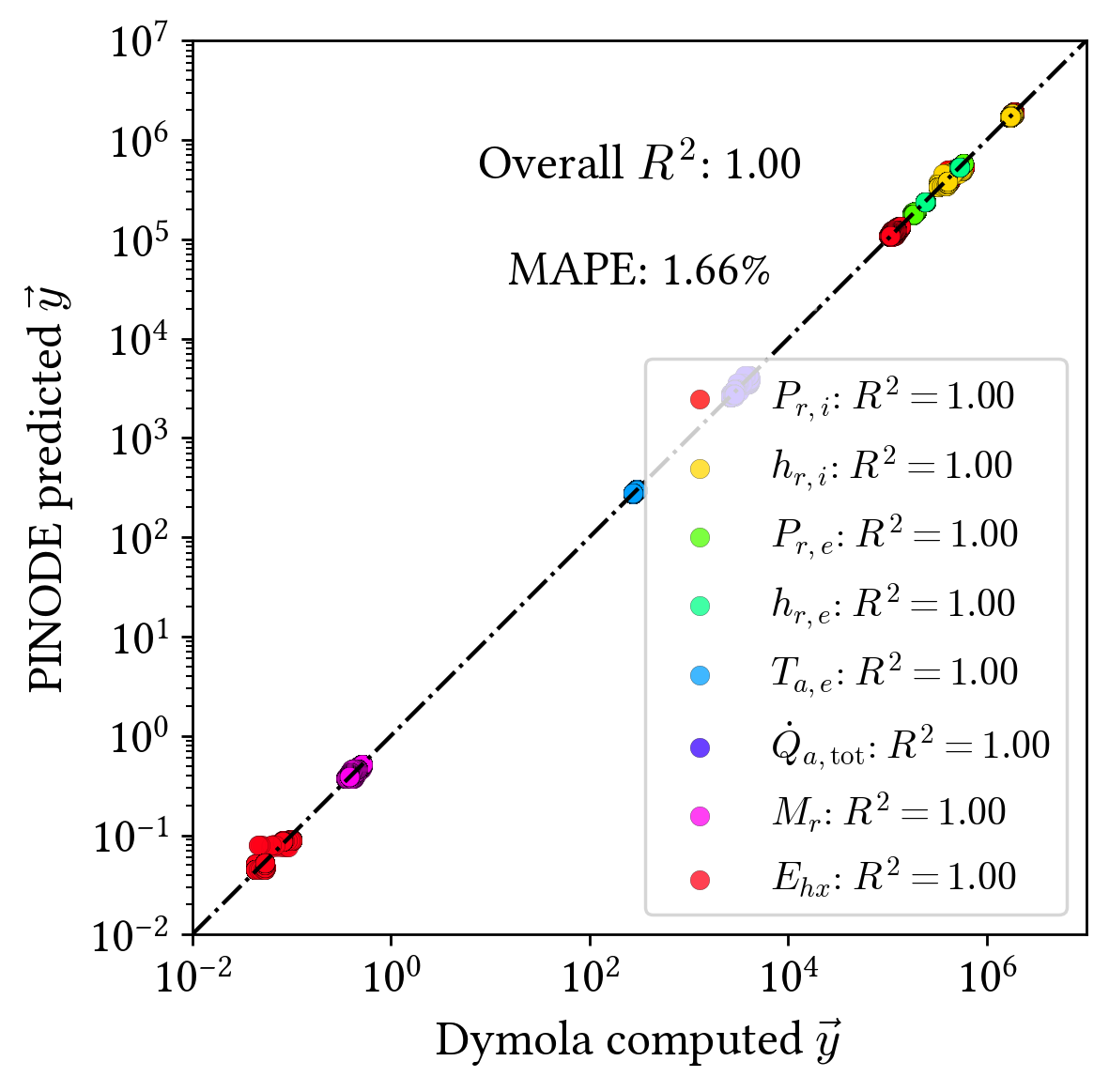}
    \caption{Comparison of parity plots (a) before and (b) after applying the corrector neural network. The corrector network significantly improves the agreement between PINODE predictions and the reference Dymola simulation results, with predictions clustering much closer to the diagonal.}
    \label{fig:parity_comparison}
\end{figure}

The effectiveness of the corrector network is immediately apparent when comparing system-level predictions.
Figure~\ref{fig:parity_comparison} shows a significant improvement in the parity plot after applying the corrector network, with predictions much closer to the reference Dymola simulation results.
The baseline predictions (without corrector) show noticeable scatter and deviation from the line of perfect prediction, particularly for extreme values, while the corrected predictions exhibit much tighter clustering around the ideal one-to-one correspondence, indicating substantially improved accuracy.

\begin{figure}[htbp]
    \centering
    \includegraphics[width=0.6\linewidth]{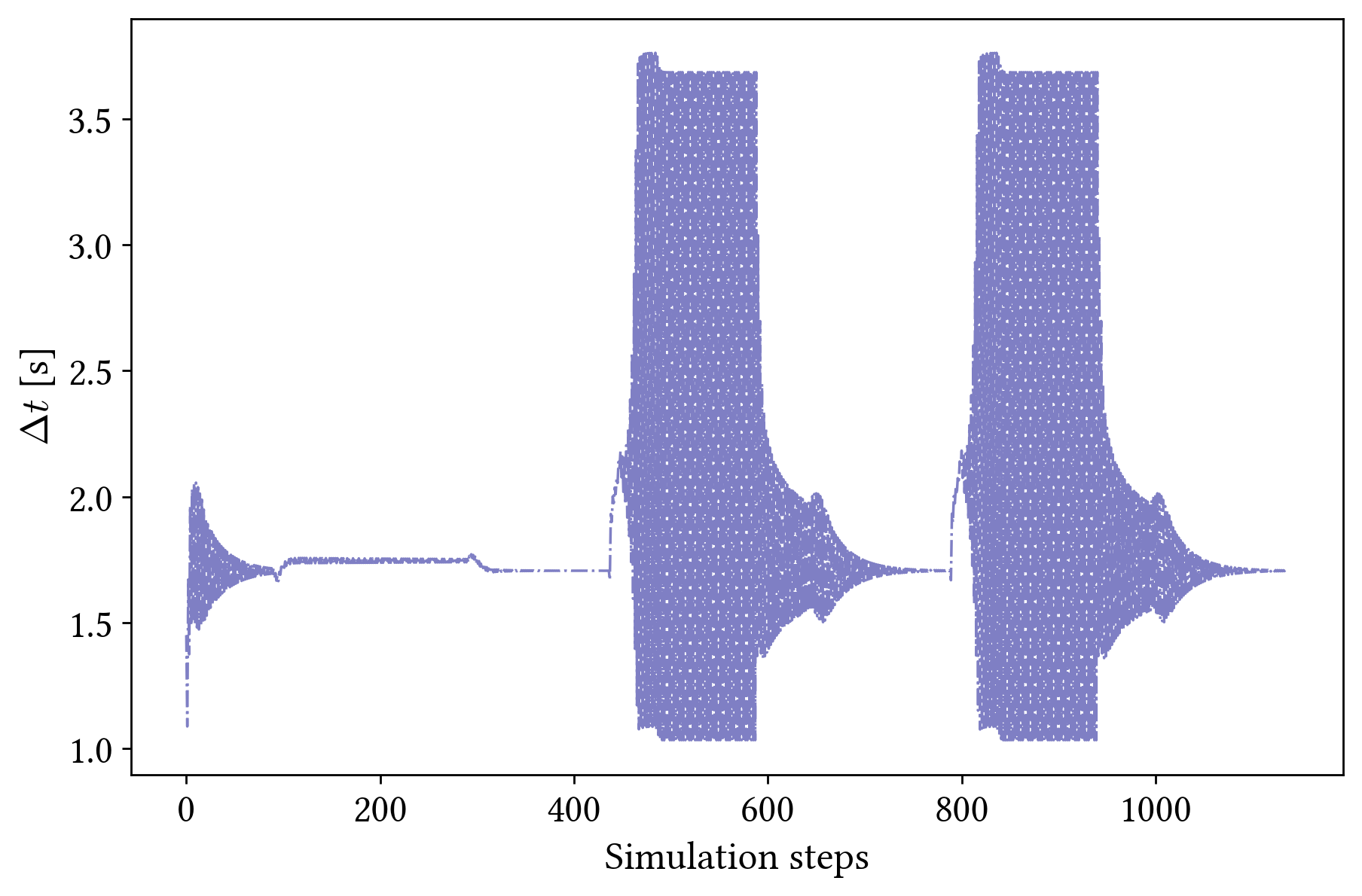}
    \caption{Adaptive time step evolution during the dual-compressor cycle simulation with the corrector network. The solver adjusts the time step size based on local error estimates, using smaller steps during transients and larger steps during steady operation.}
    \label{fig:time_stepping_corrnn}
\end{figure}

The time-stepping behavior of the dual-compressor cycle is illustrated in Figure~\ref{fig:time_stepping_corrnn}, highlighting the adaptive time step selection throughout the simulation.
The observed step size variation is closely tied to the system dynamics: the solver automatically reduces the time step during periods of rapid transients (e.g., control input changes and strong component interactions), while increasing it during quasi-steady regimes where the solution evolves more smoothly.
This behavior reflects effective local error control, allowing the solver to allocate computational effort where it is most needed without uniformly refining the entire trajectory.
Importantly, the presence of the corrector network does not introduce instability or excessive stiffness; instead, it preserves stable integration while enabling accurate tracking of system dynamics with a balanced computational cost.
Overall, the adaptive stepping pattern provides indirect evidence that the combined PINODE–corrector framework produces numerically well-behaved trajectories suitable for efficient time integration.

\begin{figure}[htbp]
    \centering
    {\small Before applying corrector NN\hspace{60pt}After applying corrector NN\quad\hspace{30pt}\quad\quad\quad}\\
    (a)\includegraphics[height=0.26\linewidth]{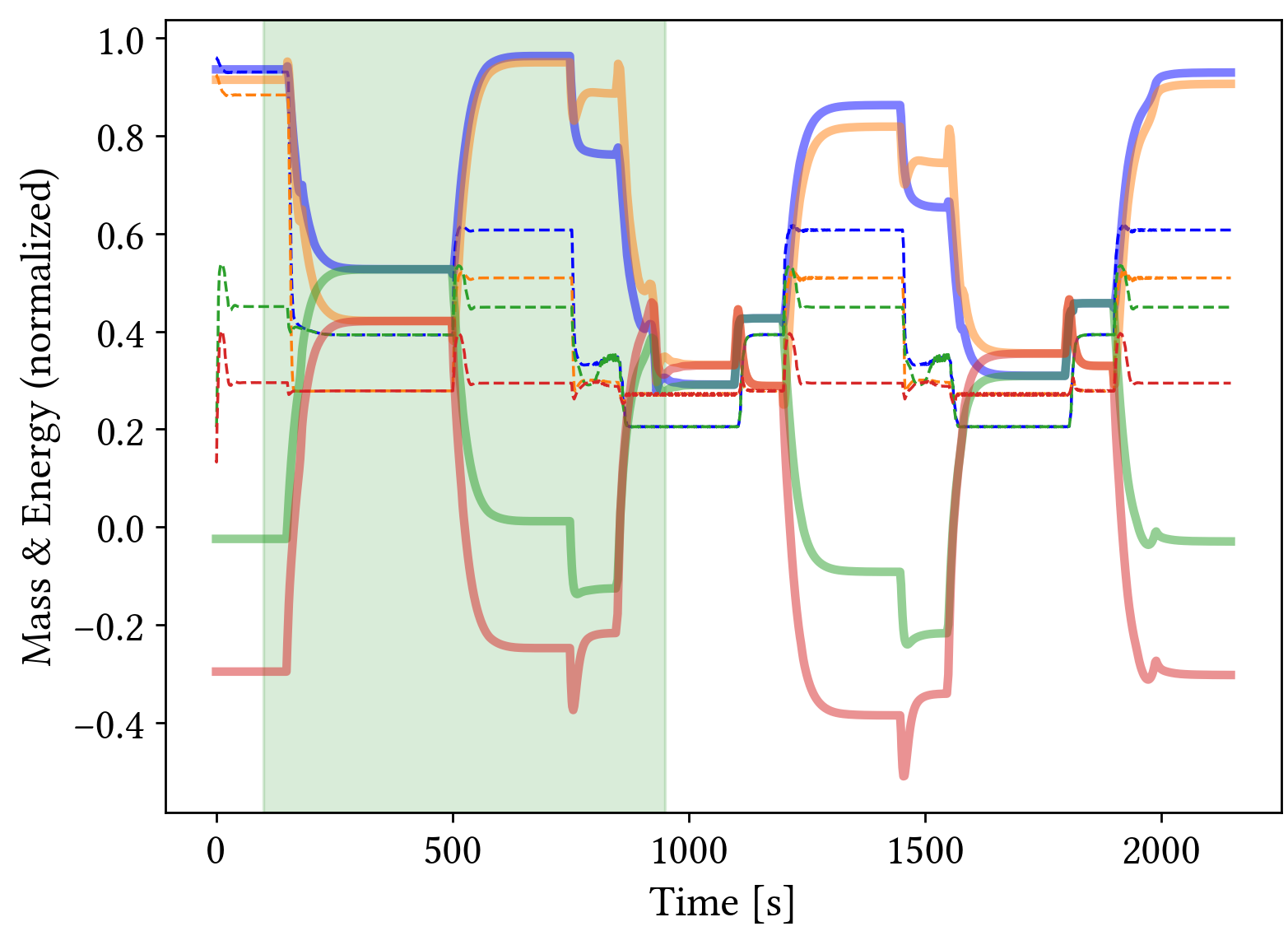}
    (b)\includegraphics[height=0.26\linewidth]{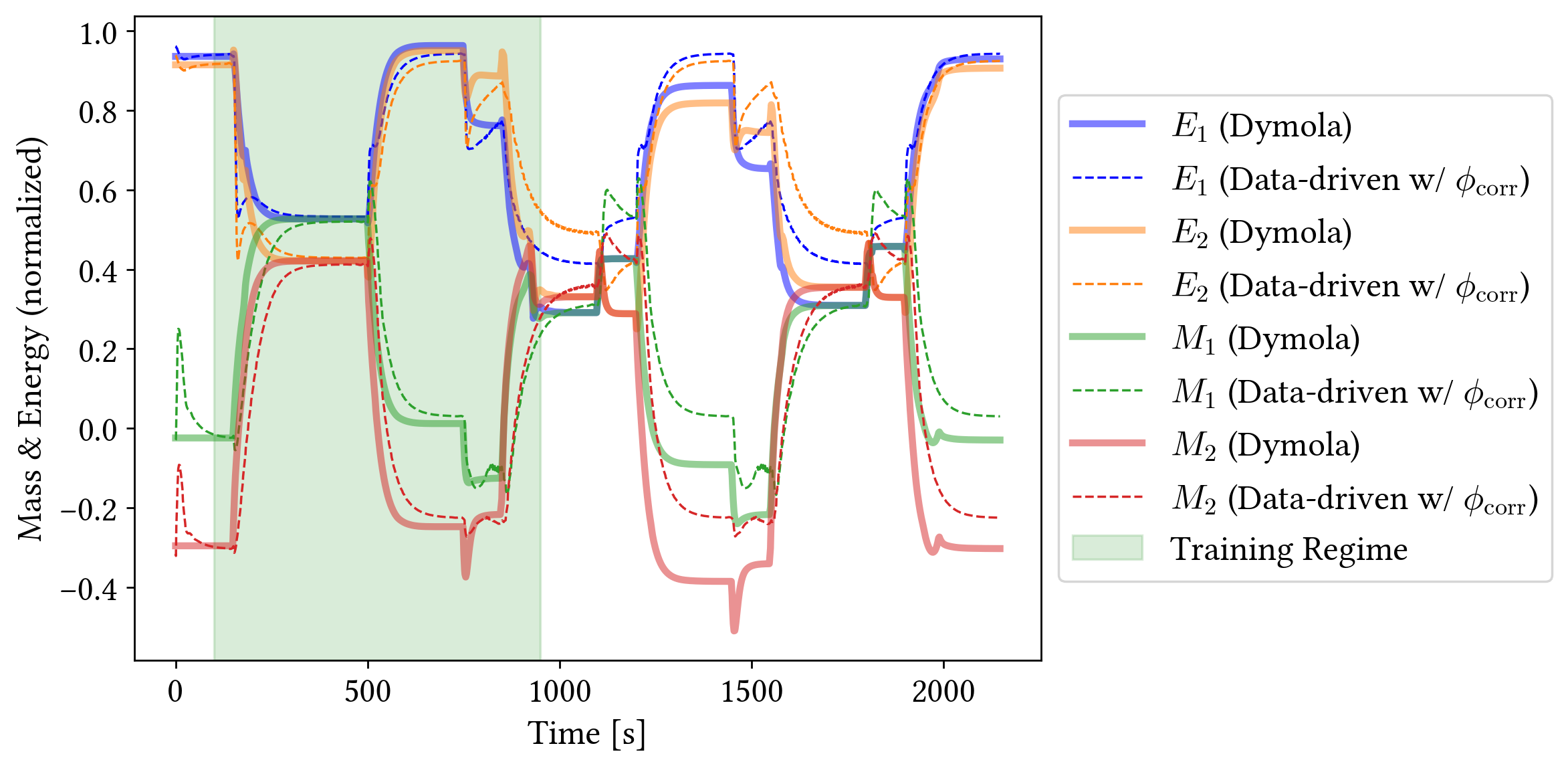}
    \caption{Comparison of normalized mass-energy (ME) terms for the dual-compressor system: (a) before and (b) after applying the corrector network. The corrector network reduces systematic deviations between predicted and reference ME values, with improvements observed across both energy terms ($E_1$, $E_2$) and mass terms ($M_1$, $M_2$).}
    \label{fig:me_norm_corrnn}
\end{figure}

\begin{figure}[htbp]
    \centering
    {\small Before applying corrector NN\hspace{70pt}After applying corrector NN\quad\quad\quad\quad\quad}\\
    (a)\includegraphics[height=0.33\linewidth]{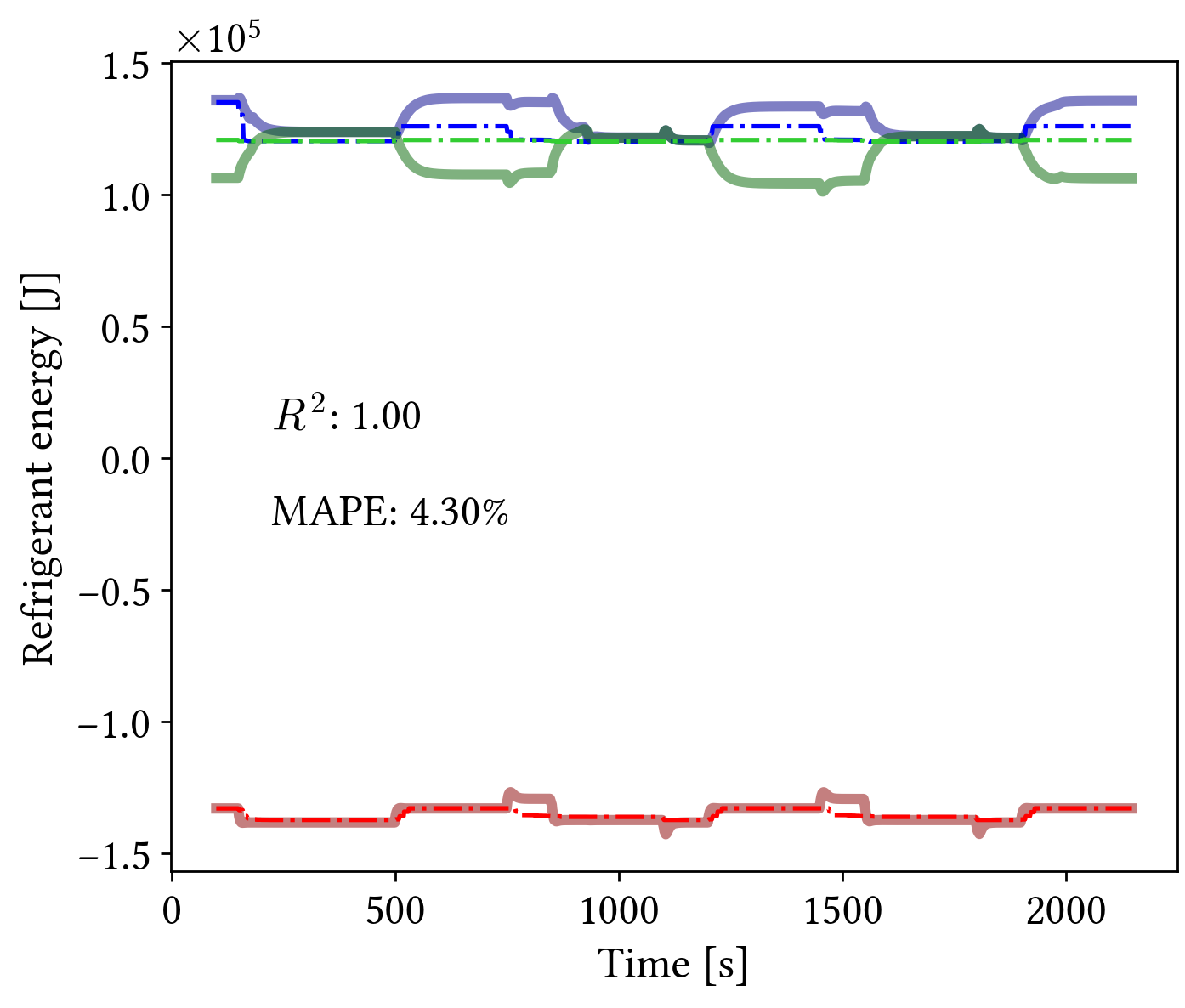}
    (b)\includegraphics[height=0.33\linewidth]{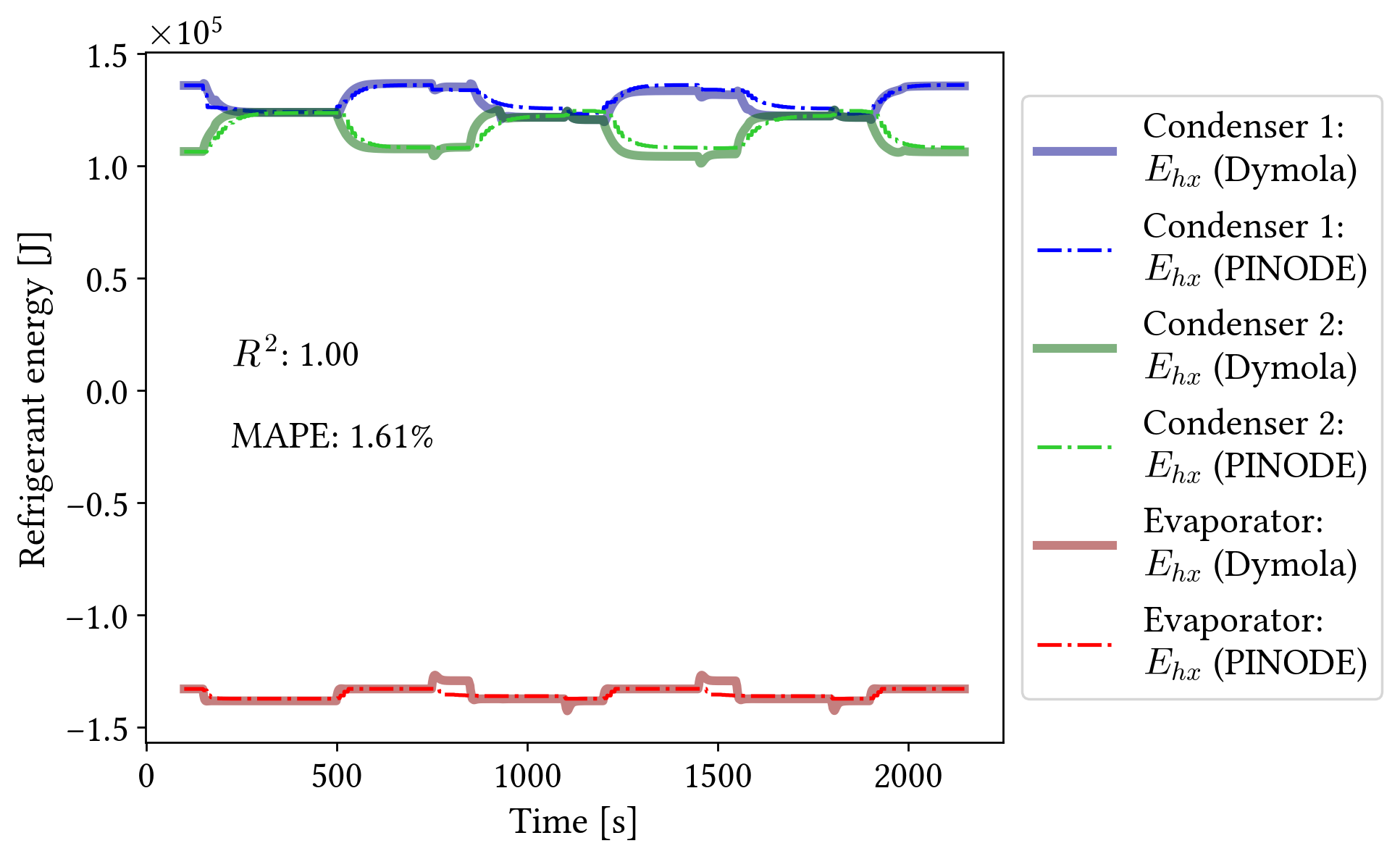}
    \caption{Refrigerant energy evolution in the dual-compressor HVAC system: (a) without and (b) with the corrector network. Improvements in local mass-energy variables translate to more consistent system-level energy predictions, with reduced discrepancies relative to the reference solution, particularly during transient regimes.}
    \label{fig:energy_corrnn}
\end{figure}

To further quantify the improvement, we examine the normalized mass-energy (ME) terms directly.
Figure~\ref{fig:me_norm_corrnn} shows the evolution of the four ME components ($E_1$, $E_2$, $M_1$, $M_2$) before and after applying the corrector network.
In the baseline case (Figure~\ref{fig:me_norm_corrnn}(a)), the predicted trajectories exhibit systematic deviations from the reference solution, most notably during transient periods where coupling effects and integration errors are more pronounced.
After applying the corrector network (Figure~\ref{fig:me_norm_corrnn}(b)), these discrepancies are significantly reduced, with the predicted trajectories more closely aligned with the reference across all components.
The improvement is particularly evident in the reduction of transient overshoots and bias, indicating that the corrector effectively compensates for accumulated system-level errors arising from component coupling and numerical integration.
While Figure~\ref{fig:me_norm_corrnn} focuses on component-level mass-energy variables, we further examine how these improvements propagate to the system-level energy evolution.
Figure~\ref{fig:energy_corrnn} shows the refrigerant energy evolution of the overall system.
In the baseline case (Figure~\ref{fig:energy_corrnn}(a)), noticeable deviations from the reference persist, particularly during transient regimes.
After applying the corrector network (Figure~\ref{fig:energy_corrnn}(b)), the predicted energy trajectory more closely tracks the reference across the entire time horizon, with reduced transient discrepancies.
This behavior is consistent with the correction of local mass-energy states leading to improved global energy consistency.

\subsection{Generalizability in Dual-Compressor Systems\label{sec:dualcomp_results}}

\subsubsection{Bayesian Optimization}

To systematically identify optimal solver parameters, we apply Bayesian optimization to characterize the design space for solver tolerances across three different solver types: the algebraic (Powell hybrid) solver, the DAE-IDA solver, and the DAE-DASSL solver.
The objective function balances simulation accuracy (measured by overall MAPE) and computational efficiency (simulation time), allowing us to find Pareto-optimal configurations that trade off these competing objectives.

\begin{figure}[htbp]
    \centering
    \includegraphics[width=0.5\linewidth]{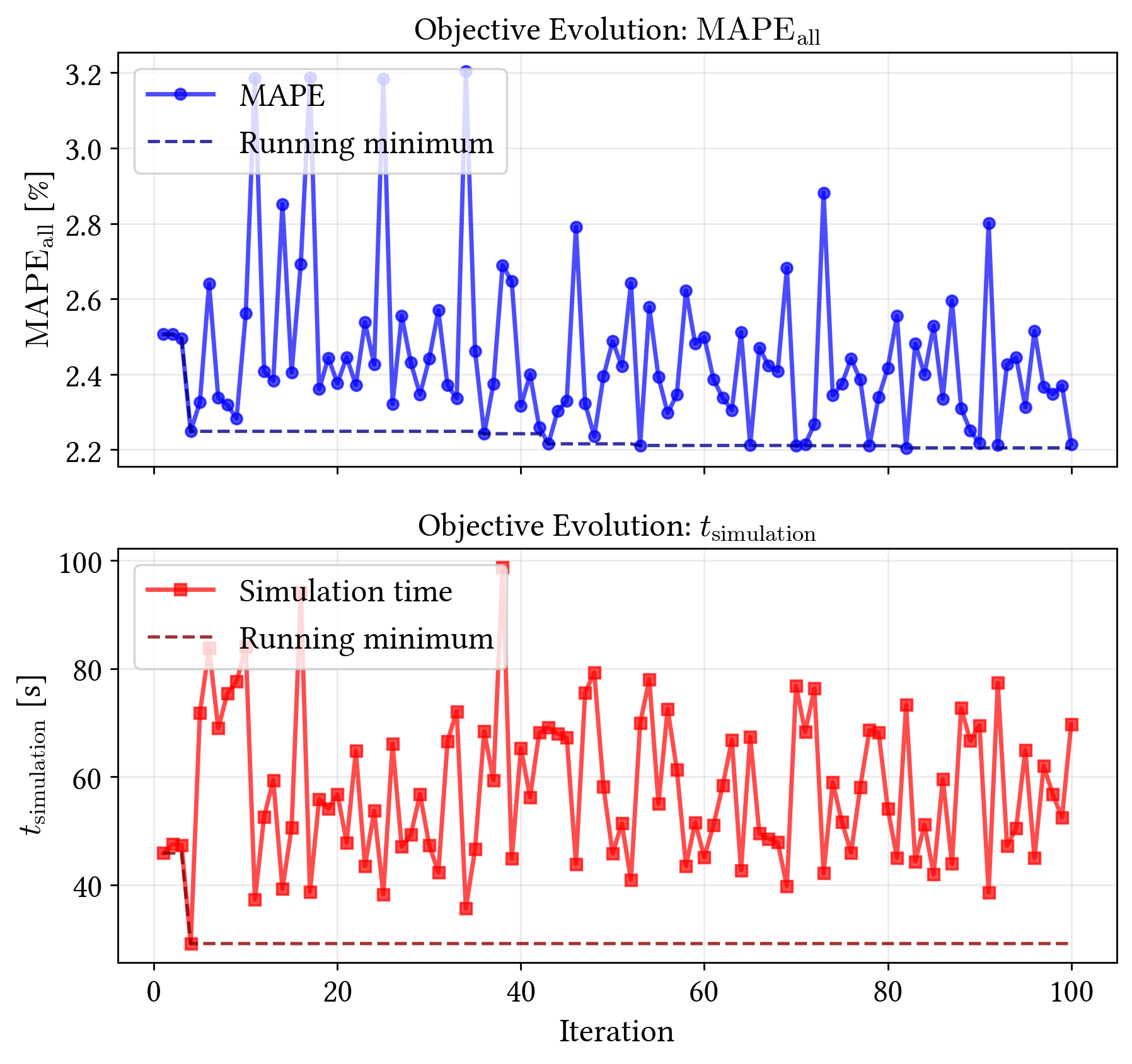}
    \caption{Evolution of the objective function during Bayesian optimization for the algebraic solver. The objective function decreases over 100 iterations, demonstrating convergence to improved parameter configurations that balance accuracy and computational efficiency.}
    \label{fig:bo_obj_algebraic}
\end{figure}

\begin{figure}[htbp]
    \centering
    \includegraphics[width=0.75\linewidth]{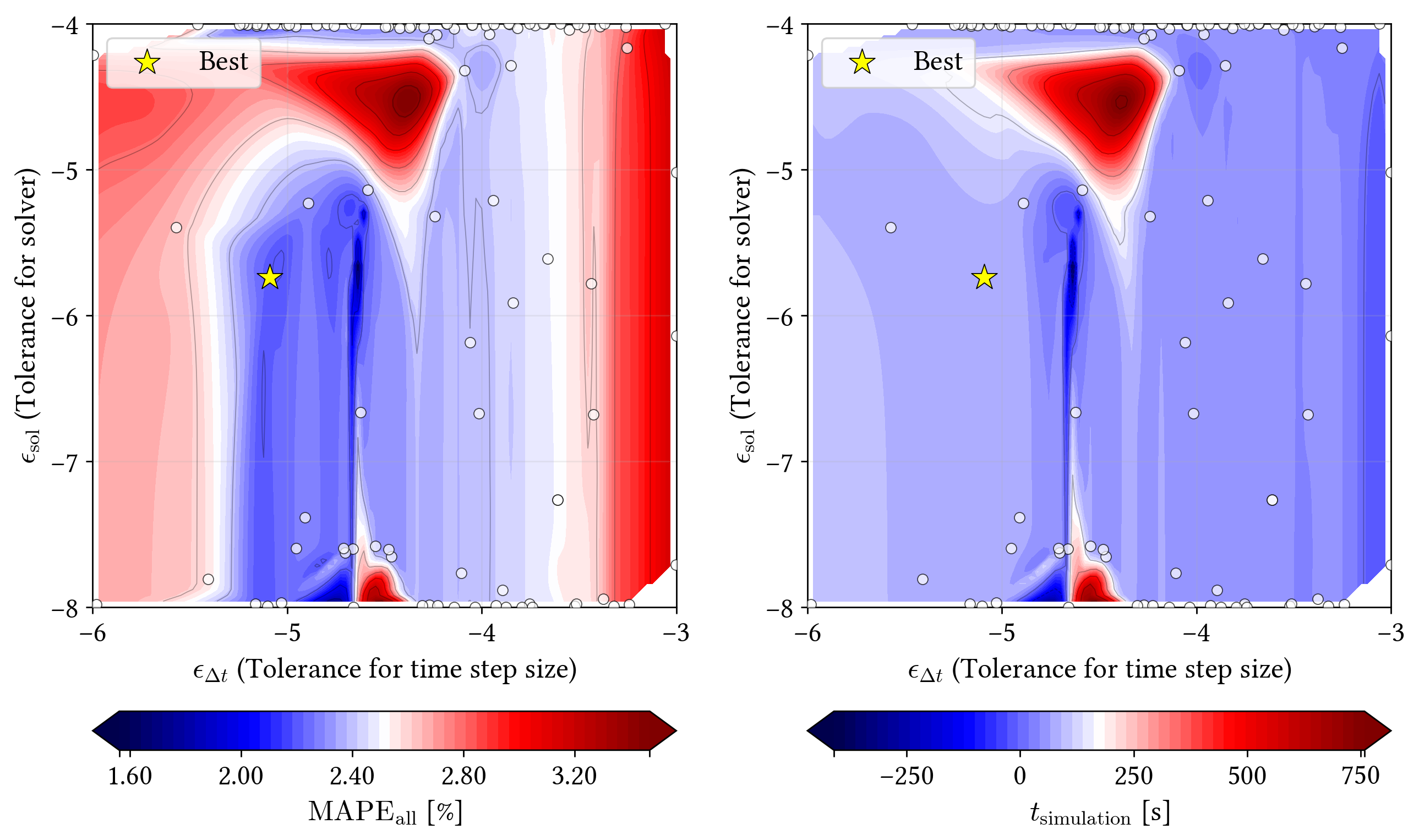}
    \caption{Design space characterization in Bayesian optimization for the algebraic solver. The contour plots show the sensitivity of (left) overall MAPE and (right) simulation time to different combinations of $\epsilon_{\Delta t}$ and $\epsilon_\text{soln}$. The optimal configuration (marked with a yellow star) achieves a balance between accuracy and efficiency.}
    \label{fig:bo_design_algebraic}
\end{figure}

For the algebraic solver, we optimize two key parameters: the time step tolerance ($\epsilon_{\Delta t}$) and the solution tolerance ($\epsilon_\text{soln}$).
Figure~\ref{fig:bo_obj_algebraic} shows the evolution of the objective function during the optimization process, demonstrating convergence to improved solutions over 100 iterations.
The objective function decreases monotonically, indicating that the Bayesian optimization successfully explores the design space and identifies progressively better parameter combinations.
The corresponding design space contours in Figure~\ref{fig:bo_design_algebraic} reveal the sensitivity of both MAPE and simulation time to different tolerance combinations, identifying regions of optimal performance.
The contour plots show that tighter tolerances generally improve accuracy but increase computational cost, creating a clear trade-off that the optimization must navigate.

\begin{figure}[htbp]
    \centering
    \includegraphics[width=0.5\linewidth]{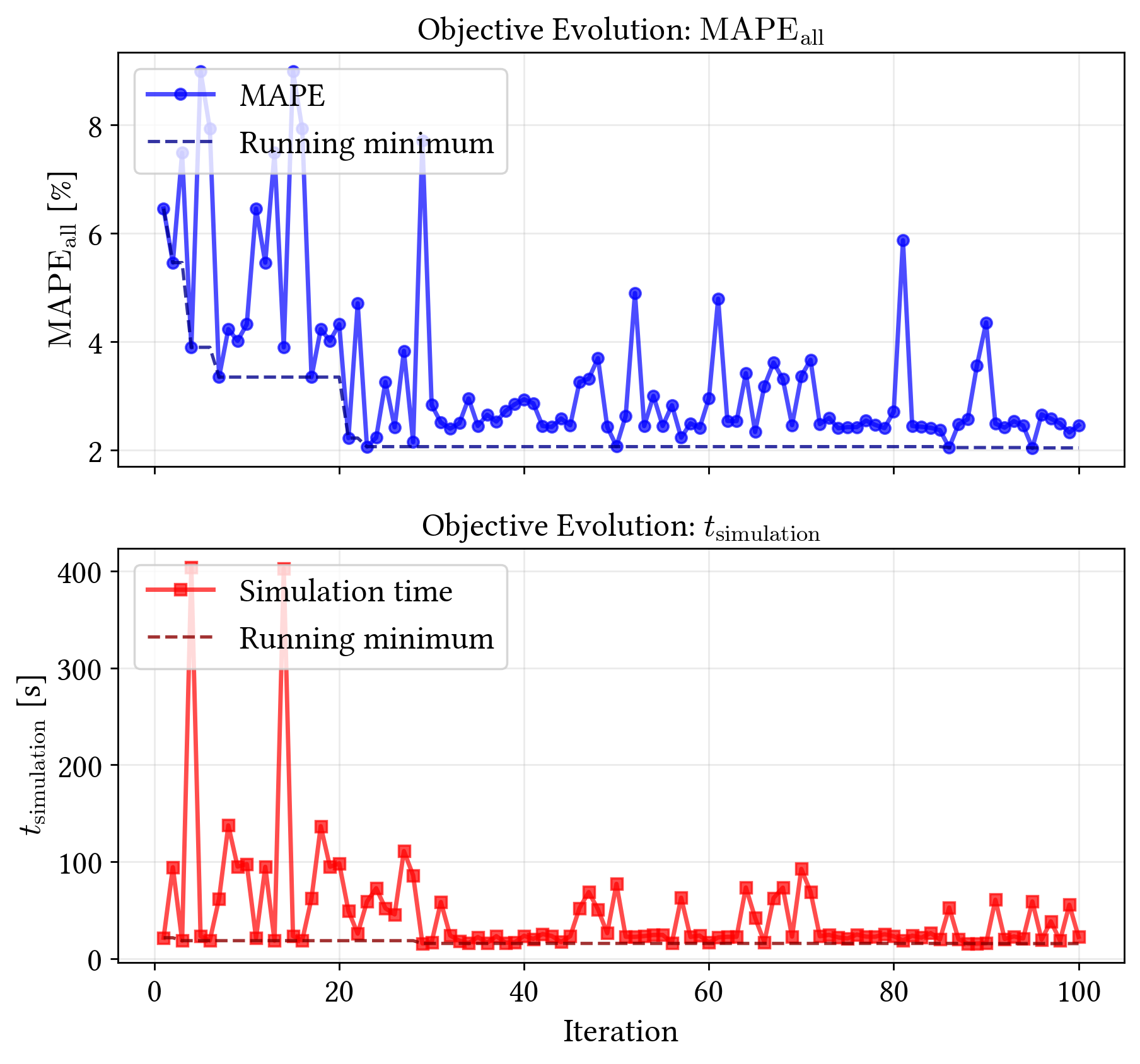}
    \caption{Evolution of the objective function during Bayesian optimization for the DAE-IDA solver. The optimization explores a 5-dimensional parameter space to find optimal configurations balancing accuracy and computational efficiency.}
    \label{fig:bo_obj_ida}
\end{figure}

\begin{figure}[htbp]
    \centering
    \includegraphics[width=0.75\linewidth]{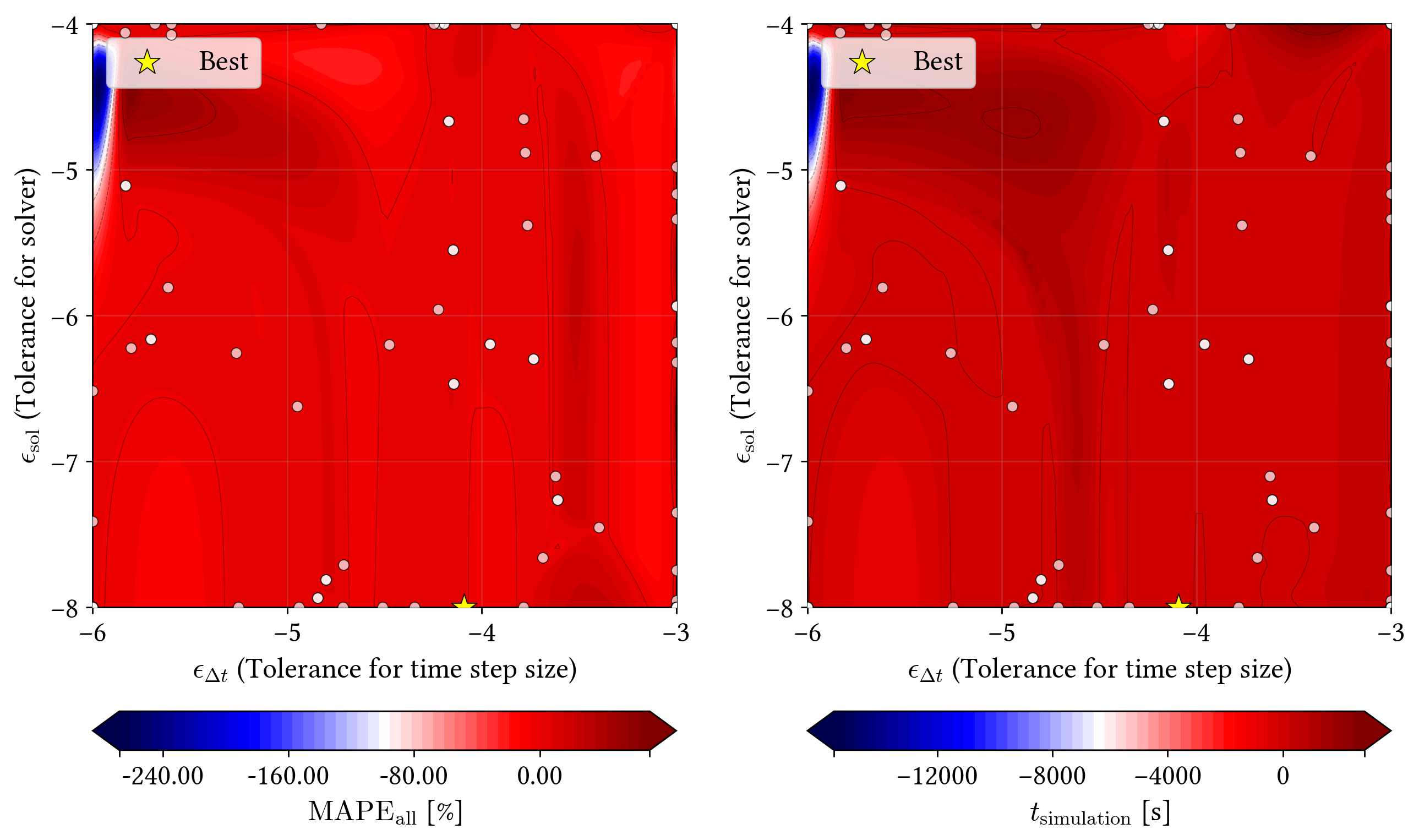}
    \caption{Design space characterization in Bayesian optimization for the DAE-IDA solver. The contour plots show the sensitivity of (left) overall MAPE and (right) simulation time to different combinations of $\epsilon_{\Delta t}$ and $\epsilon_\text{soln}$, with other parameters fixed at their optimal values.}
    \label{fig:bo_design_ida}
\end{figure}

For the two DAE solvers, we conduct a high-dimensional design space search since these solvers inherently require more parameters than the algebraic solver.
The optimization results are shown in Figures~\ref{fig:bo_obj_ida} and~\ref{fig:bo_design_ida}, revealing distinct optimal regions compared to the algebraic solver.
Unlike the algebraic solver, which optimizes over a 2-dimensional parameter space ($\epsilon_{\Delta t}$, $\epsilon_\text{soln}$), the DAE-IDA solver explores a 5-dimensional design space ($\epsilon_{\Delta t}$, $\epsilon_\text{soln}$, $h_\text{max}$, $h_\text{min}$, $\Delta t_\text{out}^\text{IDA}$), as described in Section~\ref{sec:system_solver}.
The objective function evolution (Figure~\ref{fig:bo_obj_ida}) demonstrates superior convergence behavior compared to the algebraic solver, with the Bayesian optimization algorithm efficiently navigating the higher-dimensional space to identify optimal configurations that balance accuracy and computational efficiency.
The 2D design space projections (shown for the two primary tolerances) reveal different sensitivity patterns, indicating that the DAE-IDA solver has distinct optimal operating characteristics that require careful multi-dimensional tuning.

\begin{figure}[htbp]
    \centering
    \includegraphics[width=0.5\linewidth]{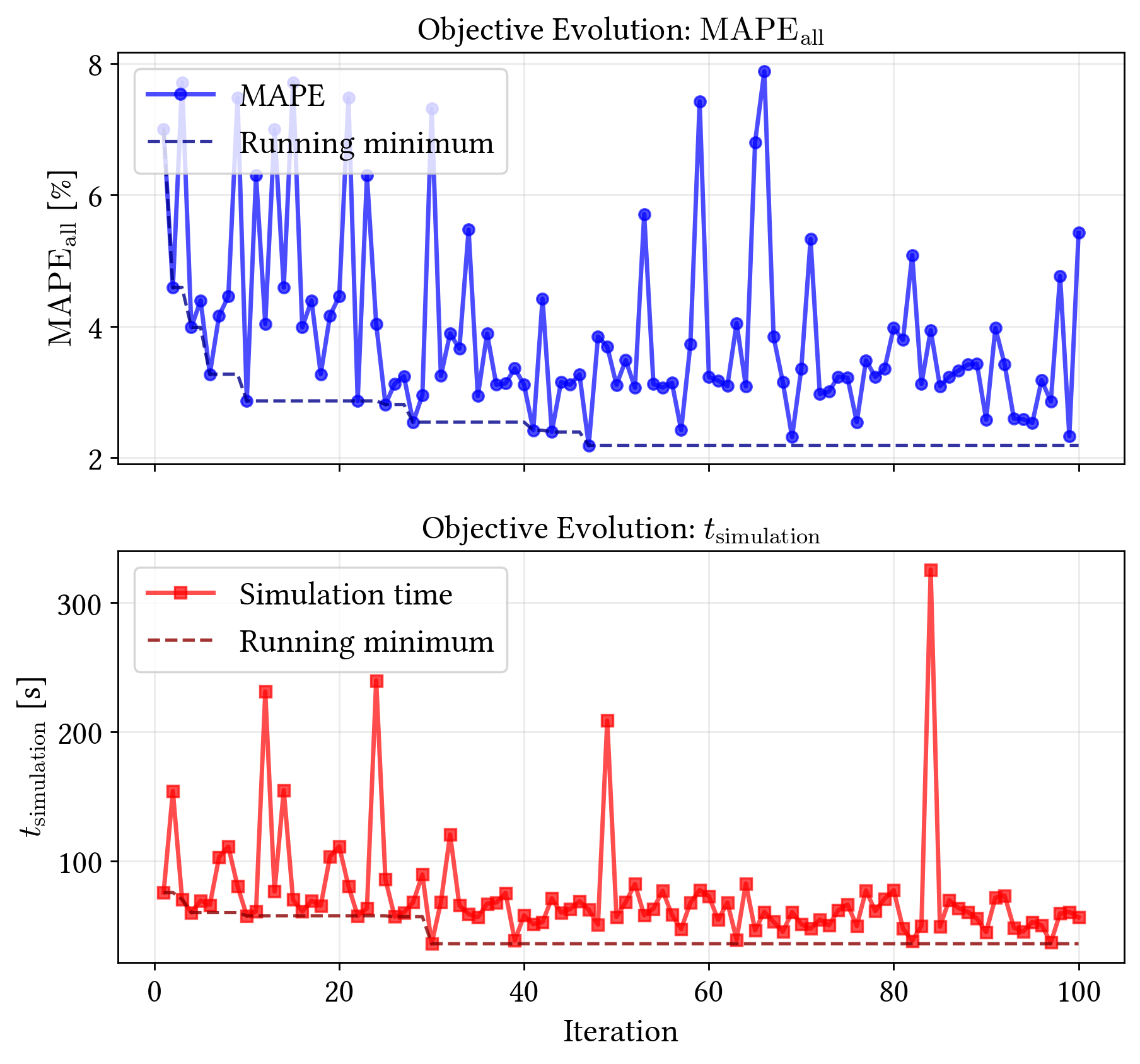}
    \caption{Evolution of the objective function during Bayesian optimization for the DAE-DASSL solver. The optimization explores a 6-dimensional parameter space to identify optimal configurations.}
    \label{fig:bo_obj_dassl}
\end{figure}

\begin{figure}[htbp]
    \centering
    \includegraphics[width=0.75\linewidth]{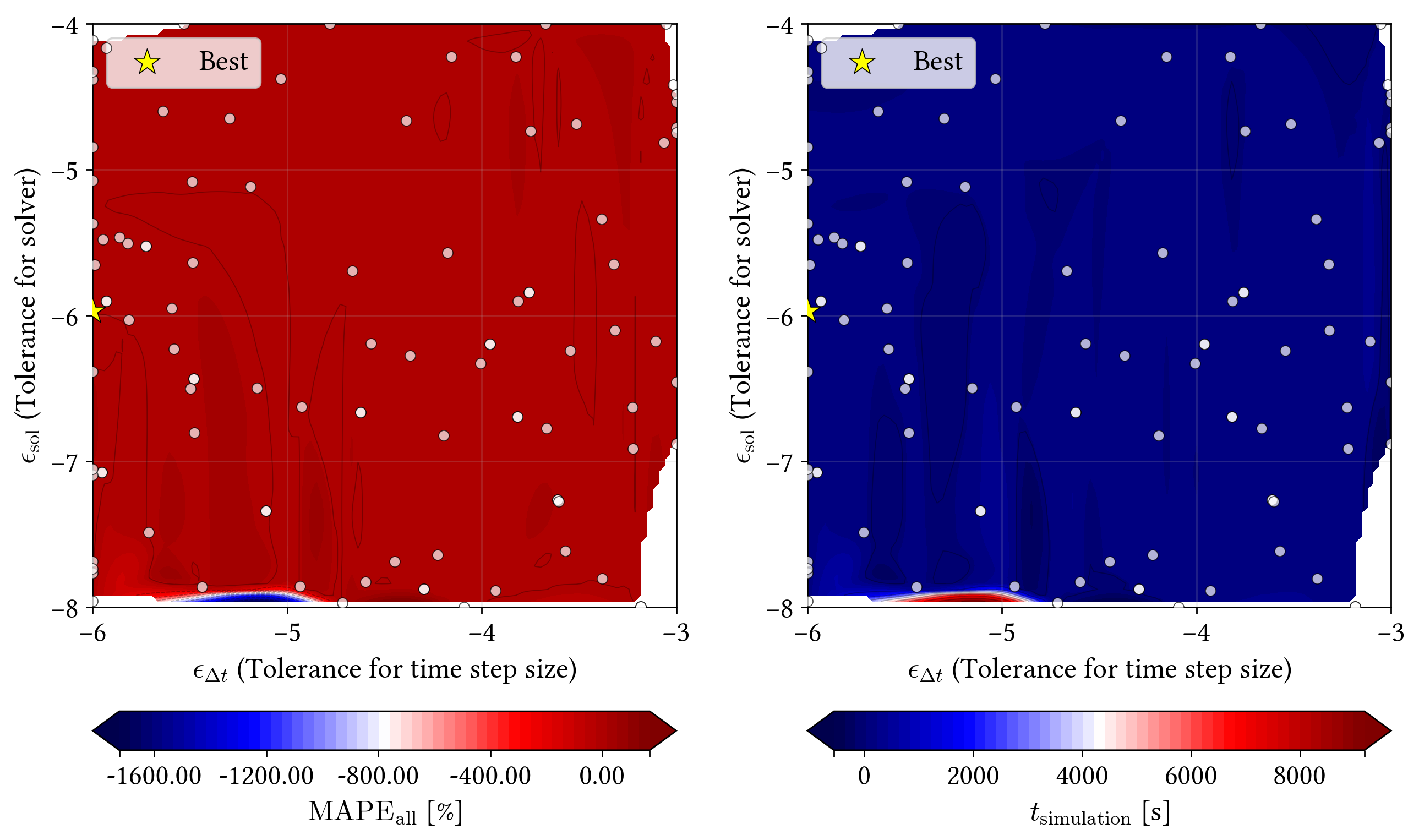}
    \caption{Design space characterization in Bayesian optimization for the DAE-DASSL solver. The contour plots show the sensitivity of (left) overall MAPE and (right) simulation time to different combinations of $\epsilon_{\Delta t}$ and $\epsilon_\text{soln}$, with other parameters fixed at their optimal values.}
    \label{fig:bo_design_dassl}
\end{figure}

Similarly, the DAE-DASSL solver optimization involves six parameters: $\epsilon_{\Delta t}$, $\epsilon_\text{soln}$, $h_\text{max}$, $h_\text{min}$, $\Delta t_\text{out,min}^\text{DASSL}$, and $N_\text{max}^\text{DASSL}$.
The optimization results are presented in Figures~\ref{fig:bo_obj_dassl} and~\ref{fig:bo_design_dassl}.
The design space exploration reveals different optimal parameter regions for DASSL compared to both the algebraic solver and the IDA solver, reflecting the unique numerical properties and adaptive time-stepping strategies of each solver.

\begin{figure}[htbp]
    \centering
    \includegraphics[width=0.5\linewidth]{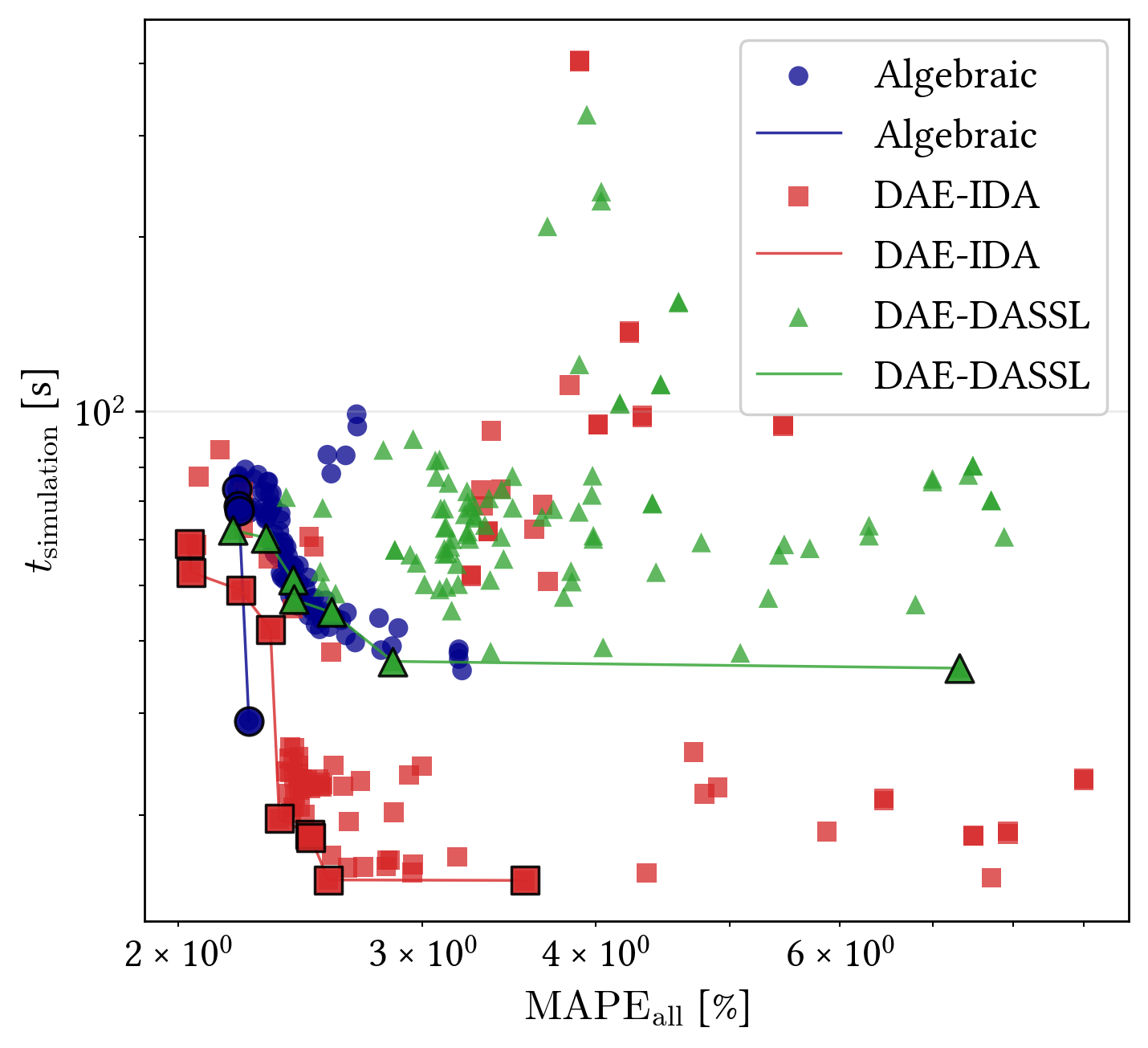}
    \caption{Comparison of Pareto fronts for the three solvers on a log-log scale. The DAE-IDA solver achieves the best overall trade-off between accuracy (MAPE) and computational efficiency (simulation time), with its Pareto front lying closest to the origin.}
    \label{fig:pareto_bayesian}
\end{figure}

To compare the overall performance across all three solvers, we examine the Pareto fronts in Figure~\ref{fig:pareto_bayesian}.
The Pareto front represents the set of non-dominated solutions, where no solution can improve one objective without worsening the other.
From the Pareto front plots, we observe that the DAE-IDA solver generally outperforms the other solvers, achieving better trade-offs between accuracy and computational efficiency across the design space.
The DAE-IDA solver's Pareto front lies closer to the origin (lower MAPE and lower simulation time), indicating superior overall performance.
The algebraic solver shows competitive performance but with a less favorable Pareto front, while the DAE-DASSL solver exhibits a wider spread of solutions, suggesting more variability in its performance characteristics.

\subsubsection{Simulation Results}

\begin{figure}[htbp]
    \centering\small\hspace{35pt}Algebraic solver\\
    (a)\includegraphics[width=.39\linewidth]{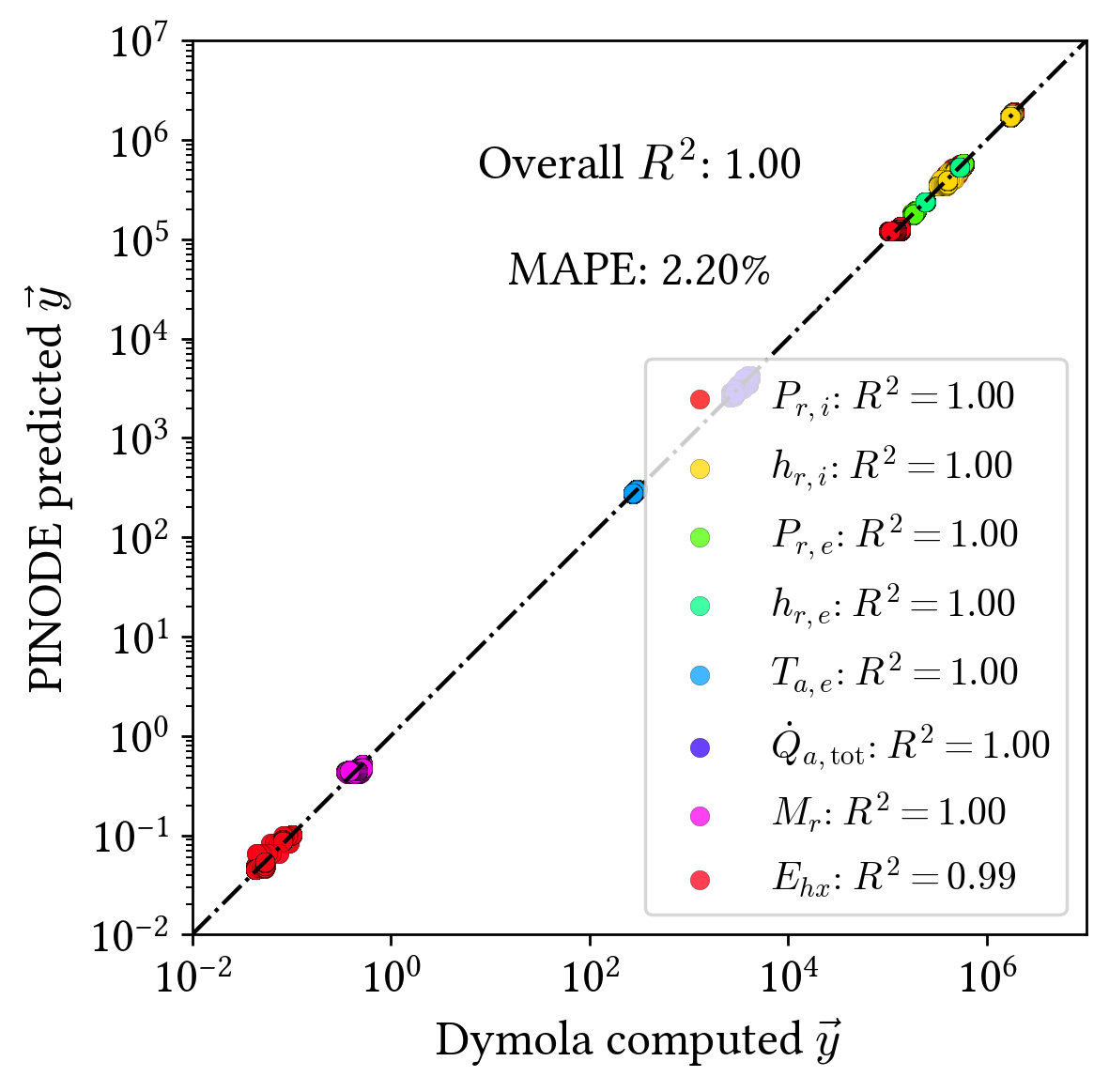}\\
    \centering\small\hspace{50pt}DAE-IDA solver\hspace{120pt}DAE-DASSL solver\\
    (b)\includegraphics[width=.39\linewidth]{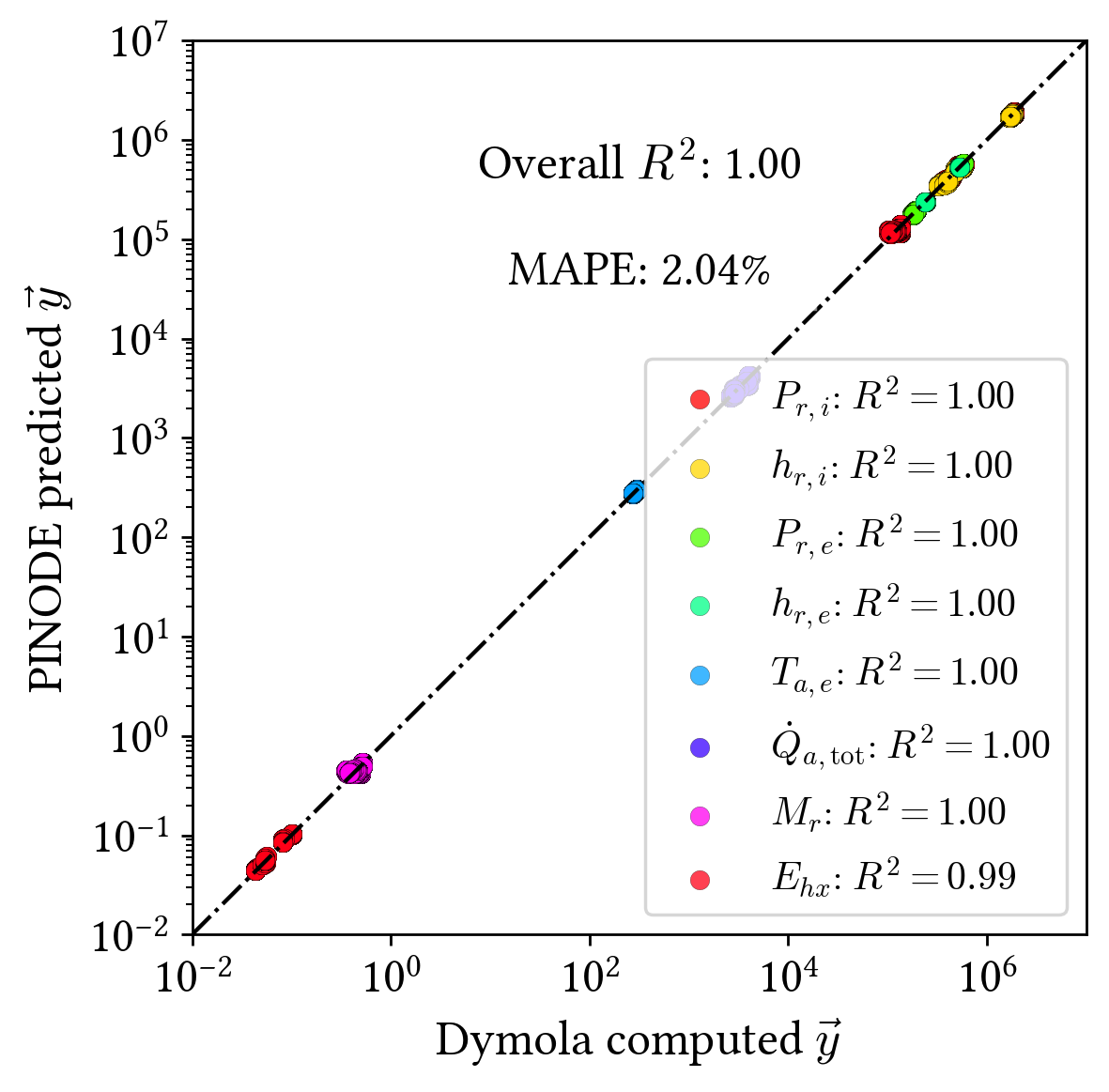}
    (c)\includegraphics[width=.39\linewidth]{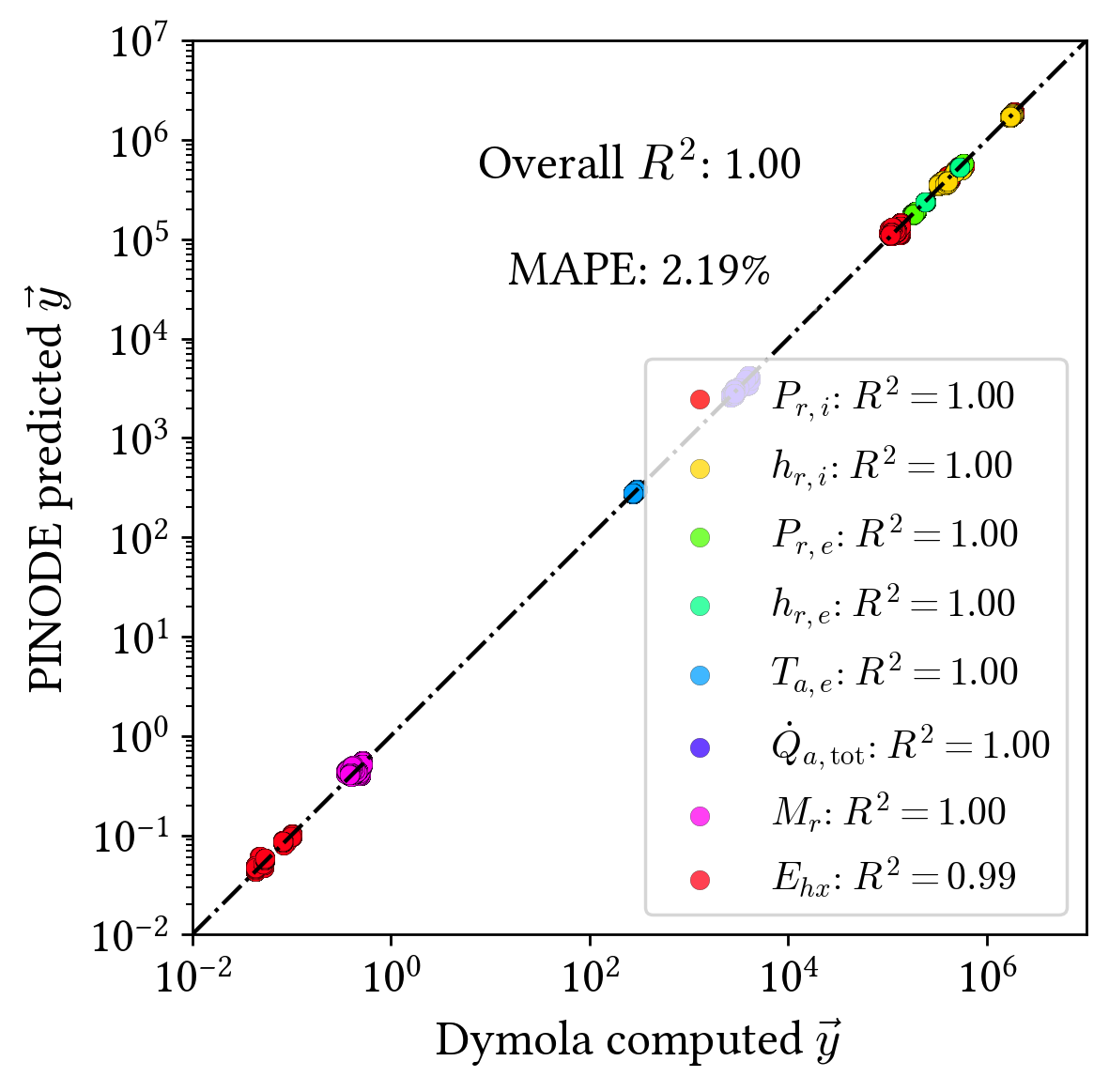}
    \caption{Comparison of parity plots for (a) algebraic solver, (b) DAE-IDA solver, and (c) DAE-DASSL solver using optimal parameters identified through Bayesian optimization. All three solvers demonstrate excellent agreement with the reference Dymola simulation, with the DAE-IDA solver showing the tightest clustering around the diagonal.}
    \label{fig:parity_system}
\end{figure}

Using the optimal parameters identified through Bayesian optimization, we evaluate the performance of each solver on the dual-compressor system.
Figure~\ref{fig:parity_system} compares the overall prediction accuracy for all three solvers, demonstrating that both DAE solvers achieve excellent agreement with the reference Dymola simulation, with the DAE-IDA solver showing particularly strong performance.
All three solvers produce predictions that cluster closely around the diagonal, indicating high accuracy, but the DAE-IDA solver shows the tightest clustering and best overall agreement with the reference data.

The quantitative performance metrics are summarized in Tables~\ref{tab:solver_best_overall} and~\ref{tab:solver_best_time}.
Table~\ref{tab:solver_best_overall} shows the best overall objective (weighted combination of MAPE and simulation time), where the DAE-IDA solver achieves the lowest MAPE (2.04\%) and fastest simulation time (58.91 s) simultaneously, representing the best balance between accuracy and efficiency.
The algebraic solver achieves a slightly higher MAPE (2.20\%) with a longer simulation time (73.33 s), while the DAE-DASSL solver shows comparable accuracy (2.19\% MAPE) but slightly better efficiency (62.30 s) than the algebraic solver.
Table~\ref{tab:solver_best_time} focuses on the fastest simulation times, where the DAE-IDA solver again excels with a simulation time of 15.43 s, though with a slightly higher MAPE of 3.56\%.
This demonstrates the DAE-IDA solver's ability to achieve both high accuracy and high computational efficiency, depending on the optimization objective.






\begin{table}[htbp]
\centering
\caption{Best Overall Objective (Weighted Combination of MAPE and Simulation Time)}
\label{tab:solver_best_overall}
\begin{tabular}{lccc}
\toprule
 & \textbf{Algebraic} & \textbf{DAE-IDA} & \textbf{DAE-DASSL} \\
\midrule
$\text{MAPE}_\text{all}$ [\%] & 2.2045\% & \textcolor{green!60!black}{2.0366\%} & 2.1888\% \\
$t_\text{simulation}$ [s] & 73.33 & \textcolor{green!60!black}{58.91} & 62.30 \\
\bottomrule
\end{tabular}
\end{table}


\begin{table}[htbp]
\centering
\caption{Best Simulation Time (Minimum Time with Corresponding MAPE)}
\label{tab:solver_best_time}
\begin{tabular}{lccc}
\toprule
 & \textbf{Algebraic} & \textbf{DAE-IDA} & \textbf{DAE-DASSL} \\
\midrule
$\text{MAPE}_\text{all}$ [\%] & \textcolor{green!60!black}{2.2486\%} & 3.5551\% & 7.3143\% \\
$t_\text{simulation}$ [s] & 29.14 & \textcolor{green!60!black}{15.43} & 35.93 \\
\bottomrule
\end{tabular}
\end{table}

The computational speedups achieved by our framework are substantial when compared to high-fidelity simulators.
The DAE-IDA solver achieves a simulation time of 15.43~s, representing approximately a $9.5\times$ speedup compared to the Dymola-based simulator reported in~\cite{Ma2024}, which requires 146.8~s for similar HVAC system simulations.
The algebraic solver achieves a $5.0\times$ speedup (29.14~s), while the DAE-DASSL solver achieves a $4.1\times$ speedup (35.93~s).
These $4$--$9\times$ computational speedups, combined with MAPE errors below $2.5\%$, demonstrate that our hybrid framework successfully bridges the accuracy--efficiency trade-off that has limited the practical deployment of physics-based HVAC simulators.

\begin{figure}[htbp]
    \centering
    \includegraphics[width=0.65\linewidth]{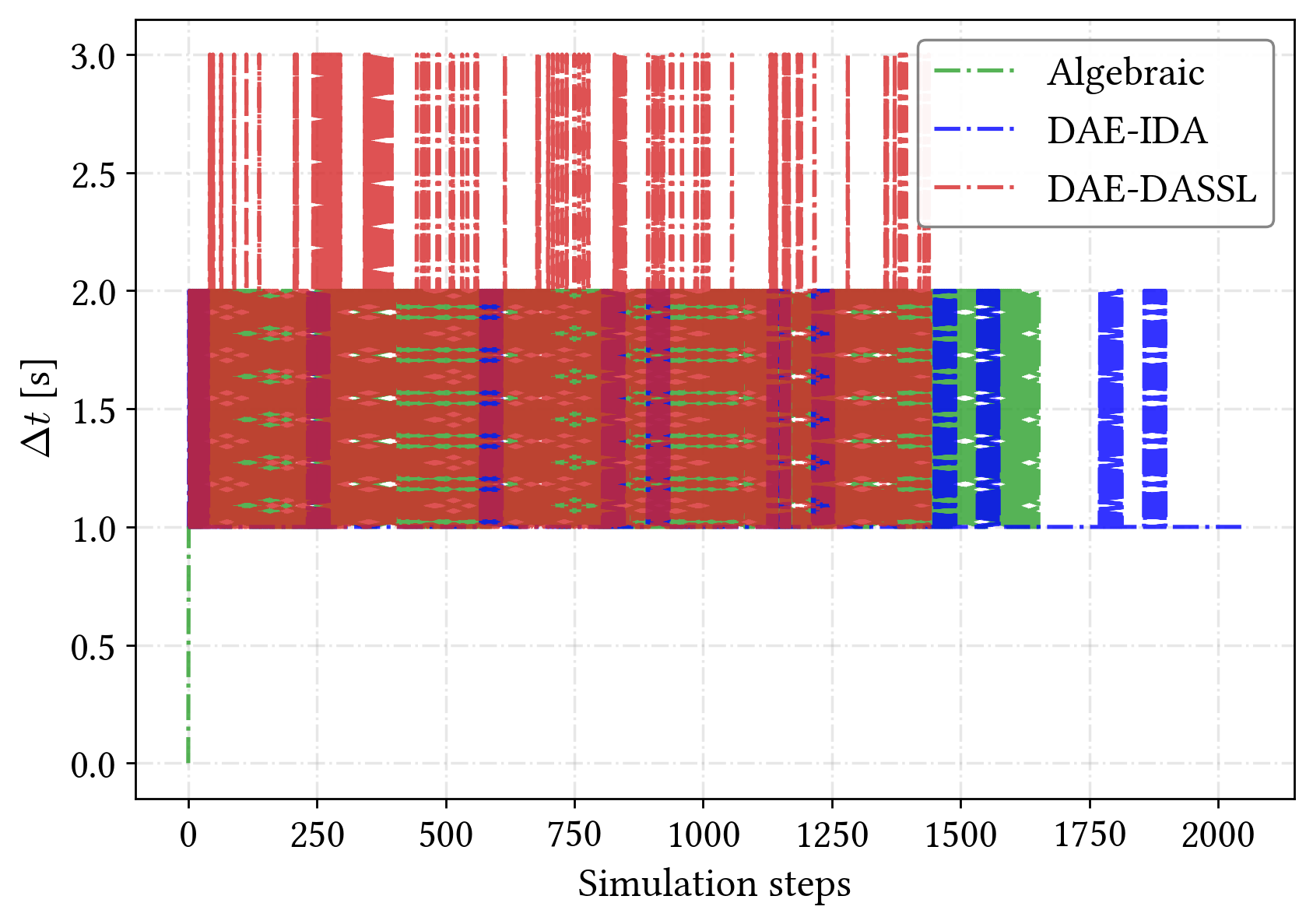}
    \caption{Comparison of optimal time step sizes ($\Delta t$) for the three solvers obtained through Bayesian optimization. The DAE-IDA solver achieves the smallest time steps, enabling higher temporal resolution, while the algebraic solver operates with larger time steps for computational efficiency.}
    \label{fig:delta_t_optimal}
\end{figure}

The optimal time step sizes selected by each solver reflect their different numerical strategies and adaptive time-stepping mechanisms.
Figure~\ref{fig:delta_t_optimal} compares the optimal $\Delta t$ evolution for the three solvers, revealing distinct temporal resolution patterns.
The DAE-IDA solver achieves smaller time steps on average, enabling higher temporal resolution and better capture of transient dynamics, which contributes to its superior accuracy.
The algebraic solver operates with larger time steps for computational efficiency, while the DAE-DASSL solver shows intermediate behavior, balancing between the two extremes.
The time step evolution patterns reflect each solver's internal error control mechanisms and their ability to adapt to local solution characteristics, as discussed in Section~\ref{sec:system_solver}.

For the algebraic solver, the RK45 adaptive stepping mechanism (see Section~\ref{sec:system_solver}) automatically adjusts the time step based on local truncation error estimates, reducing $\Delta t$ when dynamics are rapidly changing and increasing it during smooth transients.
The DAE solvers (IDA and DASSL) employ variable-order, variable-step BDF methods with adaptive error control, using weighted root-mean-square error norms to adaptively select both step size $h_n$ and BDF order $k$, enabling fine-grained temporal resolution control that is evident in the time step evolution patterns.
The observed behavior aligns with the theoretical framework for adaptive time stepping, where local error estimates drive automatic step size selection to maintain accuracy while maximizing computational efficiency.

\begin{figure}[htbp]
    \centering
    {\small Algebraic\quad\quad\quad\quad\quad}\\
    (a)\includegraphics[height=0.35\linewidth]{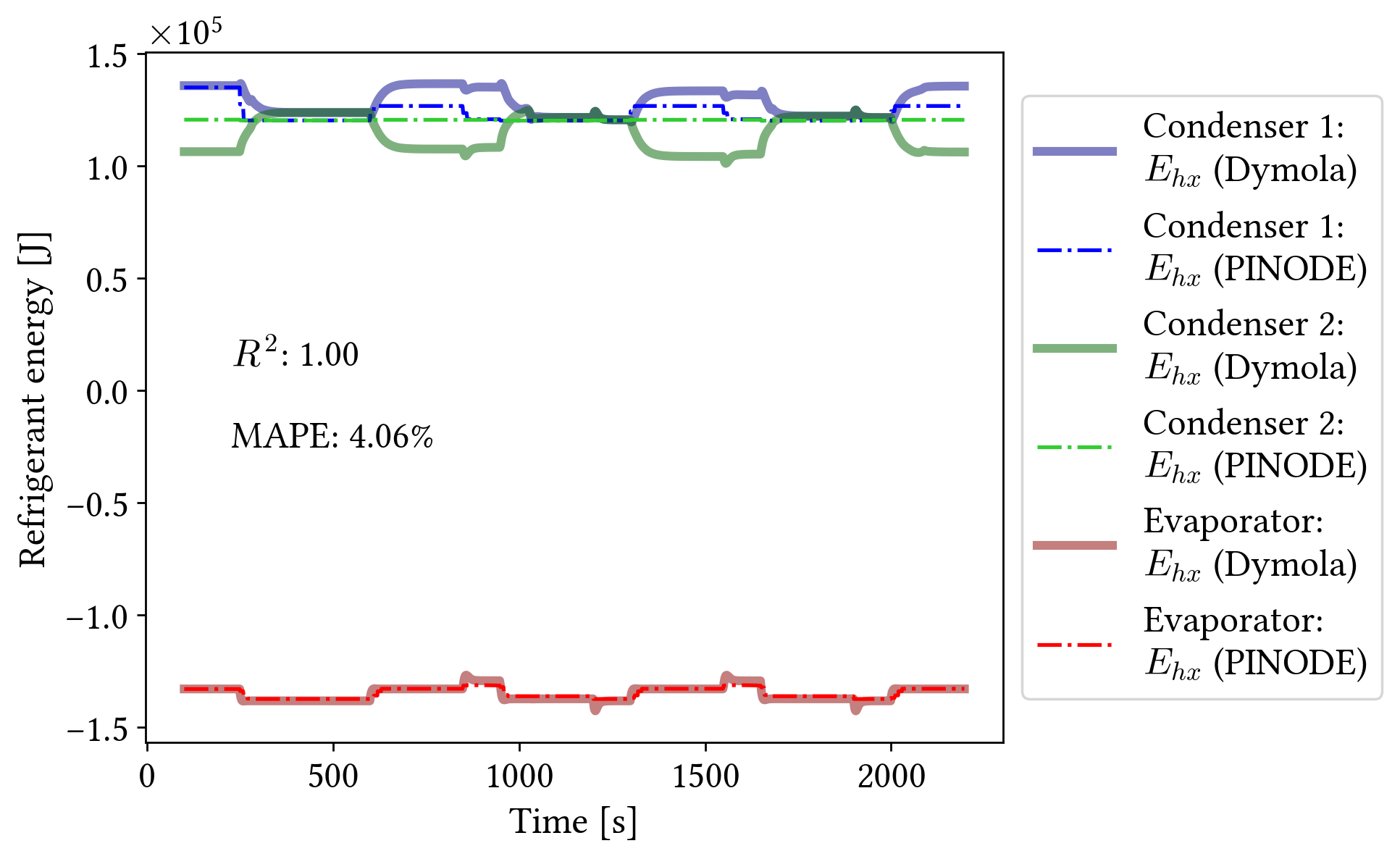}\\
    {\small\quad\quad\quad\quad\quad\quad DAE-IDA\hspace{150pt}DAE-DASSL}\\
    (b)\includegraphics[height=0.35\linewidth]{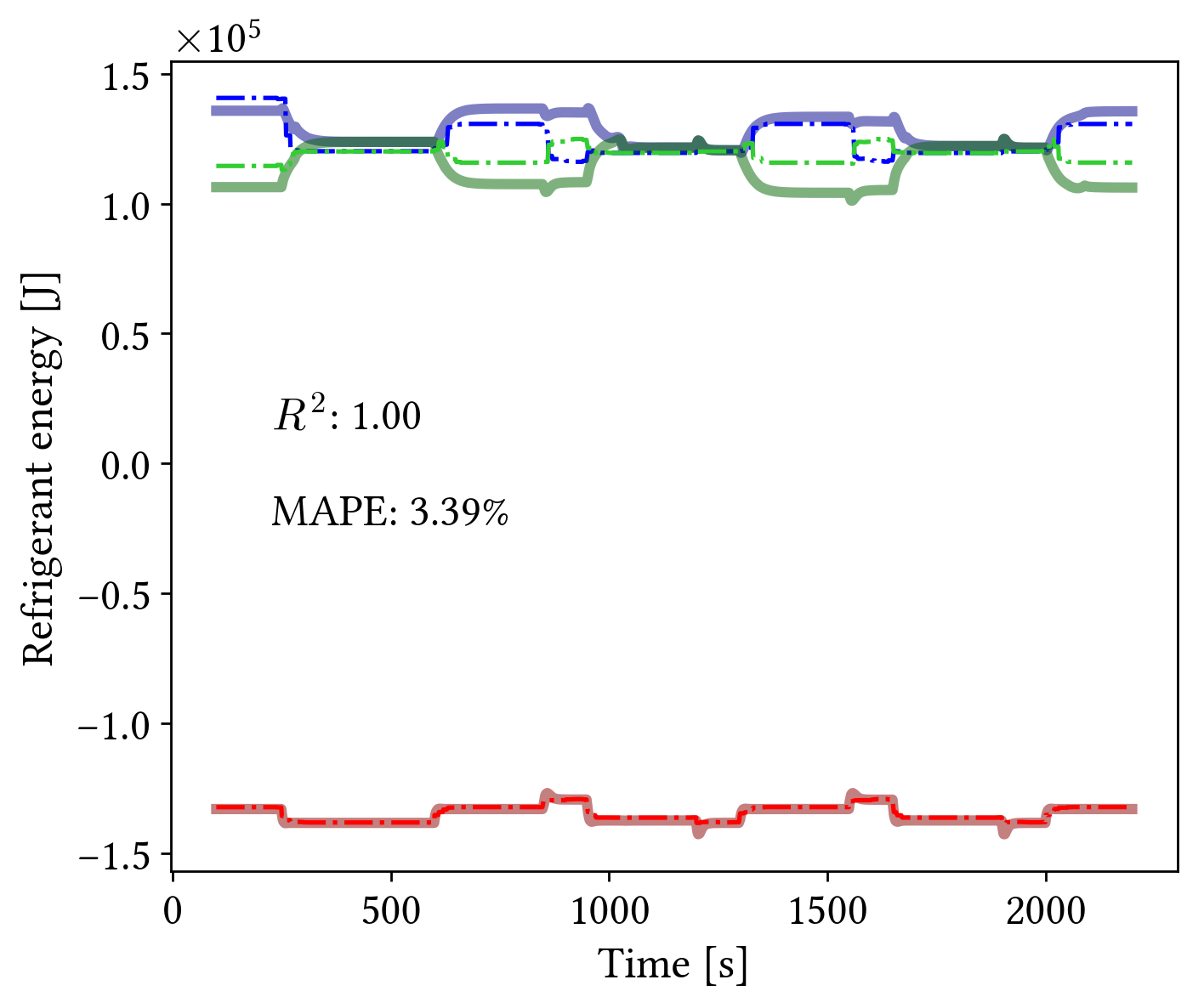}(c)\includegraphics[height=0.35\linewidth]{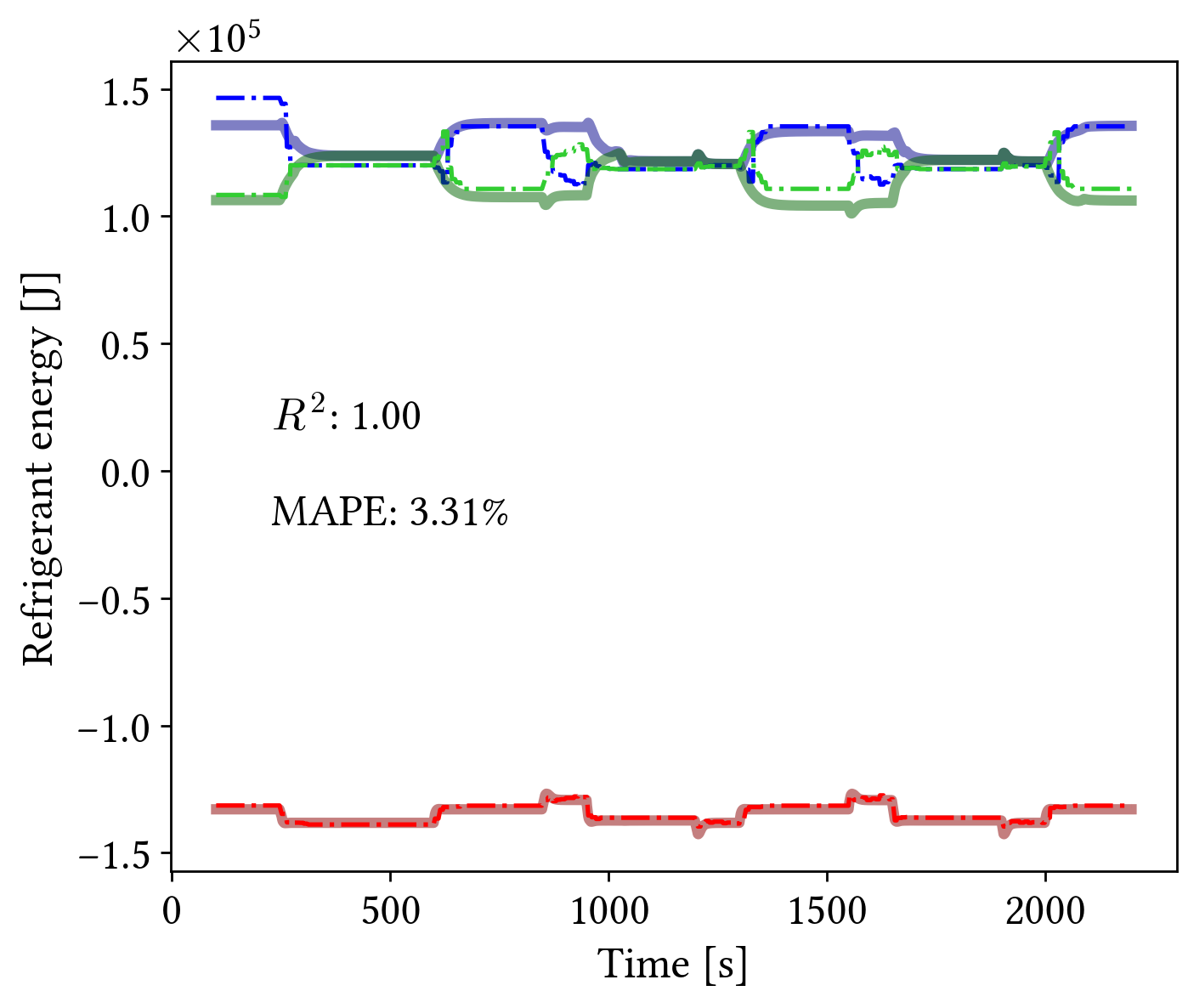}
    \caption{Refrigerant energy predictions using optimal solver parameters for (a) algebraic solver, (b) DAE-IDA solver, and (c) DAE-DASSL solver. All three solvers demonstrate good agreement with the reference Dymola simulation, with the DAE-IDA solver showing the best overall performance in terms of accuracy and computational efficiency.}
    \label{fig:energy_optimal}
\end{figure}

Finally, Figure~\ref{fig:energy_optimal} shows the refrigerant energy predictions using the optimal solver parameters for all three solvers.
All three solvers demonstrate good agreement with the reference Dymola simulation, validating the effectiveness of the Bayesian optimization approach in identifying high-performing parameter configurations.
The DAE-IDA solver consistently shows the best overall performance, achieving high accuracy while maintaining computational efficiency.
The energy predictions track the reference trajectory closely across all time periods, including rapid transients and steady-state operation, demonstrating the robustness of the optimized solver configurations.

Based on the dual-compressor simulations, the DAE-IDA solver emerges as a strong candidate for general-purpose HVAC dynamical simulations for several reasons. 
First, it consistently achieves favorable computational efficiency, often matching or exceeding the fastest configurations identified for the other solvers. 
Second, it maintains low prediction error across a wide range of operating conditions, with MAPE values comparable to or slightly better than those of the algebraic and DAE-DASSL solvers. 
Third, and most importantly, the DAE-IDA solver demonstrates superior performance on the Pareto front, indicating a more robust trade-off between accuracy and computational cost across the explored parameter space. 
This suggests that, beyond individual optimal points, the IDA solver provides more reliable performance under varying tolerance and stepping configurations. 

However, it is important to note that for smaller-scale systems, such as the dual-compressor case considered here, the performance gap between solvers is relatively modest. 
In particular, the algebraic solver remains competitive in terms of both runtime and accuracy when appropriately tuned, and in some cases can achieve comparable best-case performance. 
Therefore, while DAE-IDA offers better overall robustness and scalability, the choice of solver for small systems may still depend on specific optimization objectives, implementation simplicity, and parameter tuning considerations.

\clearpage

\section{Scalability of HVAC Systems\label{sec:scalability}}

\subsection{Scaling Up HVAC}

We scale up the HVAC systems by simultaneously increasing the number of compressor--condenser pairs ($n_c$) and valve--evaporator pairs ($n_v$), demonstrating the scalability of our algorithms.
This represents the first demonstration of such system scaling in the literature.
Compared with the previous work by Ma et al.~\cite{Ma2024}, we show that with our new system solver and effective data-driven components, large-scale systems can be solved efficiently, which was indeed a significant challenge for traditional physics-based solvers.

The large-scale system topology follows a parallel--merge architecture that generalizes the dual-compressor configuration:
\begin{enumerate}
\item \textbf{Parallel compression stage}: Each of the $n_c$ compressors discharges into its own dedicated condenser, operating in parallel.
\item \textbf{Liquid manifold}: All condenser outlets merge into a single liquid manifold node, creating a common high-pressure liquid reservoir.
\item \textbf{Parallel expansion stage}: The liquid manifold feeds $n_v$ parallel valve--evaporator pairs, each operating independently.
\item \textbf{Suction manifold}: All evaporator outlets merge into a single suction manifold node, creating a common low-pressure vapor reservoir.
\item \textbf{Compressor inlets}: The suction manifold feeds all $n_c$ compressor inlets, completing the cycle.
\end{enumerate}

This topology ensures that each compressor--condenser pair operates independently while sharing common liquid and suction manifolds, enabling efficient parallel operation and load distribution across multiple compressors.

\begin{figure}[htbp]
    \centering
    \includegraphics[width=0.65\linewidth]{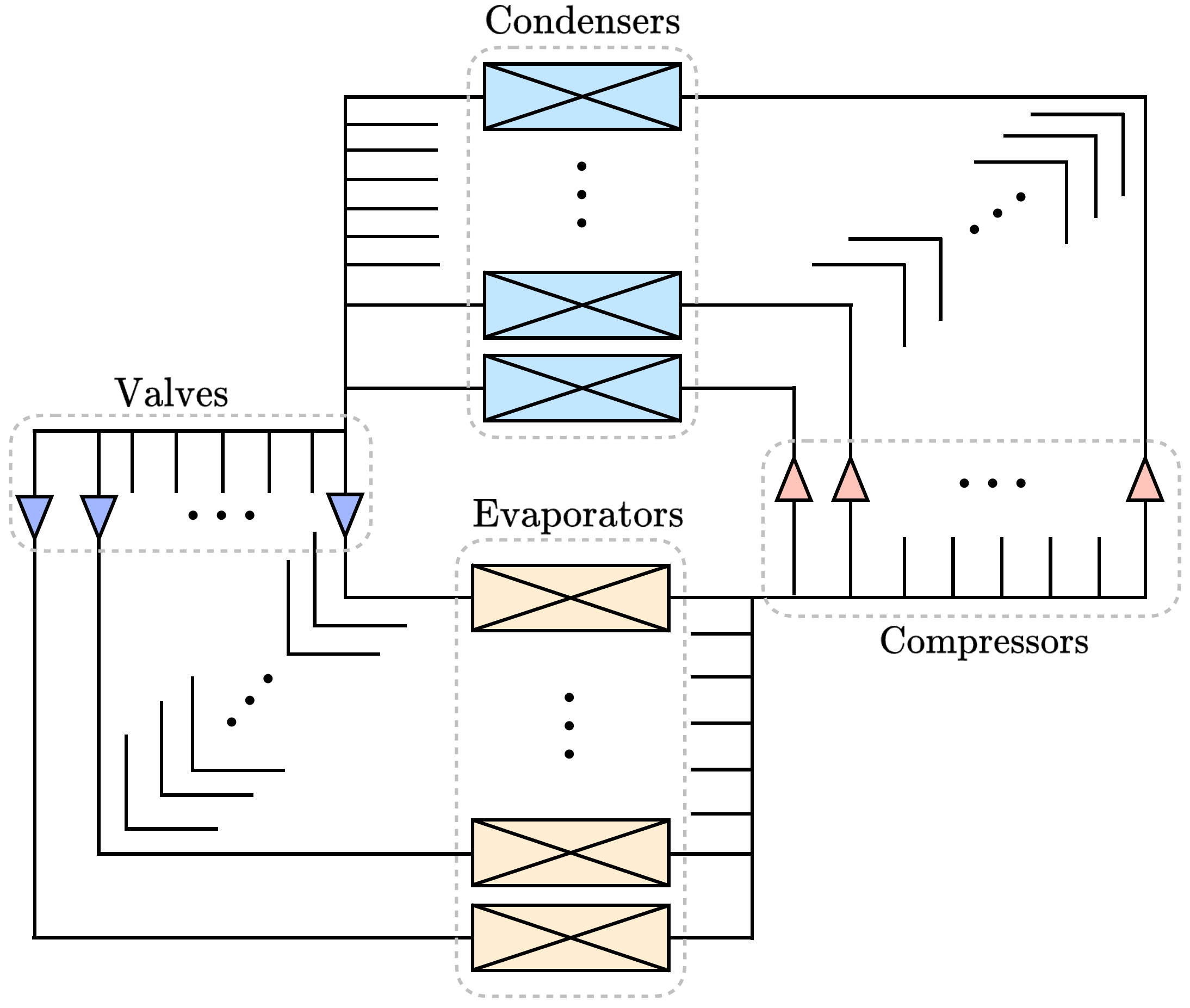}
    \caption{Scaling topology for large-scale HVAC systems. The left panel shows the dual-compressor base case ($n_c=n_v=2$). The right panel illustrates the parallel--merge architecture with \textbf{$n_c=n_v=16$} compressor--condenser and valve--evaporator branches as drawn in the schematic. Empirical runtime scaling for $n_c=n_v$ up to 16 is reported in Figure~\ref{fig:hvac_scaling}.}
    \label{fig:scaling_up_hvac}
\end{figure}

Figure~\ref{fig:hvac_scaling} reports solver cost for $n_c=n_v$ from 2 up to 16, using the same parallel--merge topology illustrated in Figure~\ref{fig:scaling_up_hvac}.

\begin{algorithm}[htbp]
\caption{Large-scale HVAC system construction}
\label{alg:large_scale_construction}
\begin{algorithmic}[1]
\Require Number of compressor--condenser pairs $n_c$, number of valve--evaporator pairs $n_v$%
\Ensure Configured system with $n_c$ compressors and $n_v$ evaporators%
\State \textbf{Initialize:} Create $n_c$ compressors $\mathcal{C} = \{C_1, \ldots, C_{n_c}\}$, $n_c$ condensers $\mathcal{H}_c = \{H_{c,1}, \ldots, H_{c,n_c}\}$, $n_v$ valves $\mathcal{V} = \{V_1, \ldots, V_{n_v}\}$, $n_v$ evaporators $\mathcal{H}_e = \{H_{e,1}, \ldots, H_{e,n_v}\}$%
\State \textbf{Connect parallel compression:}%
\For{$k = 1$ to $n_c$}
    \State Connect $C_k.\text{out} \rightarrow H_{c,k}.\text{in}$ \Comment{\textcolor{gray}{Compressor discharge to condenser inlet}}%
\EndFor
\State \textbf{Create liquid manifold:}%
\State Set $p_\text{liq} = H_{c,1}.\text{out}$ \Comment{\textcolor{gray}{Liquid manifold pressure node}}%
\For{$k = 2$ to $n_c$}
    \State Connect $H_{c,k}.\text{out} \rightarrow p_\text{liq}$ \Comment{\textcolor{gray}{Merge condenser outlets}}%
\EndFor
\State \textbf{Connect parallel expansion:}%
\For{$k = 1$ to $n_v$}
    \State Connect $p_\text{liq} \rightarrow V_k.\text{in}$ \Comment{\textcolor{gray}{Liquid manifold to valve inlet}}%
    \State Connect $V_k.\text{out} \rightarrow H_{e,k}.\text{in}$ \Comment{\textcolor{gray}{Valve outlet to evaporator inlet}}%
\EndFor
\State \textbf{Create suction manifold:}%
\State Set $p_\text{suct} = H_{e,1}.\text{out}$ \Comment{\textcolor{gray}{Suction manifold pressure node}}%
\For{$k = 2$ to $n_v$}
    \State Connect $H_{e,k}.\text{out} \rightarrow p_\text{suct}$ \Comment{\textcolor{gray}{Merge evaporator outlets}}%
\EndFor
\State \textbf{Connect compressor inlets:}%
\For{$k = 1$ to $n_c$}
    \State Connect $p_\text{suct} \rightarrow C_k.\text{in}$ \Comment{\textcolor{gray}{Suction manifold to compressor inlet}}%
\EndFor
\State \Return Configured system with topology $\mathcal{G}(n_c, n_v)$%
\end{algorithmic}
\end{algorithm}

The system solver automatically adapts to the increased number of junction pressures, where $n_p$ scales with $n_c$ and $n_v$ according to the system topology.
For systems with $n_p > 10$, the algebraic solver automatically switches from Powell hybrid method to bounded least-squares optimization (Levenberg--Marquardt) to ensure robust convergence.
For DAE solvers, the state vector dimension scales linearly with the number of heat exchangers: $\bm{y} \in \mathbb{R}^{2(n_\text{cond} + n_\text{evap})}$, where $n_\text{cond} = n_c$ and $n_\text{evap} = n_v$ for the parallel--merge architecture, and each heat exchanger contributes two differential variables (mass charge $M_r$ and internal energy $E_\text{hx}$).

\subsection{Stability and Solver Parameters}

To evaluate the stability and performance of our solvers on large-scale systems, we conduct simulations for systems ranging from $n_c = 2$ to $n_c = 16$ compressor--condenser pairs, with an equal number of valve--evaporator pairs ($n_v = n_c$).
Each simulation runs for $N_\text{steps} = 500$ time steps to assess both short-term stability and computational efficiency.
The solvers are configured with optimized parameters identified through Bayesian optimization (Section~\ref{sec:system_solver}), ensuring fair comparison across different system sizes.
The optimization parameters include tolerances ($\epsilon_{\Delta t}$, $\epsilon_\text{soln}$), step size bounds ($h_\text{max}$, $h_\text{min}$), and output intervals ($\Delta t_\text{out}^\text{IDA}$ for IDA, $\Delta t_\text{out,min}^\text{DASSL}$ and $N_\text{max}^\text{DASSL}$ for DASSL).

\begin{figure}[htbp]
    \centering
    \includegraphics[width=0.65\linewidth]{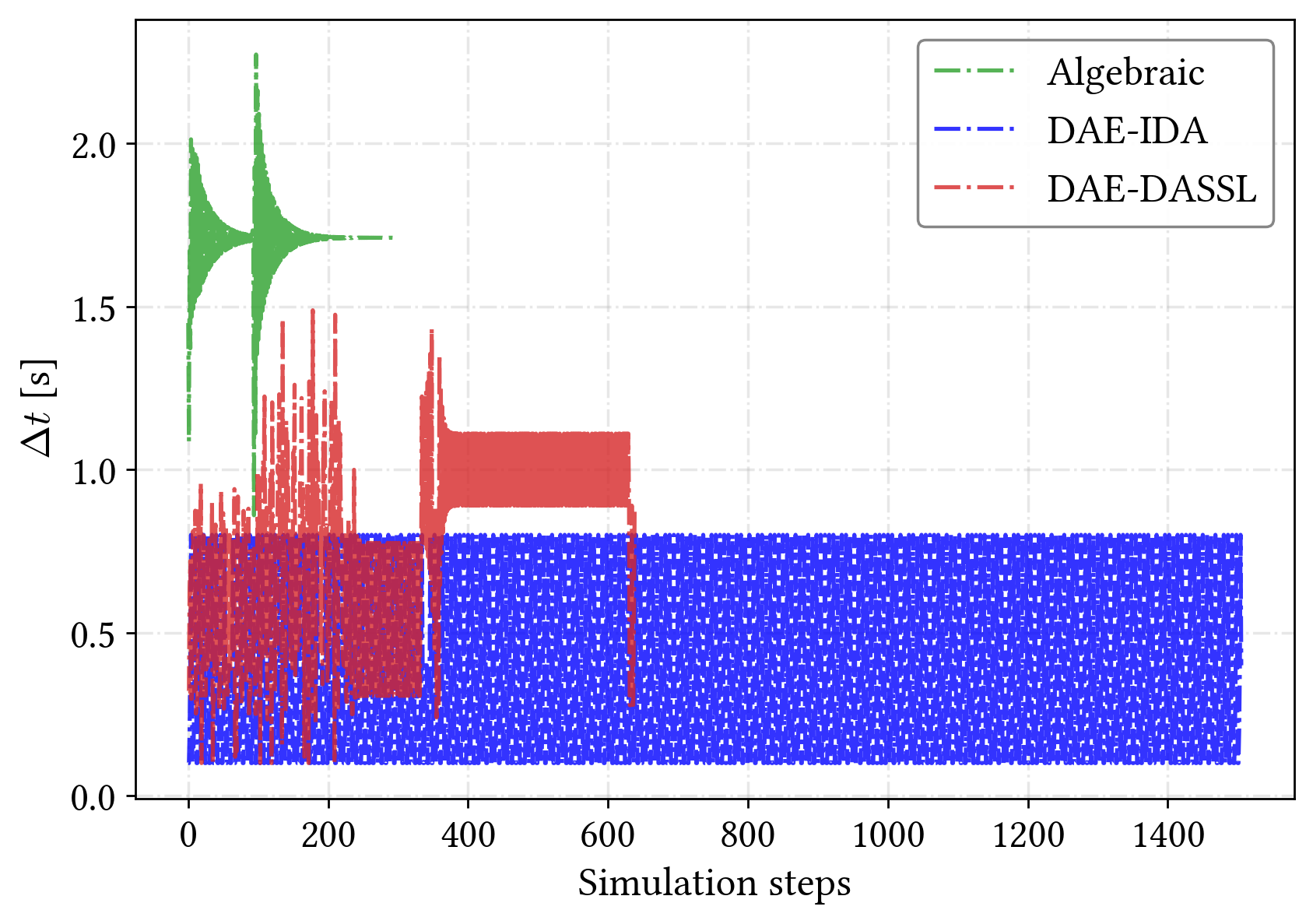}
    \caption{Time step ($\Delta t$) evolution comparison for a large-scale system with $n_c = 2$ compressors and $n_v = 2$ evaporators. The DAE-IDA solver achieves the finest temporal resolution with smaller, adaptive time steps, while the algebraic solver uses larger, more uniform steps. All three solvers successfully complete the $N_\text{steps} = 500$ simulation, demonstrating stability for medium-scale systems.}
    \label{fig:hvac_scaling_2x2_delta_t}
\end{figure}

Figure~\ref{fig:hvac_scaling_2x2_delta_t} compares the time step evolution for a system with $n_c = 2$ compressors and $n_v = 2$ evaporators across all three solver types.
The DAE-IDA solver achieves the finest temporal resolution with smaller, more adaptive time steps ($\Delta t \approx 0.1$--$1.0$ s), enabling better capture of transient dynamics.
The DAE-DASSL solver shows intermediate behavior with time steps in the range $\Delta t \approx 0.5$--$2.0$ s, while the algebraic solver operates with larger, more uniform time steps ($\Delta t \approx 1.0$--$5.0$ s) for computational efficiency.
All three solvers successfully complete the $N_\text{steps} = 500$ simulation, demonstrating stability for this system size ($n_c = n_v = 2$).

\begin{figure}[htbp]
    \centering
    \includegraphics[width=0.5\linewidth]{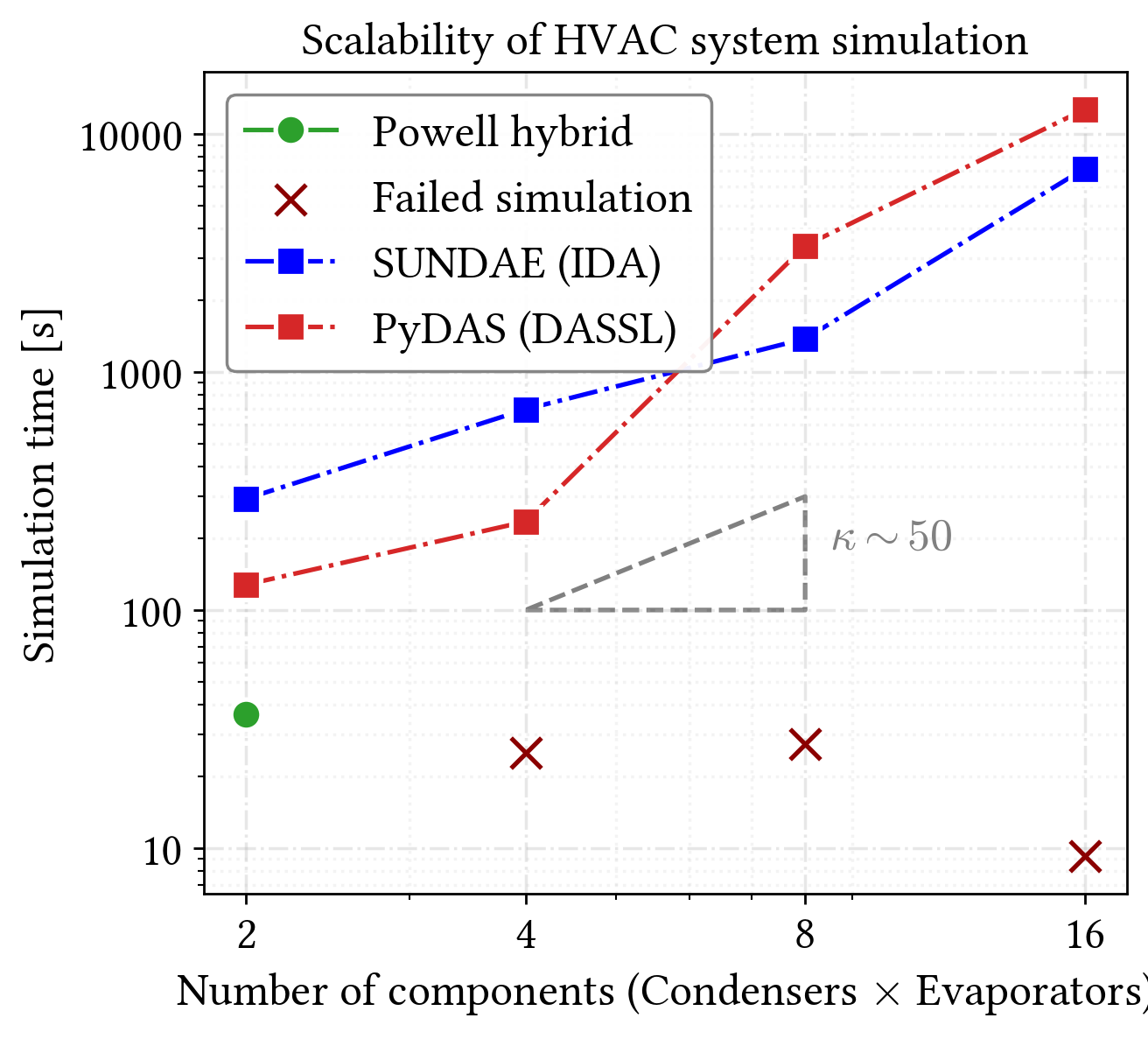}
    \caption{Computational scaling comparison for HVAC systems with increasing numbers of compressor--condenser pairs ($n_c$ ranging from 2 to 16). The simulation time $t_\text{simulation}$ exhibits clear superlinear growth with system size. Over the tested range, the increase in runtime is broadly consistent with polynomial scaling, with an effective exponent between quadratic and cubic, although the limited number of data points precludes a precise characterization. This trend reflects the increasing complexity of the coupled algebraic pressure system with $n_p = \mathcal{O}(n_c)$ junction pressures, as well as the growth of the state space $\bm{y} \in \mathbb{R}^{2(n_c + n_v)}$ for DAE solvers. Among the methods, the DAE-IDA solver demonstrates comparatively favorable scaling behavior, maintaining moderate simulation times even for larger systems (e.g., $n_c = 16$). Failed simulations are marked with red crosses.}
    \label{fig:hvac_scaling}
\end{figure}

Figure~\ref{fig:hvac_scaling} shows the computational scaling behavior as the number of components increases from $n_c = 2$ to $n_c = 16$.
Across all solvers, the simulation time $t_\text{simulation}$ exhibits clear superlinear growth with system size.
Over the range considered, the increase in runtime is broadly consistent with polynomial scaling, with an effective exponent between quadratic and cubic, although the limited number of data points precludes a precise characterization.
This trend is expected given the increasing complexity of the coupled algebraic pressure system with $n_p = \mathcal{O}(n_c)$ junction pressures, as well as the growth of the state space $\bm{y} \in \mathbb{R}^{2(n_c + n_v)}$ for the DAE solvers.
Among the methods, the DAE-IDA solver demonstrates comparatively favorable scaling behavior, maintaining moderate simulation times even for larger systems (e.g., $t_\text{simulation} \lesssim 200$~s at $n_c = 16$).
The algebraic solver performs competitively for small to medium system sizes ($n_c \leq 8$), but its runtime increases more rapidly at larger scales, indicating reduced scalability.
The DAE-DASSL solver exhibits intermediate behavior, with consistently higher computational cost than DAE-IDA but more stable scaling than the algebraic formulation across the tested range.

The time step evolution patterns observed in Figure~\ref{fig:hvac_scaling_2x2_delta_t} directly reflect the adaptive time-stepping mechanisms described in Section~\ref{sec:system_solver}: the algebraic solver's RK45 method adaptively adjusts $\Delta t$ based on local error estimates, while the DAE solvers' BDF methods use adaptive error control to select both step size and order, resulting in the distinct temporal resolution behaviors observed across system sizes.

The results demonstrate that our framework successfully scales to large HVAC systems, with the DAE-IDA solver providing the best balance between accuracy and computational efficiency across all system sizes tested.
This scalability is crucial for practical applications where HVAC systems may contain dozens of compressors and heat exchangers, enabling system-level optimization and control design for real-world installations.

\section{Conclusions\label{sec:conclusions}}

This work presents a comprehensive framework for modeling large-scale HVAC systems using physics-informed neural ODEs (PINODEs) integrated with differential-algebraic equation (DAE) solvers.
The framework combines data-driven component models with physics-based constraints, enabling accurate and efficient simulation of complex multi-compressor systems.
Our key contributions include: (1) physics-informed neural ODE models for individual heat exchangers that capture transient dynamics while respecting thermodynamic principles, (2) a corrector neural network that compensates for system-level errors by learning from short training segments, (3) integration of DAE solvers (IDA and DASSL) to handle algebraic constraints arising from pressure equilibrium and mass flow conservation, and (4) Bayesian optimization for automatic parameter tuning across multiple solver types.

The results demonstrate that the DAE-IDA solver achieves superior performance across all metrics, achieving an overall MAPE of 2.04\% with a simulation time of 58.91 seconds for the dual-compressor system, outperforming both the algebraic solver and DAE-DASSL solver.
The corrector network significantly improves prediction accuracy, reducing systematic biases in mass and energy predictions and bringing the system-level results into close agreement with reference Dymola simulations.
Bayesian optimization successfully identifies optimal solver parameters, revealing distinct optimal regions for each solver type and enabling automated tuning that balances accuracy and computational efficiency.

The framework demonstrates excellent scalability, successfully simulating systems with up to 16 compressor--condenser pairs while maintaining computational efficiency.
The DAE-IDA solver exhibits superior scaling behavior, maintaining reasonable simulation times even for large systems, while the algebraic solver shows steeper computational scaling due to the increasing complexity of the pressure system.
This scalability is crucial for practical applications, as real-world HVAC installations often contain dozens of compressors and heat exchangers, requiring efficient system-level simulation for optimization and control design.

The combination of physics-informed neural networks, system-level correction, and robust DAE solvers provides a powerful approach for modeling complex HVAC systems that was previously challenging with traditional physics-based methods.
This work opens new possibilities for system-level optimization, predictive control, and design exploration of large-scale HVAC installations, with potential applications in building energy management, fault detection, and real-time control systems.




\section*{Declaration of Competing Interests}%

None.%

\section*{CRediT authorship contribution statement}

Hanfeng Zhai: Conceptualization, Data curation, Formal analysis, Funding acquisition, Investigation, Methodology, Software, Validation, Visualization, Writing - original draft, Writing - review \& editing.

Hassan Mansour: Conceptualization, Data curation, Formal analysis, Funding acquisition, Investigation, Methodology, Software, Validation, Visualization, Writing - original draft, Writing - review \& editing.

Hongtao Qiao: Conceptualization, Data curation, Formal analysis, Funding acquisition, Investigation, Methodology, Software, Validation, Visualization, Writing - original draft, Writing - review \& editing.

Christopher Laughman: Conceptualization, Data curation, Formal analysis, Funding acquisition, Investigation, Methodology, Software, Validation, Visualization, Writing - original draft, Writing - review \& editing.
\section*{Data availability}

The data that has been used is confidential.

\section*{Appendix}

This appendix provides additional results for the outdoor heat exchanger (evaporator) PINODE model, complementing the indoor heat exchanger results presented in Section~\ref{sec:pinode_results}.
The outdoor heat exchanger model follows the same training procedure and architecture as the indoor model, ensuring consistency across both components used in the system-level simulations.

\begin{figure}[htbp]
    \centering
    \includegraphics[width=0.85\linewidth]{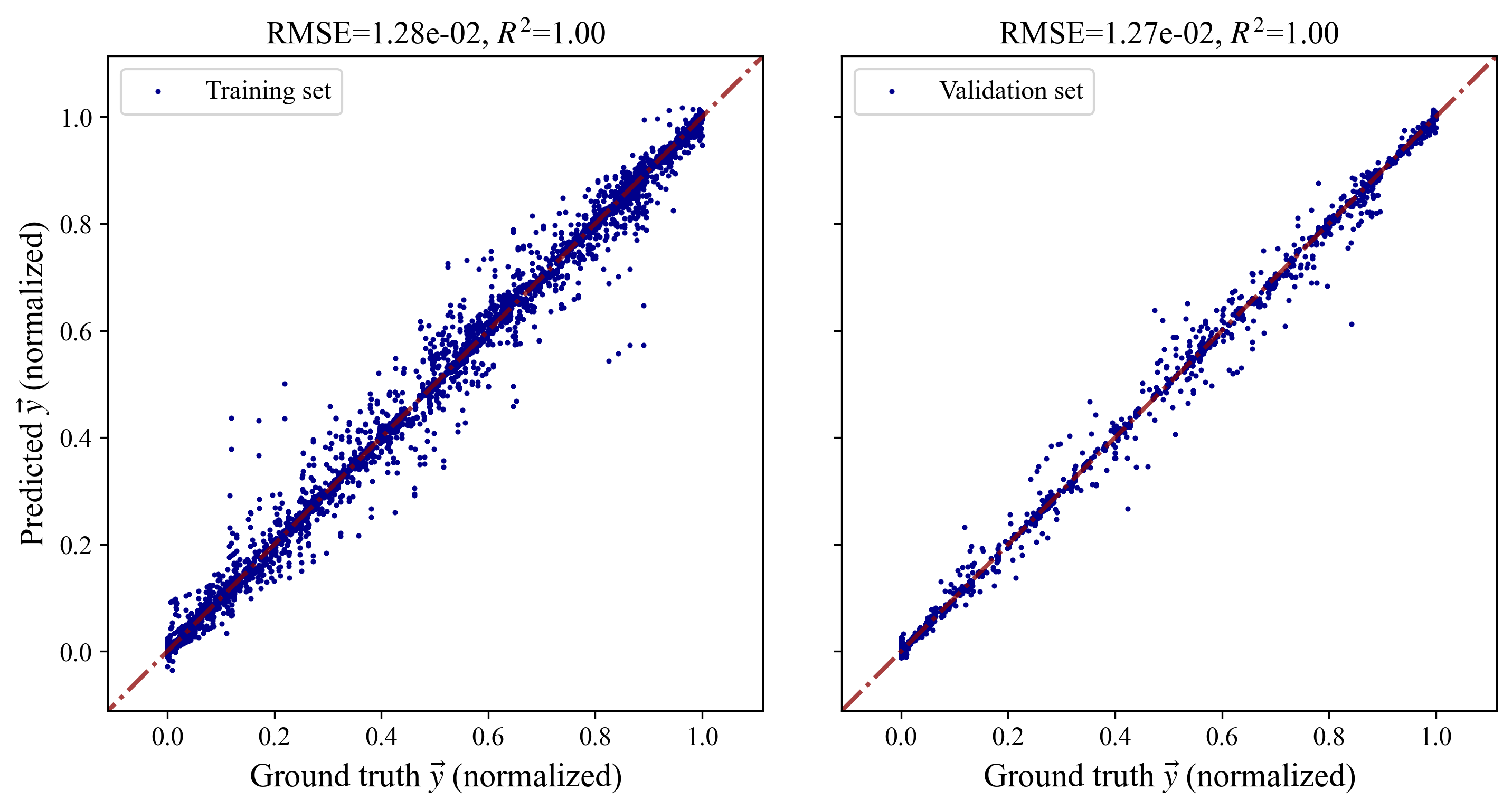}
    \caption{Parity plot comparing predicted versus true outputs for the training and testing sets of the evaporator (outdoor heat exchanger) PINODE model. The model demonstrates excellent agreement with the reference data across both datasets, with data points closely aligned along the diagonal, indicating high prediction accuracy for the outdoor heat exchanger component.}
    \label{fig:parity_outdoor}
\end{figure}

Figure~\ref{fig:parity_outdoor} shows the parity plot for the outdoor heat exchanger model, demonstrating excellent performance on both training and testing datasets.
The strong linear correlation and tight clustering around the diagonal indicate that the PINODE model successfully captures the dynamics of the evaporator component, providing accurate predictions for use in system-level simulations.

\begin{figure}[htbp]
    \centering
    \includegraphics[width=0.85\linewidth]{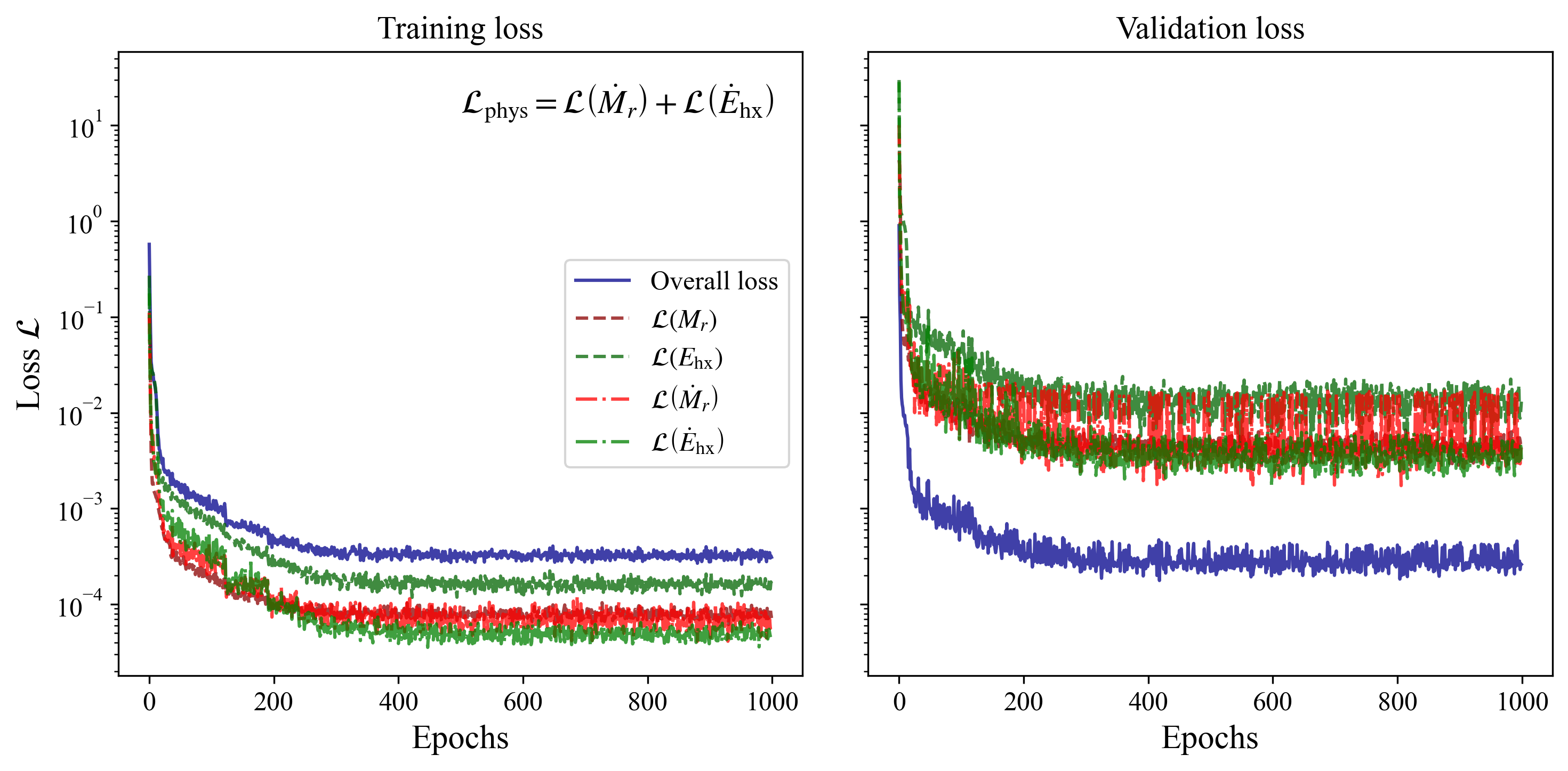}
    \caption{Loss history for the training and testing sets of the evaporator (outdoor heat exchanger) PINODE model. The stable convergence with minimal gap between training and testing losses indicates good generalization and minimal overfitting, validating the model architecture and training procedure for the outdoor heat exchanger component.}
    \label{fig:loss_outdoor}
\end{figure}

The loss history in Figure~\ref{fig:loss_outdoor} confirms stable convergence with minimal overfitting, as evidenced by the parallel trajectories of training and testing losses throughout the training process.
This validates the model architecture and training procedure for the outdoor heat exchanger, ensuring that both indoor and outdoor components are well-trained and ready for integration into the system-level simulation framework.


\end{document}